\pdfoutput=1

\documentclass[11pt]{article}

\usepackage[preprint]{acl}

\usepackage{times}
\usepackage{latexsym}

\usepackage[T1]{fontenc}

\usepackage[utf8]{inputenc}

\usepackage{microtype}

\usepackage{inconsolata}

\usepackage{graphicx}
\usepackage{subcaption}
\usepackage{booktabs}
\usepackage{colortbl}
\definecolor{LightCyan}{rgb}{0.88,1,1}
\usepackage{tcolorbox} %
\tcbuselibrary{most}
\usepackage[inline]{enumitem}

\title{Training Bilingual LMs  with Data Constraints in the Targeted Language}

\author{Skyler Seto\thanks{Equal contribution}, Maartje ter Hoeve\footnotemark[1], Richard He Bai, Natalie Schluter, David Grangier \\
  Apple \\
  \texttt{\{sseto,m\_terhoeve,richardbai,natschluter,d\_grangier\}@apple.com}}

\begin{document}
\maketitle
\begin{abstract}
    Large language models are trained on massive scrapes of the web, as required by current scaling laws. Most progress is made for English, given its abundance of high-quality pretraining data. For most other languages, however, such high quality pretraining data is unavailable. 
    In this work, we study how to boost pretrained model performance in a target language with insufficient pretraining data for training a high performing language model, by enlisting data from an auxiliary language for which high quality data is available. 
    We study this by quantifying the performance gap between training with data in a data-rich auxiliary language compared with training in the target language, exploring the benefits of translation systems, studying the limitations of model scaling when data is limited in the target languages, and proposing new methods for upsampling data from the auxiliary language.
    Our results show that stronger auxiliary datasets result in performance gains without modification to the model or training objective for close languages, and, in particular, that performance gains due to the development of more information-rich English pretraining datasets can extend to targeted language settings with limited data.
\end{abstract}

\section{Introduction}
\label{sec:intro}

The abundance of high quality English data has given rise to exceptional language modeling performance in English~\citep{brown2020language,bubeck2023sparks, OpenAI2023GPT4TR}. However, there are many other languages for which a reasonably large amount of data is available, yet not in equally large amounts as for English. Taking mC4~\cite{xue2020mt5} as an example, languages like German, French, Chinese, and Japanese have $\sim$10-100 times less data than English.
Consequently, most non-English progress comes from relatively small bilingual models~\cite[e.g.,][]{le2019flaubert, de2019bertje, martin2019camembert, scheible2020gottbert, wei2023skywork, faysse2024croissantllm}, or larger massively multilingual models~\cite[e.g.,][]{le2023bloom, intrator-etal-2024-breaking, ustun-etal-2024-aya}. Other LLMs such as Llama-2, GPT-3, and PaLM-2 that perform well across a variety of languages are trained primarily on English, with less than 20\% of data from other languages \cite{xu2024survey}. 

Several works study cross-lingual transfer, but are limited to finetuning smaller models ($\sim$100M parameters) with limited data, or classical NLP tasks, such as part-of-speech tagging, named entity recognition, and natural language inference~\cite[e.g.,][]{chang2023multilinguality,de2022make,faisal2024efficient}. They do not evaluate pretrained language models on contemporary knowledge-based downstream tasks. It is still unclear how to optimally make use of bi- or multilingual data, to optimally increase performance in a target language.  This is especially relevant in data constrained settings. For example, imagine one has access to only $500$K documents in a target language, and wants to pretrain a language model for this language. $500$K documents is sufficient to attain a reasonable perplexity score, but this amount of data is insufficient to do well on modern knowledge-based downstream tasks. Many languages in the tail of mC4 fall in this category. Currently, a systematic analysis of how to use an auxiliary language in this scenario is lacking, leaving practitioners to a trial-and-error based approach, mostly based on intuition.

In this work, we provide such a systematic analysis. In particular, we are interested in whether ``higher quality'' auxiliary data also leads to higher model performance in the target language. We choose English as the main auxiliary language, matching a common practical setting. To investigate the effect of linguistic similarity, we also experiment with Chinese as the auxiliary language, although being somewhat bottlenecked by the amount of available data in Chinese.

\begin{figure*}[ht]
    \centering
       \begin{subfigure}{0.32\textwidth}
        \centering
        \includegraphics[width=\textwidth]{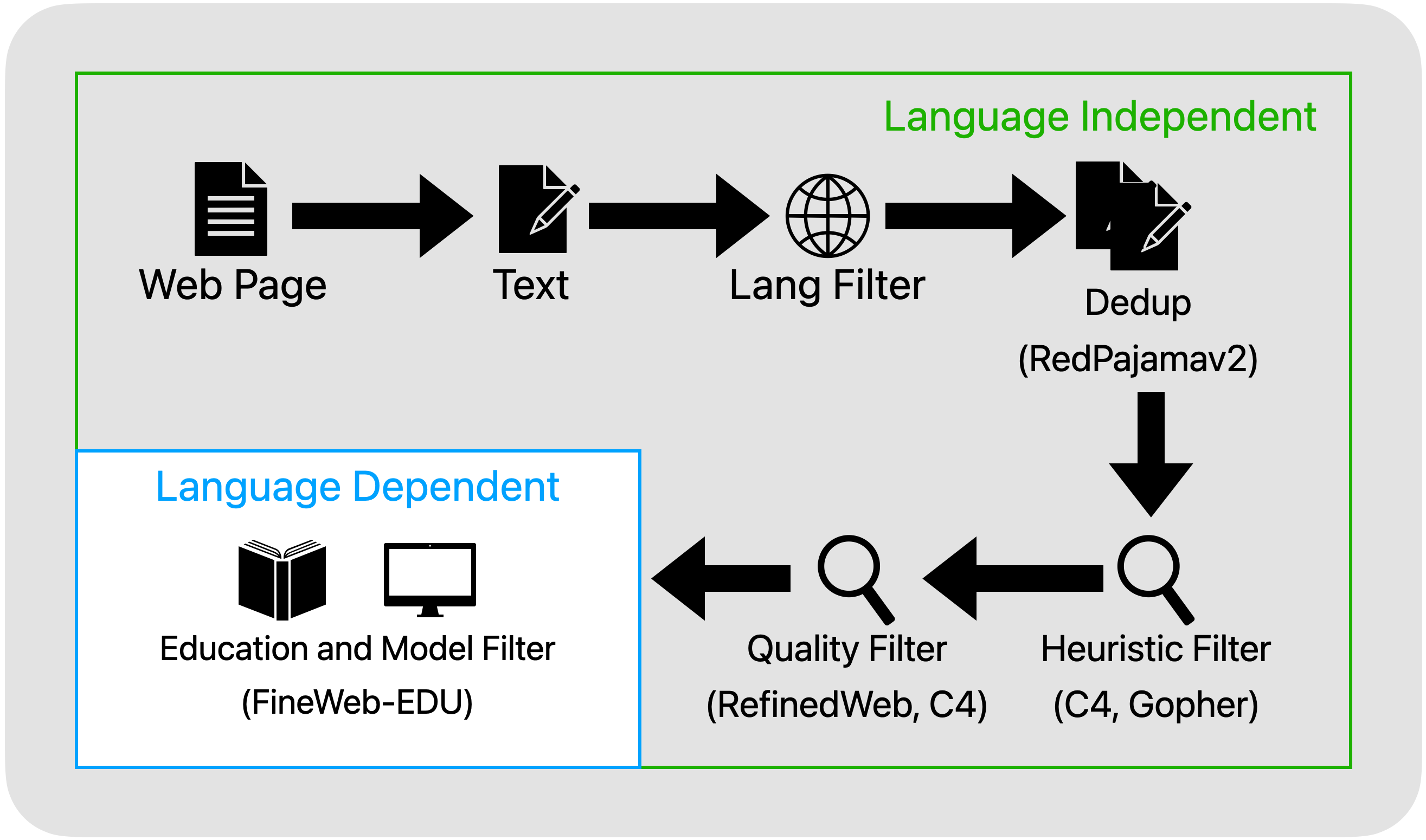}
        \caption{Data Pipeline}
        \label{fig:data_pipeline}
    \end{subfigure}
    \hfill
    \begin{subfigure}{0.32\textwidth}
        \centering
        \includegraphics[width=\textwidth]{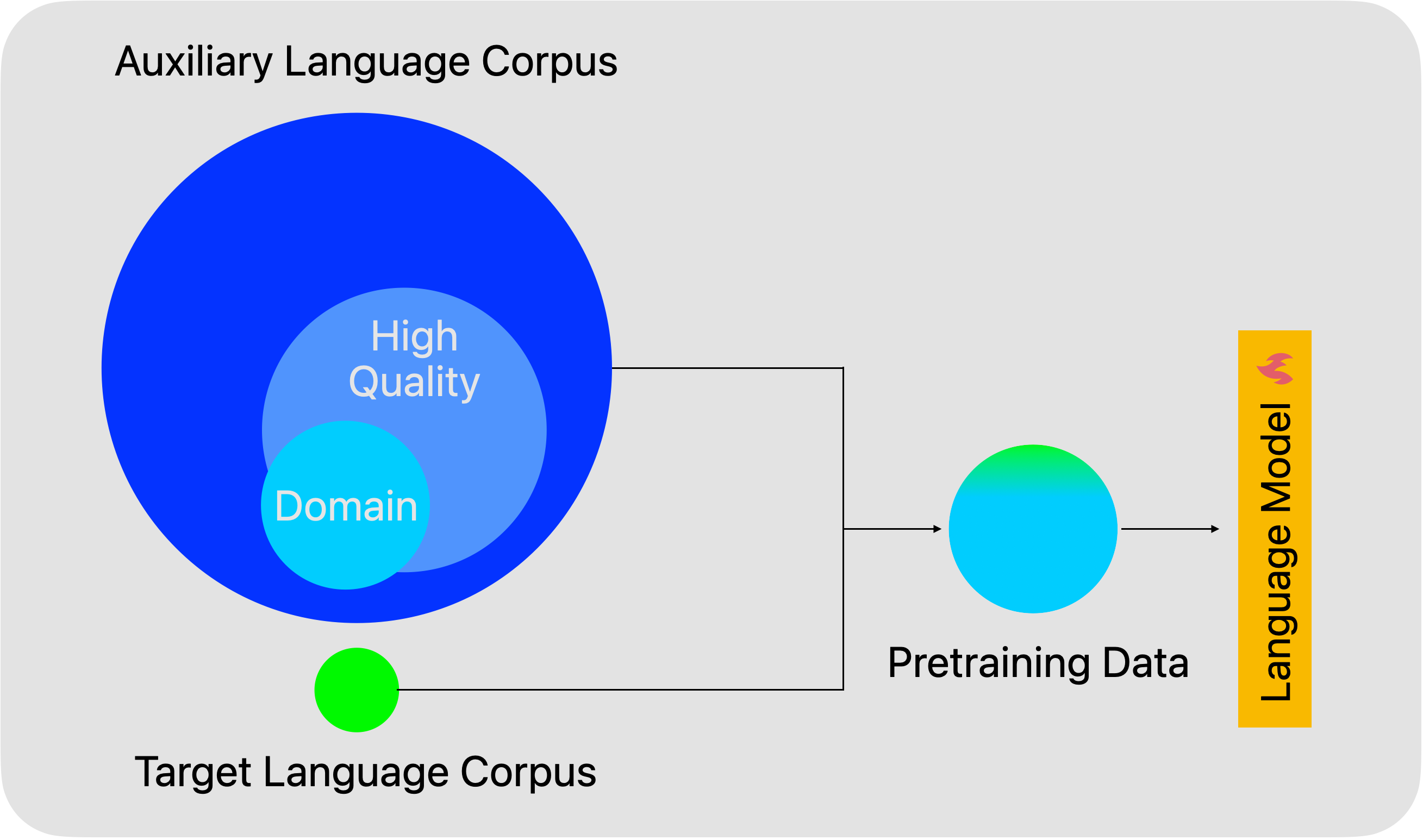}
        \caption{Auxiliary Data Pretraining}
        \label{fig:aux_pretrain}
    \end{subfigure}
    \hfill
    \begin{subfigure}{0.32\textwidth}
        \centering
        \includegraphics[width=\textwidth]{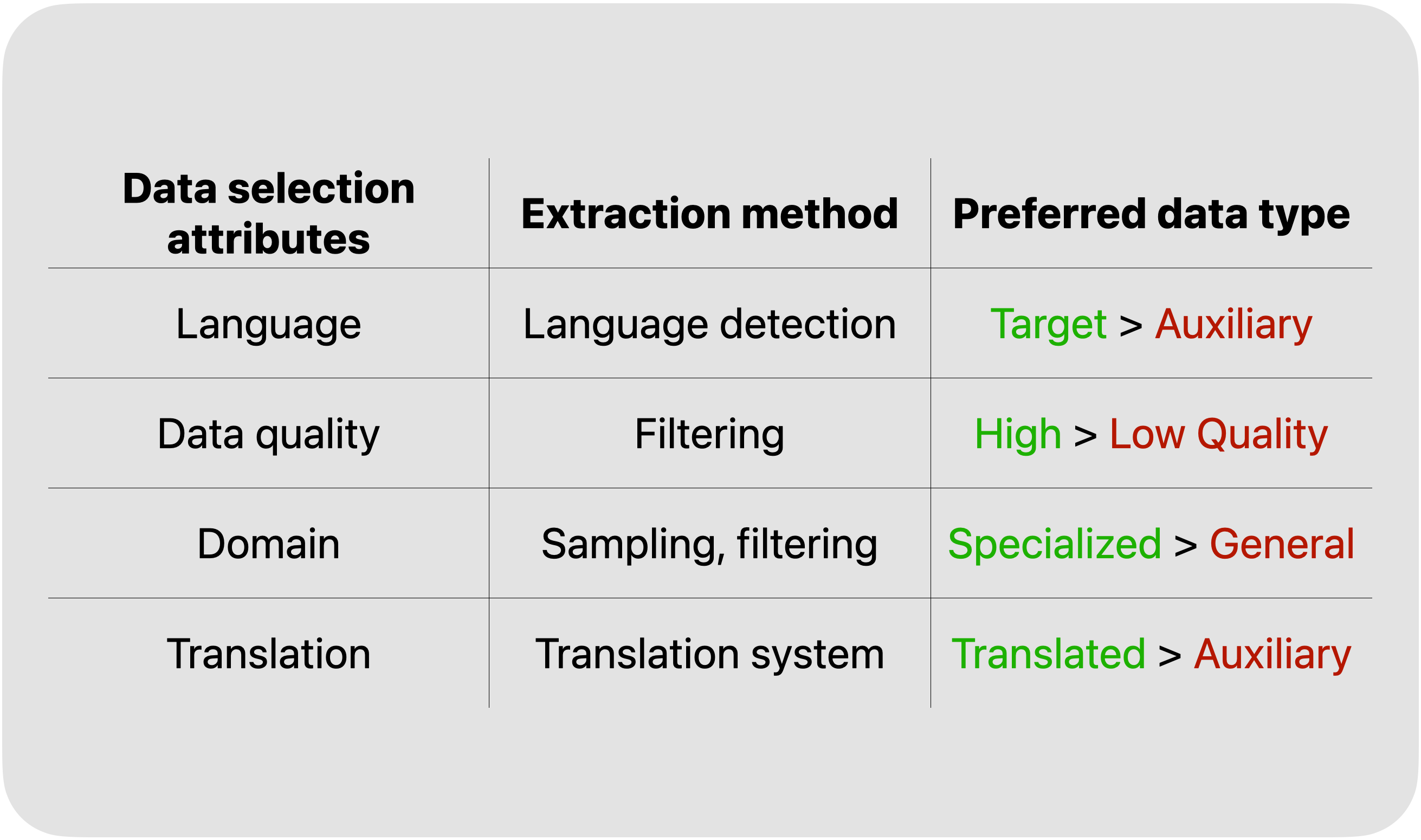}
        \caption{Data Transforms}
        \label{fig:data_transforms}
    \end{subfigure}
       
    \caption{(a) \textbf{Data Pipeline}: English data pipeline used for building large pretraining corpora in \citep{penedo2024fineweb}. (b) \textbf{Auxiliary Data Pretraining}: Combining high quality domain-specific pretraining data with a small amount of data from the target language for pretraining with limited target data. (c) \textbf{Data Transforms}: Many considerations when building datasets in languages with limited data.}
    \label{fig:overview}
\end{figure*}

To measure the impact of ``better'' auxiliary data, we explore well-tested filtering techniques for a monolingual, English, setting. We focus on the impact of dataset size, filtering for data quality and style, and data selection for specialized information relevant to downstream evaluations (Figure~\ref{fig:data_transforms}), matching recent advancements in state-of-the-art open-source English datasets such as FineWebEDU \cite{penedo2024fineweb}, and DataComp \cite{li2024datacomp} (Figure~\ref{fig:overview}). 
Currently, it is still unclear how these techniques extend to a bilingual setting;
as we confirm in our experiments, not every filtering technique that performs well in a monolingual setting performs well in a bilingual setting. 
Specifically, our \textbf{key findings} are:

\begin{enumerate}[nosep,leftmargin=*]
    \item Auxiliary English data that is generated by some of the existing model-based data filtering pipelines for English \textit{can} be helpful to complement limited data in a target language (\S\ref{sec:dataset_comparisons});
    \item Findings are not the same across multiple target languages. We hypothesize that for languages that are ``far'' from English, gains from better English datasets do not help (\S\ref{sec:multiple_languages});
    \item When training with higher quality auxiliary data, slightly more gains can be attributed to \textit{relevant information} in the auxiliary data (\S\ref{sec:mb_filter}-\ref{sec:is_information});
    \item There are limits to the model sizes that are practical to pretrain with limited target data: data size should scale linearly with model size \cite{kaplan2020scaling}, and performance in the target language saturates without increasing target data, regardless of increasing auxiliary data (\S\ref{sec:model_size_scaling}).
\end{enumerate}

\section{Related Work}
\label{sec:rw}

\paragraph{Multilingual Language Models} While much of the LLM research has focused on English, large-scale transformer-based multilingual language models have been trained on large multilingual corpora including mBERT \cite{pires2019multilingual}, XLM \cite{conneau2019cross}, mT5 \cite{xue2020mt5}, PolyLM \cite{wei2023polylm}, and Bloom \cite{le2023bloom}.  These works focus on training models that are language balanced and can reason in over 100 languages.  Other SOTA language models such as Llama 2 \cite{touvron2023llama}, Falcon \cite{almazrouei2023falcon}, and Palm 2 \cite{anil2023palm} have multilingual capabilities, but over 90\% the training data is English, and these models perform poorly across a variety of languages, such as south east Asian languages \cite{nguyen2023seallms}.

Other works focus on training smaller bilingual language models in French \cite{faysse2024croissantllm,le2019flaubert,martin2019camembert}, German \cite{scheible2020gottbert},  Dutch \cite{de2019bertje}, or Chinese \cite{wei2023skywork}, but require a substantial amount of bilingual data: English and the respective language for pretraining.  Other works focus on understanding languages LLMs reason in \cite{wendler2024llamas}, and languages LLMs cannot learn \cite{borenstein2024languages,kallini2024mission}. Still, little work has examined how information seen during pretraining in one language can help down-stream task performance in another language.

\paragraph{Cross-Lingual Transfer} \citet{philippy-etal-2023-towards} present a comprehensive survey on cross-lingual transfer in multilingual language models. The survey explains that cross-lingual transfer is a well studied topic for classic NLP tasks, such as part-of-speech tagging, named entity recognition, dependency parsing, machine translation, etc., although findings are not always consistent. %
Cross-lingual transfer is less well studied for the language modeling objective, and for modern downstream evaluation tasks, such as ARC~\cite{clark2018think}, HellaSwag~\cite{zellers2019hellaswag}, PIQA~\cite{bisk2020piqa}, SCIQ~\cite{welbl2017crowdsourcing}, WinoGrande~\cite{sakaguchi2021winogrande}, etc. \citeauthor{philippy-etal-2023-towards} end with a number of recommendations, one of which is to study cross-lingual transfer in more detail for generative models, given their recent exceptional performance. Our work differs from prior work on cross-lingual transfer in three ways: 
\begin{enumerate*}[label=(\roman*)]
    \item we do not directly study cross-lingual transfer as we do not first train on the auxiliary language, and then on the target language,
    \item we pretrain 1B and 3B parameters models that are 10-30x the size of typical BERT models, and
    \item instead of finetuning, we focus on how models learn to share between languages during pretraining.
\end{enumerate*}

\paragraph{Data Selection} Selecting high quality data for pretraining LLMs remains an active area of research.  Early research on data selection was based on heuristics.  For example, the original GPT-2 model was pretrained on outbound links filtered from Reddit, based on heuristic indicators of whether users found the links interesting, educational, or funny \cite{radford2019language}. Prior approaches also upsampled documents from high quality sources like Wikipedia   \cite{gururangan2022whose}, or used combinations of heuristics such as absence of stop words,  document length, word length, etc. \cite{rae2021scaling}. Other works select data based on quality filters and time of collection \cite{longpre2023pretrainer}, model-based quality filtering \cite{sachdeva2024train,li2024datacomp}, or textbook quality knowledge \cite{gunasekar2023textbooks,li2023textbooks,kong2024large}.  An alternative approach to data selection is re-weighting data samples to select the best data mixtures for training \cite{fan2023doge,xie2024doremi}, or importance sampling based on a downstream task \cite{grangier2024specialized,grangier2024task,xie2023data}.  Still a majority of these filtering techniques are applied to English-only datasets, and multilingual datasets such as mC4 have limited data filtering \cite{xue2020mt5}.

\section{Using English Data Selection Pipelines to Complement Limited Target Data}
\label{sec:exp}

Existing data selection pipelines have been shown to be effective in monolingual (English) pretraining. We investigate whether these pipelines are useful in the bilingual setup with limited target data. 
In \S\ref{sec:dataset_comparisons}, we report our findings for experiments with English as the auxiliary language and German as the example target language. In \S\ref{sec:multiple_languages} we discuss how our findings extend across multiple languages.

\subsection{General Implementation Details}
\label{sec:general}
\paragraph{Model.}
  We train decoder-only transformer models \cite{vaswani2017attention} with 1.3B parameters. Models use the PolyLM tokenizer \cite{wei2023polylm}, with a total vocabulary size of 256K tokens using BPE.
 Models are trained for 100K steps  with batch size 1024.
 Additional hyperparameters and model details are in Appendix~\ref{sec:hyperparams}. We report results for the 1.3B model in the main body, and refer to Appendix~\ref{sec:app_300M} for results on a 300M model trained for 30K steps. These model sizes are chosen as they provide reasonable (above random) performance on several benchmark QA tasks, and are commonly used for benchmarking and ablating pretraining of language models \cite{penedo2023refinedweb,penedo2024fineweb}. %

\paragraph{Data.}
To meaningfully and systematically study a data constrained scenario, we impose the following trade-off:
\begin{enumerate*}[label=(\roman*)]
    \item the number of tokens in the target language needs to be sufficient to start training a language model in the target language (i.e., perplexity needs to go down sufficiently), and
    \item the number of tokens in the target language should be small enough to expect meaningful differences from the auxiliary language.
\end{enumerate*}
Through empirical exploration, we choose to include 250M tokens in the target language. Not only does this choice satisfy the trade-off posed above, but it also simulates a realistic scenario, as 250M tokens is representative for the amount of data that exists in the tail of mC4 ($\sim$20 languages).
For our experiments, we choose languages with more than 250M tokens available in mC4. This allows us to compare with a monolingual setting, to better assess the impact of the auxiliary language.

We consider three baselines: 
\begin{enumerate*}[label=(\roman*)]
\item \textit{Target (S)}: 250M tokens from the target language, 
\item \textit{Target (L)}: enough target language for 1 training epoch, and
\item \textit{Auxiliary (L)}: enough auxiliary language for 1 training epoch.
\end{enumerate*}
For the other experiments, we consider $\sim$250M tokens from the target language, as stated above, which is repeated to account for 5\% of the training steps. The remaining 95\% of the training steps consists of the auxiliary data.  We refer to the small 250M target language data as ``(S)'' and to the large (auxiliary) data pool as ``(L)''.

\paragraph{Evaluation.}
We consider the average over six general understanding QA tasks: ARC-Easy, ARC-Challenge, Hellaswag, PIQA, SciQ, and Winogrande. These are knowledge-based tasks that small models with limited data still perform well on. %
Non-English evaluations are conducted via translation of the original dataset for which we use a mix of proprietary large language models. Additional details are provided in \S\ref{app:eval_desc}.

\subsection{Better English Datasets}
\label{sec:dataset_comparisons}

\paragraph{Methodology.} We first compare the performance of models trained on combinations of German (target) and English (auxiliary) with varying existing English datasets based on the common crawl: mC4 \cite{xue2020mt5}, RedPajamav2 (RPJv2) \cite{together2023redpajama}, RefinedWeb (RFW) \cite{penedo2023refinedweb}, and FineWebEDU (FWE)~\cite{penedo2024fineweb}. These datasets have been constructed following a pipeline of different filtering steps outlined in Figure~\ref{fig:data_pipeline}. The resulting datasets are of higher quality and cover different snapshots of the common crawl. %
Further details are available in Appendix~\ref{sec:app_dataset}. %

\begin{figure}
    \centering
    \includegraphics[width=\columnwidth]{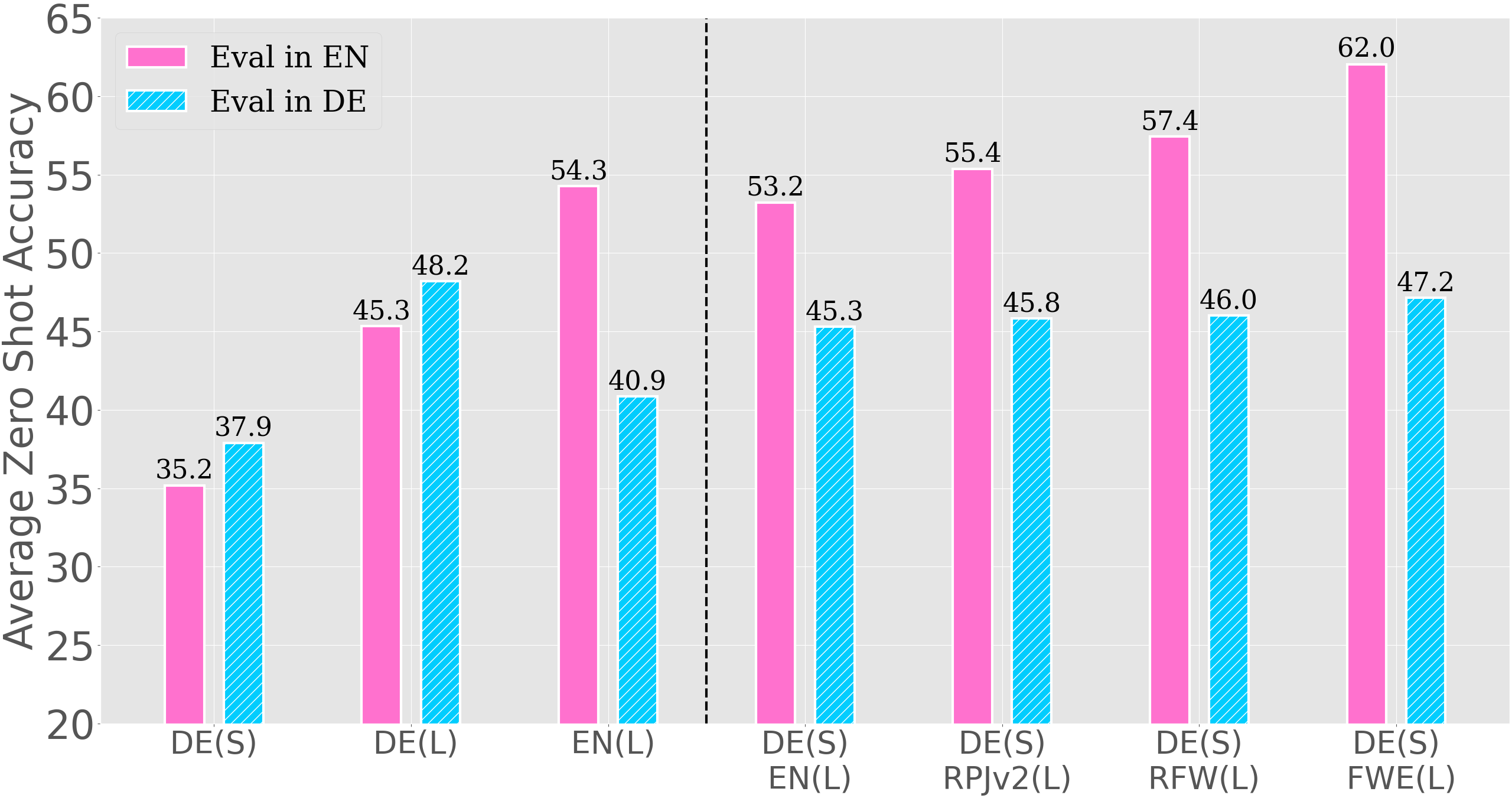}
    \caption{Zero-shot accuracy of models trained with  higher quality English auxiliary data.  Results are averaged over six eval datasets. We compare training with different auxiliary datasets on English and German evaluations. Better English datasets show large increases in English and smaller increases in German.}
    \label{fig:diff_datasets}
    \vspace{-5mm}
\end{figure}

\paragraph{Findings.}  We report results in Figure~\ref{fig:diff_datasets}. For English evaluations, better quality English datasets attain substantially higher performance on the English tasks (up to 9\%). For the same benchmarks translated into German, the performance increase is around 2\%. Of all compared English datasets, FineWebEDU achieves the best average down-stream performance, and within 1\% of the DE (L) comparison. There are two primary factors we hypothesize contribute to FineWebEDU achieving better performance on downstream tasks: high-quality data filtering, and relevant information filtering. We investigate this in more detail in \S\ref{sec:methods_improve}.

\subsection{Experiments Across Multiple Languages}
\label{sec:multiple_languages}

\begin{figure}[ht]
    \centering
    \includegraphics[width=\linewidth]{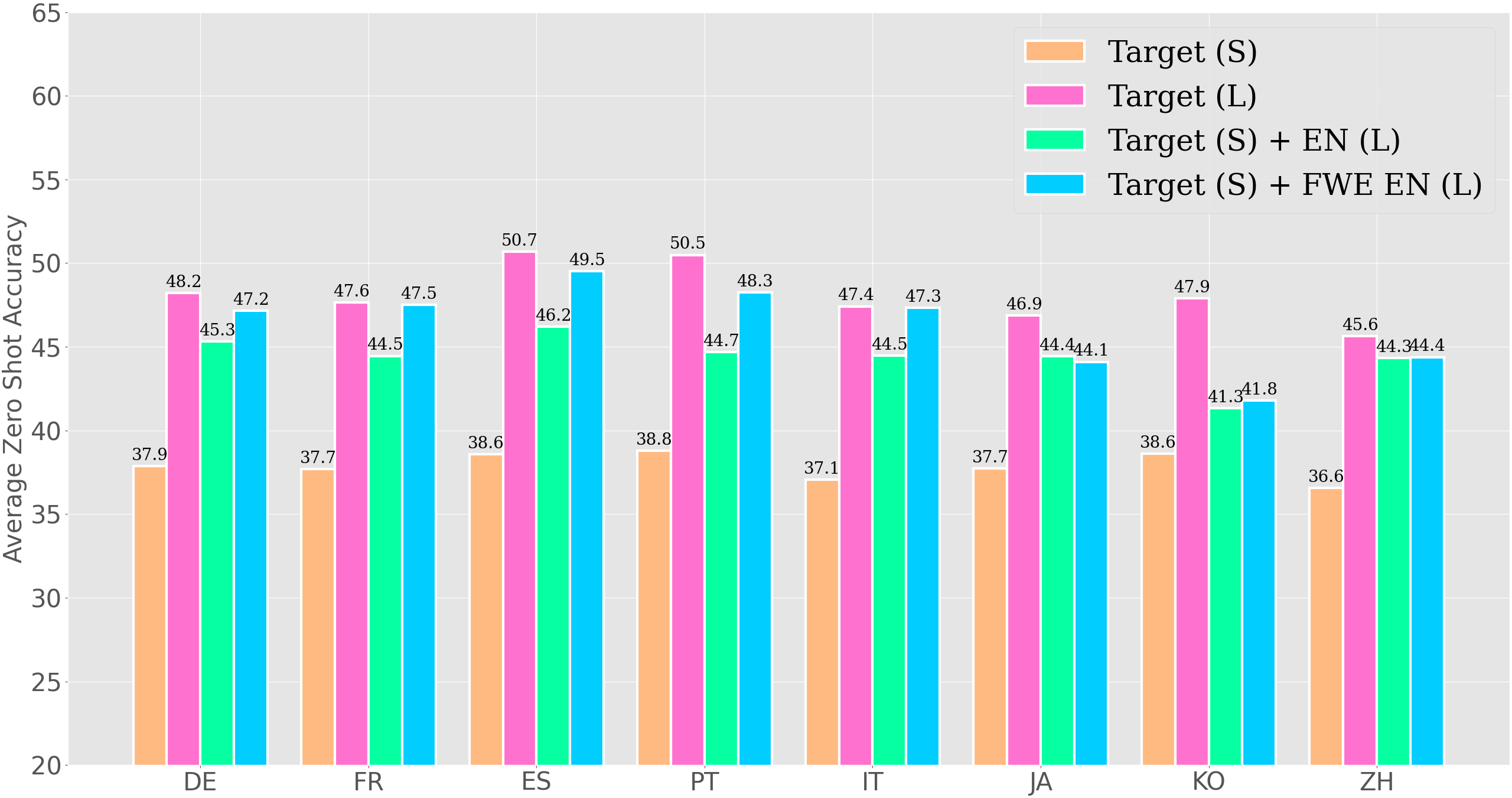}
    \vspace{-5mm}
    \caption{Average zero-shot accuracy in the target language summarized for eight languages. Models trained on 100B tokens. Comparisons between a small and large amount of monolingual data from the target language, a small amount of data from the target language and mC4 English data (same distribution), and a small amount of data from the target language and FineWebEDU.}
    \label{fig:monolingual_summary}
    \vspace{-4mm}
\end{figure}

\paragraph{Motivation.} We add seven languages, across four language families:  French, Italian, Portuguese, Spanish  (Indo-European, same as German), Chinese (Sino-Tibetan), Japanese (Japonic), and Korean (Koreanic) \cite{ethnologue2015}. We choose these languages because of their variety in language families, the amount of data available in mC4, and the access to translated evaluation data. %

\paragraph{Methodology.} %
We train with approximately 250M tokens\footnote{Note that data distributions from the $\sim$250M data in mC4 looks similar for each language and are not specialized to any particular domain as evidenced by the cluster distribution visualized in Appendix E.} from the mC4 corpus in the respective language and denote these models as ``target language (S)''.  The monolingual models are denoted as ``target language (L)''. The Chinese and Korean mC4 corpora contain fewer than 100B tokens and thus the data is repeated for multiple epochs in the base language experiments. Based on our previous results, we use FineWebEDU as the auxiliary language source. We also conduct additional experiments with Chinese as the auxiliary language and Japanese as the target language, to investigate the effect of linguistic similarity. %
Although beyond the immediate scope of our investigation, we also investigate a multilingual setup (instead of bilingual) in Appendix~\ref{sec:multilingual_mix}.

\begin{figure}[ht]
    \centering
    \includegraphics[width=\linewidth]{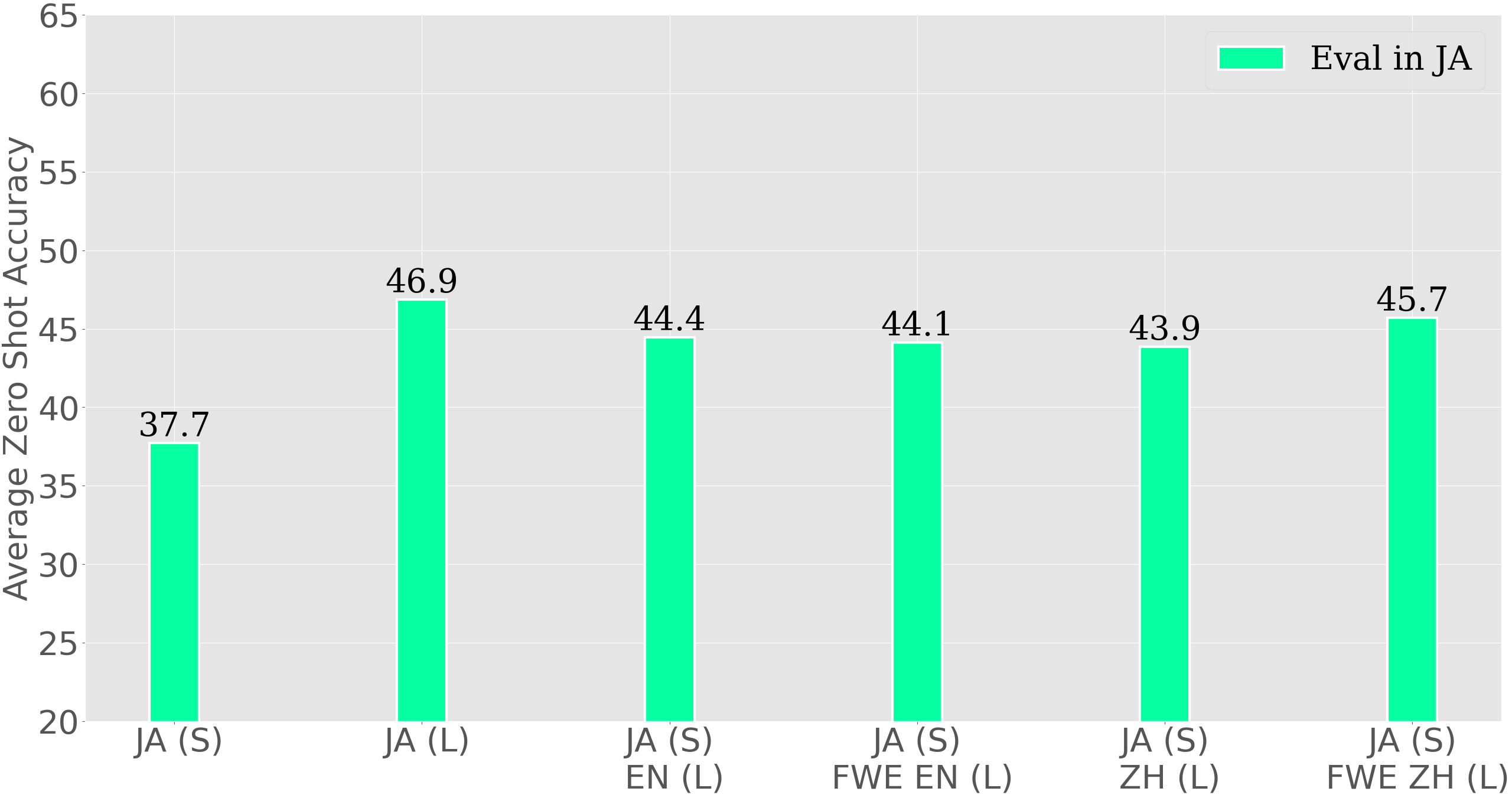}
    \vspace{-5mm}
    \caption{Average accuracy over zero-shot benchmark tasks in translated Japanese, comparing Chinese and English auxiliary data.}
    \label{fig:aux_ja_chinese_1B}
    \vspace{-2mm}
\end{figure}

\paragraph{Findings.}
We report results for multiple target languages and Chinese auxiliary data in Figure~\ref{fig:monolingual_summary}-\ref{fig:aux_ja_chinese_1B}.
We only observe improvements from FineWebEDU for the Indo-European languages in our set, with English as the auxiliary language (Figure~\ref{fig:monolingual_summary}).
Figure~\ref{fig:aux_ja_chinese_1B} shows better performance for Chinese instead of English FineWebEDU as the auxiliary language for Japanese. This is a first indication of the applicability of using different auxiliary languages, based on linguistic similarity.

\section{The Effect of Individual Data Transformations}
\label{sec:methods_improve}

In \S\ref{sec:dataset_comparisons} we investigated the effect of using existing data filtering pipelines in the auxiliary language, but this does not isolate which components of the pipeline provide the best downstream improvements. Here, we study individual data transformations that comprise the training pipeline, including model based quality filtering and educational content upsampling, as well as transformations on top of auxiliary data,  such as translation.
For all experiments, we use the same experimental setup and baselines as in \S\ref{sec:exp}, and we focus on German as the target and English as the auxiliary language.

\subsection{High Quality Filtering} 
\label{sec:mb_filter}

\paragraph{Motivation.}
Prior work shows that model-based filters and higher quality data lead to better performance in English.  A key component of FineWeb is custom filters based on text quality \cite{penedo2024fineweb}, and DCLM models are trained on data which has been highly filtered based on heuristic and model-based filters \cite{li2024datacomp}. To isolate the impact high quality auxiliary data can have on target language performance, we apply a recent model-based quality filter on top of mC4 English. We choose this filter as it aims primarily at stylistic quality, and filtering with this model leads to better performance than training with RFW and FWE~\cite{li2024datacomp}, two of the best performing datasets in Figure~\ref{fig:diff_datasets}. Note that while filtering strategies achieve strong results on English downstream evaluations, training a filtering model can require more data than available, and high quality datasets may not be available in the target language. %

\paragraph{Methodology.}
To test the impact of quality filtering, we use the DCLM fast text classifier. This filters data that is aimed at instruction following and high scoring posts in {\tt r/ExplainLikeImFive}, and was found to be the best quality filter over RFW and other model-based filters~\cite{li2024datacomp}.  %
We compare the performance of models trained with English filtering and without, holding the German data the same for evaluation.  We refer to this as DCLM Filter (following~\cite{li2024datacomp}) and filter to the top 10\% of mC4.

\begin{figure}[t]
    \centering
    \includegraphics[width=\columnwidth]{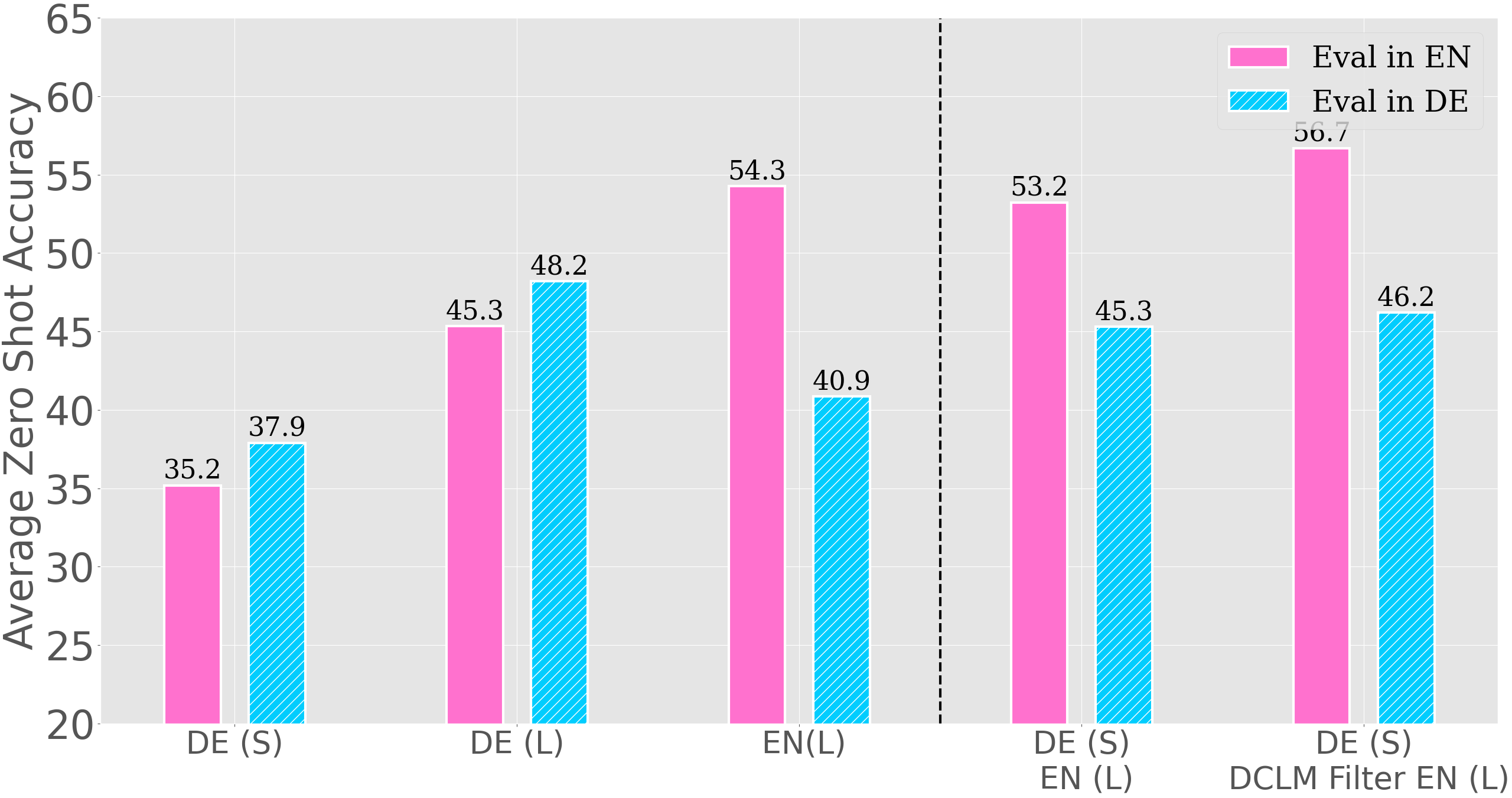}
    \caption{Zero-shot accuracy of models trained with model based filtering of English auxiliary data.  Results are averaged over six evaluation datasets. For each setting evaluation is done in English and German.}
    \label{fig:mb_filter}
    \vspace{-2mm}
\end{figure}

\paragraph{Findings.}
The results are in Figure~\ref{fig:mb_filter}. 
English evaluations improve by 3\%. Translated German evaluations are under 1\%, within 1 standard error.

\subsection{Clustered Dataset Importance Sampling} 
\label{sec:is_information}

\paragraph{Motivation.}
Prior work shows that LLMs reason in English and that information may be stored in a language agnostic space \cite{wendler2024llamas}. However, it is unclear whether the relevant information is only seen in English (the predominant language), or also in other languages. Our experiments on FineWebEDU indicate the presence of information sharing, but FineWebEDU filters for both high quality data and for educational quality content~\cite{penedo2024fineweb}. In this section we experiment with isolating the information sharing.

\paragraph{Methodology.}
To explicitly test whether information is shared between auxiliary and target languages,  we upsample topics in English and evaluate on the target language. Given access to some small target set, importance sampling weights are computed based on the amount of data from a target evaluation set that is assigned to each cluster following \cite{grangier2024specialized}.  

To train the clustering model, we take a small subset of the training set, produce embeddings from a smaller sentenceBERT model~\cite{reimers-2020-multilingual-sentence-bert}, and cluster the data according to the embeddings. %
We upweight a subset of roughly 300B tokens from the English dataset. Given a small target set (typically on the order of 1000-10000 samples), we assign each sample to a cluster and upweight the original training set based on the cluster assignment proportions.  In practice, we do not optimize for the cluster parameters jointly with the model weights, and instead precompute them based on the pretraining and target task data.

For clustering hyperparameters, we use a lightweight SentenceTransformers multilingual model\footnote{The particular model is called {\tt paraphrase-multilingual-MiniLM-L12-v2} model and is obtained from \url{https://huggingface.co/sentence-transformers/paraphrase-multilingual-MiniLM-L12-v2}.} for extracting features \cite{reimers-2019-sentence-bert}, and a balanced $K$-means algorithm to cluster the embeddings into 64 clusters.  

Upsampling is done in English at the cluster level, to facilitate having specialized information in the auxiliary language that is unavailable in the target language. We compare two settings:
\begin{enumerate*}[label=(\roman*)]
    \item upsampling based on HellaSwag (general knowledge and instruction style), and
    \item upsampling based on ARC (general science knowledge).
\end{enumerate*}
We also add a comparison where we downsample relevant data to the target task, to compare with having only low quality data, which we call ``no ARC DE (L)''.

\begin{figure}[t!]
    \centering
    \includegraphics[width=\columnwidth]{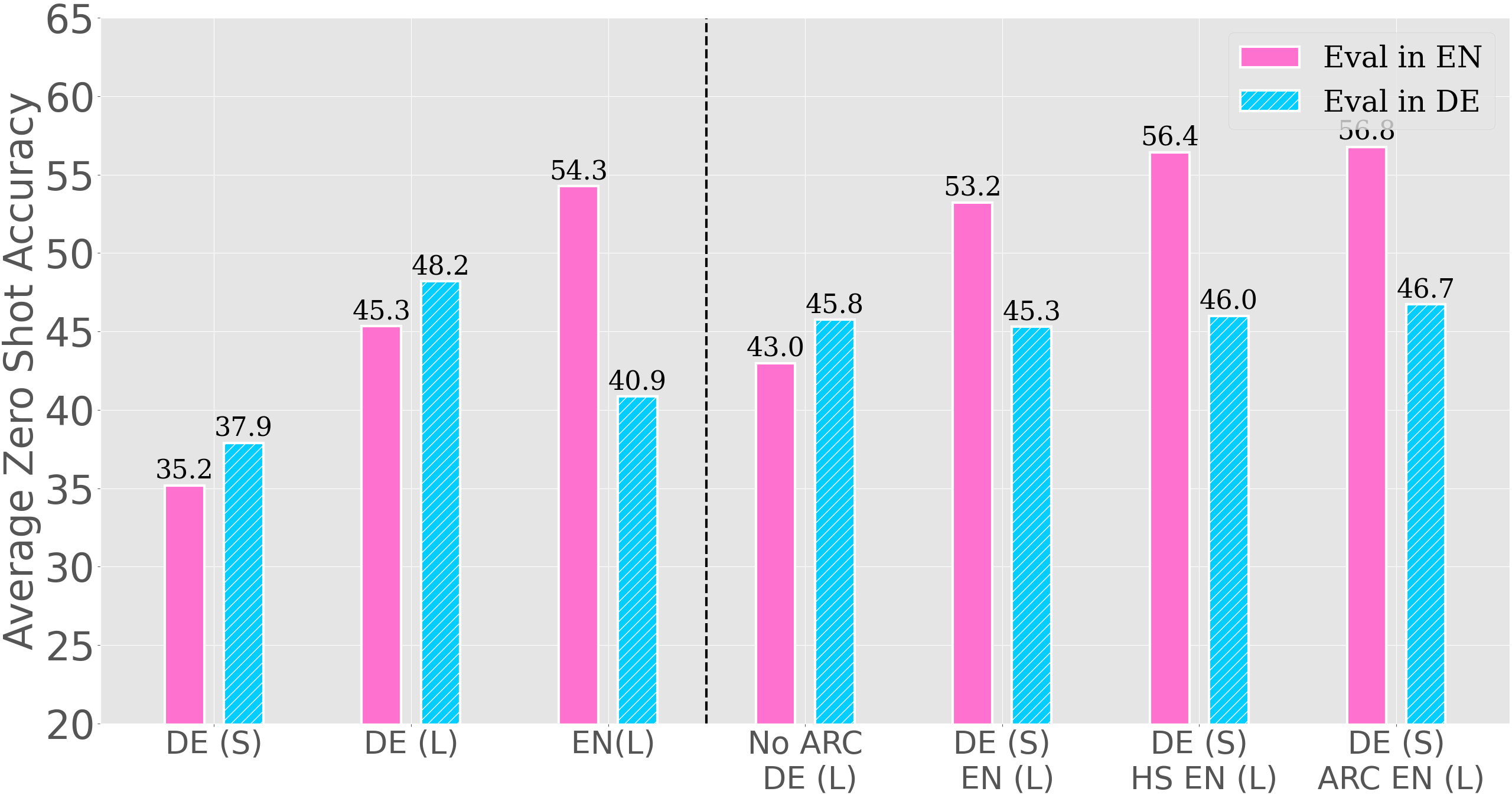}
    \caption{Zero-shot accuracy of models trained with  upsampling of English auxiliary data.  Results are averaged over six evaluation datasets. For each setting evaluation is done in English and German.}
    \label{fig:data_selection}
    \vspace{-2mm}
\end{figure}

\paragraph{Findings.}
Figure~\ref{fig:data_selection} shows the results. We see 4\% improvement in English evaluations, and 2\% improvement in the target language. This highlights that models can take advantage of information in the auxiliary language, and the performance improvements are higher for the information upsampling than for model-based filtering. %

\begin{figure}
    \centering
    \includegraphics[width=\columnwidth]{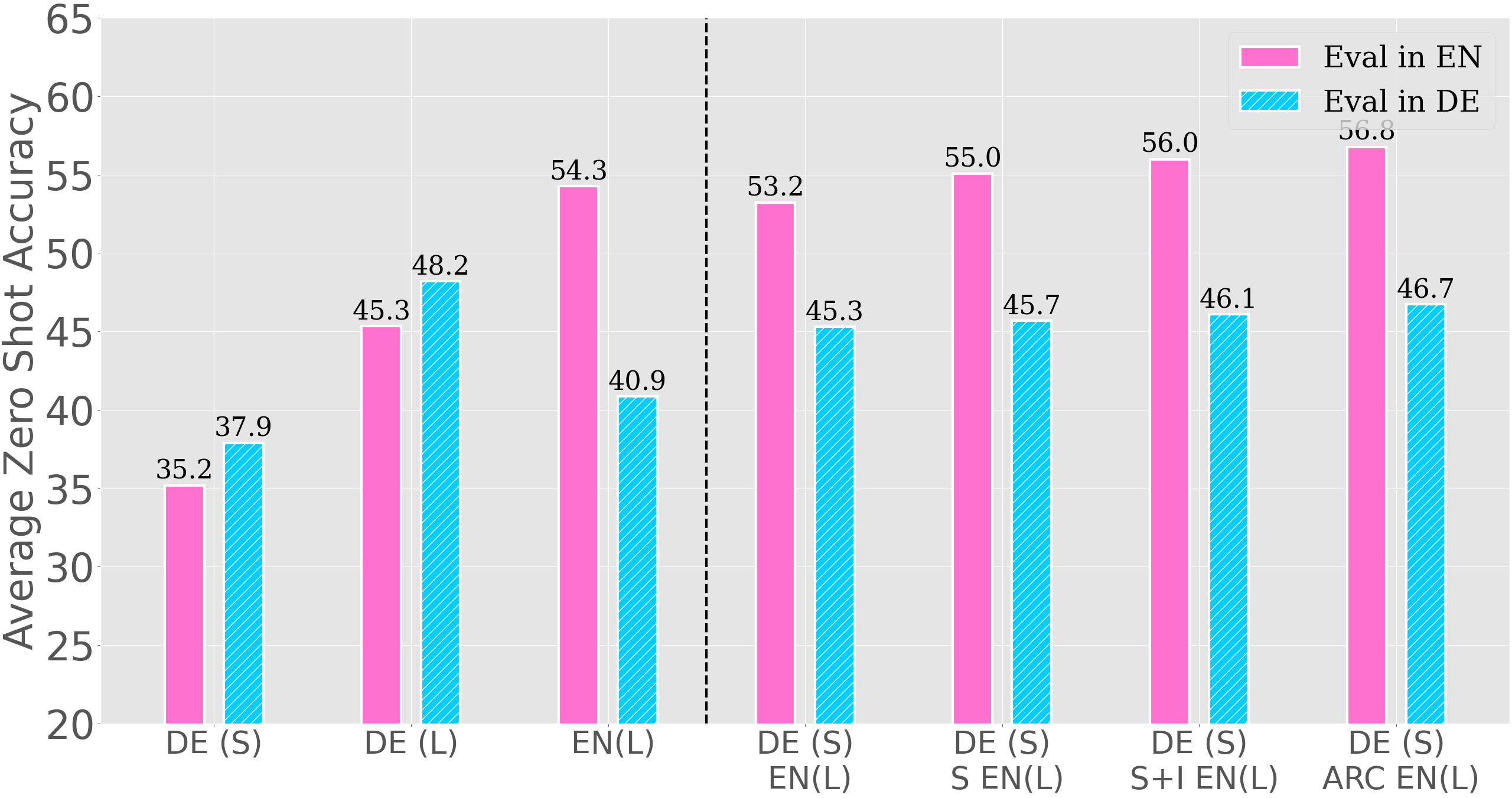}
    \caption{Zero-shot accuracy of models trained with  synthetic data upsampling of English auxiliary data.  Results are averaged over six evaluation datasets. For each setting evaluation is done in English and German.}
    \label{fig:syn_selection}
    \vspace{-2mm}
\end{figure}

Generally, our findings indicate that filtering for high quality data has limited improvement for the target language (0.8\% averaged across all languages), and further improvements come from data selection over important topics.

\subsection{Upsampling with Synthetic Examples} 
\label{sec:synthetic_information}

\paragraph{Motivation.}
While we achieved performance improvements by selecting data based on target downstream evaluations, having such data available at pretraining can be restrictive.  For this reason, it can be desirable to be able to \textit{generate} the necessary data for upsampling. While prior work has examined the use of LLM-generated data for pretraining \cite{maini2024rephrasing} and finetuning \cite{li2023self,yuan2024self} language models, to our knowledge there is no prior work that investigates data selection based on synthetic examples.

\paragraph{Methodology}
We generate a small set of synthetic examples, following the approach in \cite{maini2024rephrasing}. The synthetic data is created by prompting an off-the-shelf instruction finetuned language model to generate sets of questions relating to the topic of interest. We use the prompts in Appendix~\ref{sec:app_syn}. Generating synthetic data using an off-the-shelf language model can be both computationally expensive and challenging. However, for the purpose of computing sampling weights, a small number of questions is sufficient.  

For our experiments, we generate data using a frozen Mistral-7B instruction tuned model\footnote{\url{https://huggingface.co/mistralai/Mistral-7B-Instruct-v0.3}} \cite{jiang2023mistral}. We create around 2000 science QA pairshile with filteog that cnfirms only a question and answer are specified. We refer to this setting as ``S EN (L)'', S referring to Science.

We also generate general instruction data, aimed at broad QA style data, and general factual information helpful for downstream tasks.  This data is generated in two stages: we use the frozen Mistral-7B model to generate a set of questions, and then the model is prompted to answer the questions. 

Using the sets of questions, we identify the cluster sample weights and upsample data accordingly as in \S\ref{sec:is_information}. We refer to this setting as ``S+I EN (L)''.  %

\paragraph{Findings}
We show results in Figure~\ref{fig:syn_selection}. Comparing synthetic data upsampling with upsampling from the downstream tasks (Figure~\ref{fig:data_selection}) demonstrates that synthetic data can be sufficient for incorporating information into the auxiliary English dataset. Upsampling synthetic data is within 1\% for the English downstream tasks, and within 0.5\% for the translated German evaluations. Results for transformations in other languages are reported in \S\ref{sec:app_multilang_avg}.

\subsection{Translation Systems} 
\label{sec:translation}

\paragraph{Motivation.} Training directly on auxiliary language data can lead to improvements. An alternative strategy is to translate the auxiliary data into the target language, assuming a machine translation system is available. This approach offers the benefit of training the model exclusively in one language, and, if the translation system is of high quality, it allows for training on high-quality data in the target language at the expense of translating the corpus. It is, therefore, important to also assess the level the translation system must possess to effectively translate data for pretraining.  

\paragraph{Methodology.}
For our experiments, we use light-weight translation systems of roughly 100-200M parameters.  We consider three models with BLEU scores 16.0, 26.5, and 31.6 on the WMT-17 EN-DE benchmark task.  We denote these models as v1, v2, and v3, corresponding to increasing BLEU score. All models are trained on translated versions of the mC4 English corpus.  No other English data is included in the dataset, but we keep the 250M tokens of real German data as in prior experiments. 

\begin{figure}
    \centering
     \includegraphics[width=\columnwidth]{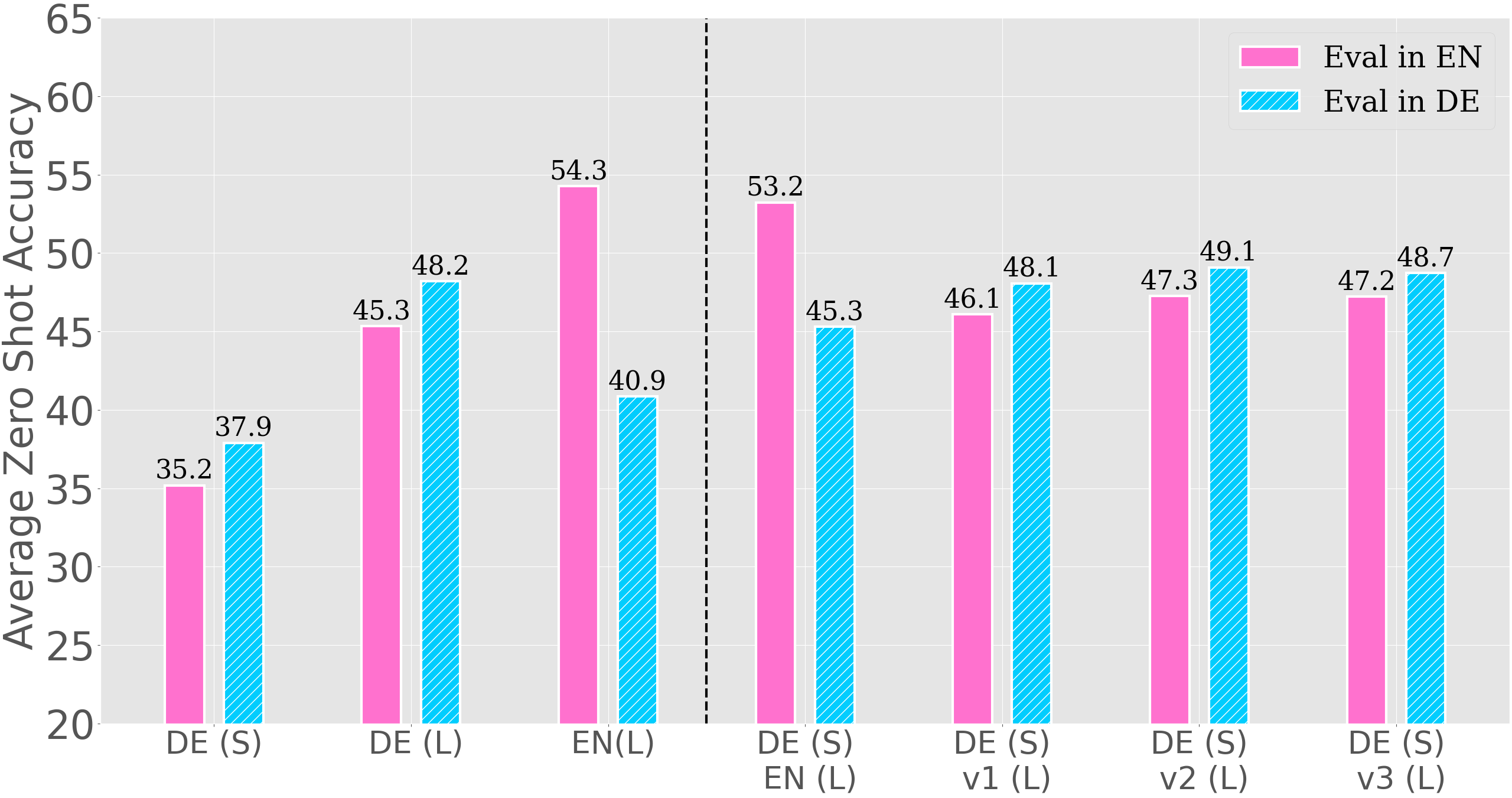}
    \caption{Zero-shot accuracy of models trained with  translations of English auxiliary data.  Results are averaged over six evaluation datasets. For each setting evaluation is done in English and German.}
    \label{fig:translation}
    \vspace{-2mm}
\end{figure} 

\paragraph{Findings.}
Figure~\ref{fig:translation} shows the results. We find little difference from the quality of the translation. Both stronger translation models achieve similar performance. The v1 model performs comparably to training on real German data, and the data translated by the v2 and v3 models obtain around 1\% improvement in comparison to real German data.
 
 We hypothesize a few reasons this may be possible, but leave investigation of each of these to future work: 
 \begin{enumerate*}[label=(\arabic*)]
     \item  English CC data is of higher quality than German CC data, as there may be more data from more diverse sources.
     \item Translated German data has a different distribution from real German data, and this better matches the translated test evaluations.
     \item Translated data from a small translation system might simplify language, which makes it easier for models to learn, following \cite{eldan2023tinystories}.
     \item Portions of the dataset could not be translated by the systems and are removed. These portions might be noisy, and some unintended filtering may lead to slightly higher performance.
 \end{enumerate*}

\section{Data and Model Scaling}
To study the impact of data and model scaling, we investigate two scaling questions:
 \begin{enumerate*}[label=(\arabic*)]
    \item How does the required quantity of additional auxiliary data relate to the amount of data required if one had enough target language data to train on? (\S\ref{sec:aux_data_scaling})
    \item How do findings extend to a larger bilingual model? (\S\ref{sec:model_size_scaling})
 \end{enumerate*}
 Following our setup in the previous section, we target our investigation to German (target) and English (auxiliary).

\subsection{Auxiliary Data Scaling}
\label{sec:aux_data_scaling}

\paragraph{Motivation.} 
We measure the amount of data needed in the \textit{target language} to match training on high quality auxiliary data. Our goal is to quantify the advantage of training models with additional target data, beyond the 250M tokens used before.

\paragraph{Methodology.}
We investigate data scaling over mC4 and FineWebEDU. %
We train models at different dataset sizes, from 0.1B to 100B tokens.

\begin{figure}
    \centering
     \includegraphics[width=\columnwidth]{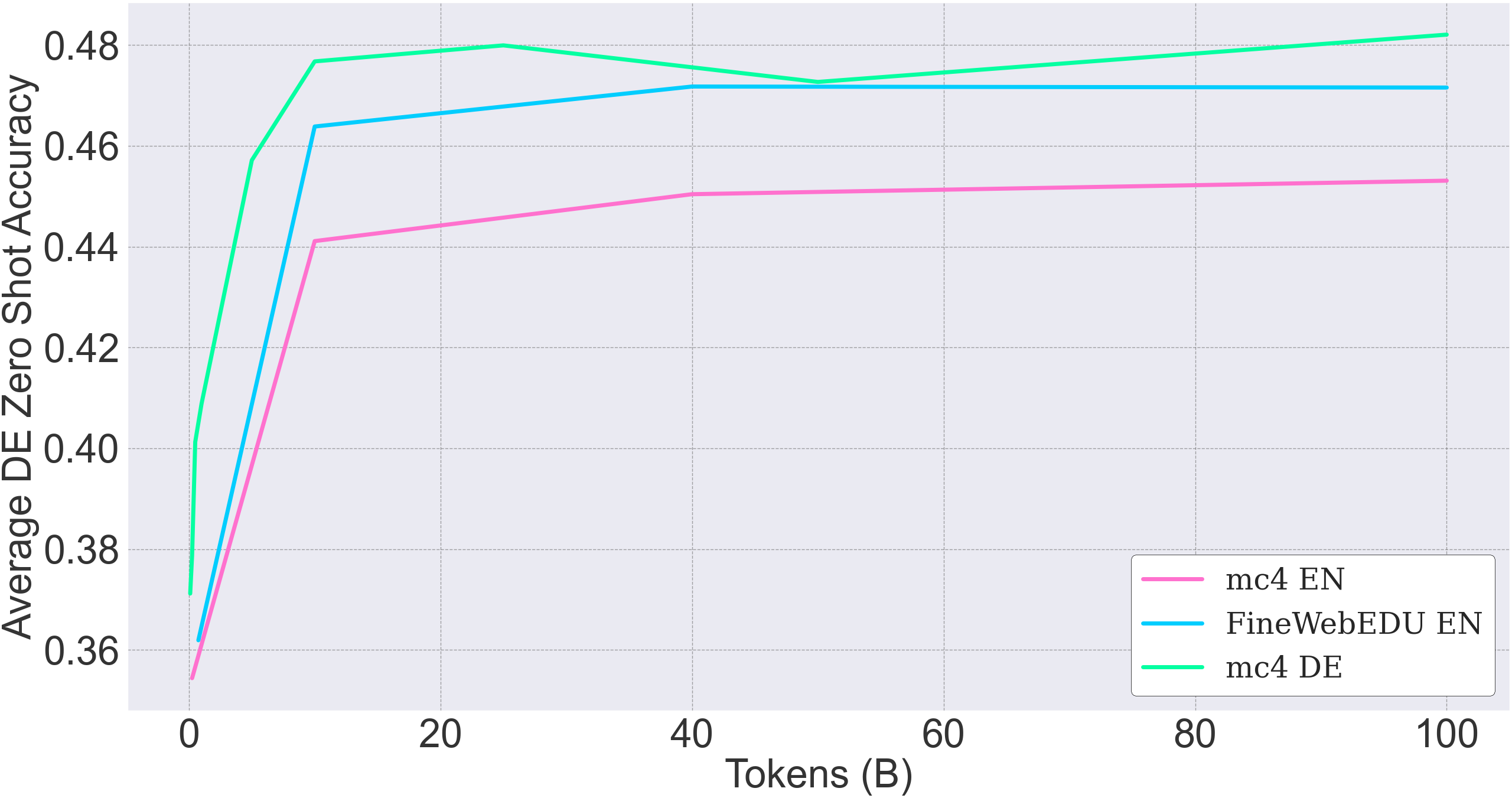}
    \caption{Average accuracy over zero-shot benchmark tasks in translated German, with increasing number of tokens in both target and auxiliary languages. Models are trained for 100B tokens.}
    \label{fig:data_scaling}
    \vspace{-2mm}
\end{figure}

\paragraph{Findings.}

Our results are summarized in Figure~\ref{fig:data_scaling}. Models trained on the English mC4 data achieve similar performance to models trained with only 5B tokens of German mC4 data. In contrast, training on a high quality data such as FineWebEDU increases the amount of German data needed to around 10B tokens, 2x the amount of German data. %
Note that the curves all plateau quickly at around 10B tokens, corresponding to around 10 repetitions of the data, matching  \cite{muennighoff2024scaling}.

There are two important data scaling considerations from Figure~\ref{fig:data_scaling} that justify training with auxiliary language data. First, for languages that are data constrained, it may be infeasible to collect twice as much data.  Second, models trained on FineWebEDU attain similar performance at a rate of 5x the number of tokens (roughly 50B tokens of FineWebEDU matches the performance of 10B tokens of mC4 German data).  An important avenue is to investigate data scaling at larger quantities of tokens. In particular, the FineWebEDU corpus totals 5.4T tokens and would require access to 1T tokens of German data, which is 3x the amount of data in mC4. As a result, while the data scaling shows large improvements from little German data, the large amount of readily available English data can make training on auxiliary data practical.

\subsection{Model Size Data Scaling}
\label{sec:model_size_scaling}

\begin{figure}
    \centering
    \includegraphics[width=\linewidth]{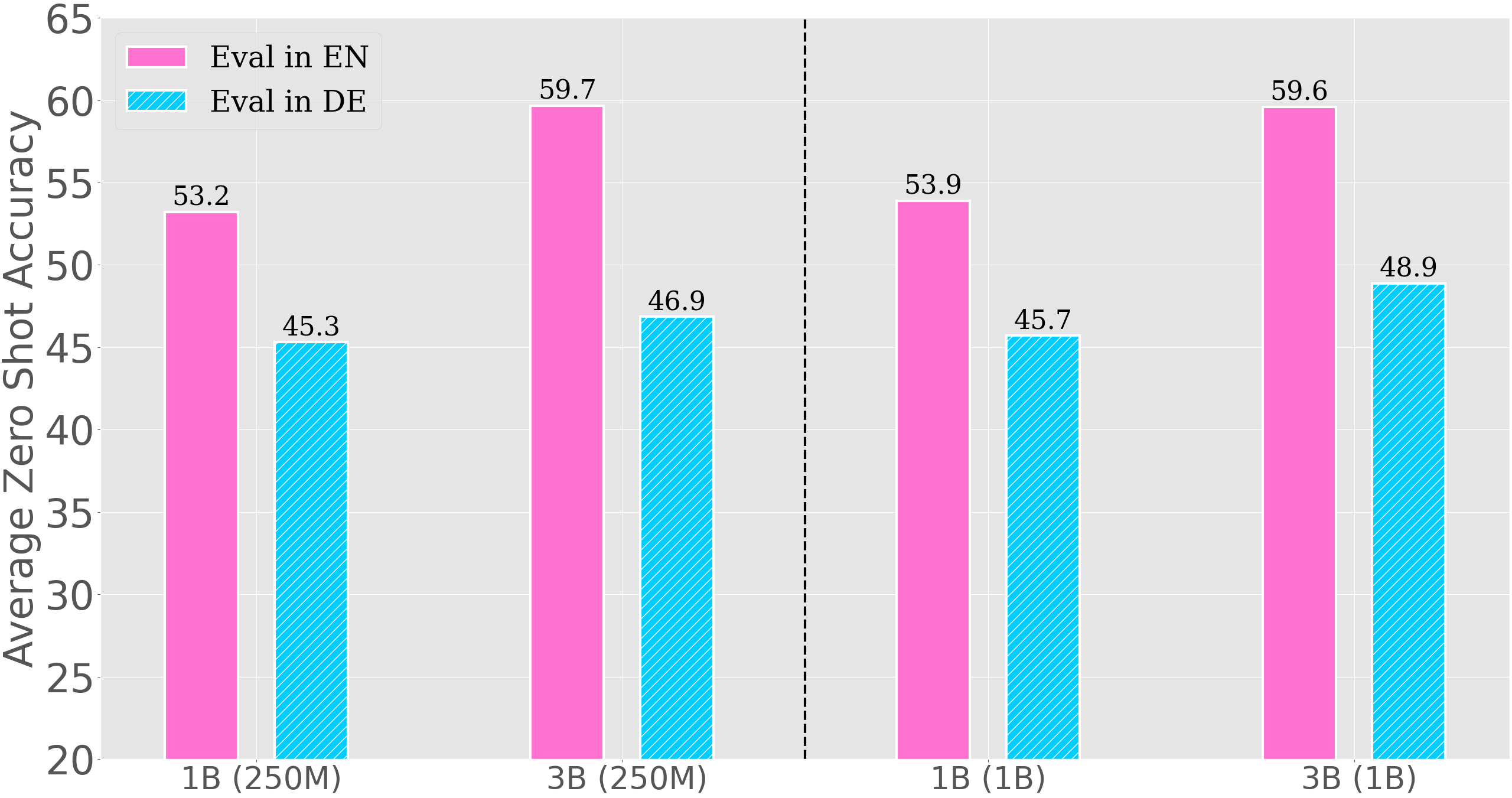}
    \caption{Zero-shot accuracy for 1B and 3B models trained with 250M  or 1B tokens in the target language denoted in brackets. Models are trained on the target language data for different ratios of the training steps . See \S\ref{sec:model_size_scaling_app} for details on data ratios.}
    \label{fig:1B_3B}
    \vspace{-2mm}
\end{figure}

Next, we investigate to what extent results scale for training larger language models when data remains constrained in the target language. Given a fixed amount of data from the target language (250M tokens) training a larger model may be impractical, as, according to Chinchilla scaling laws \cite{kaplan2020scaling}, increasing model size necessitates increasing the number of tokens seen during training.  This is challenging for data constrained languages, as the number of repetitions increases with increased amount of training. This can lead to overfitting and saturated model performance. 

 We train a  $\sim$2.7B parameter  model for 150K steps, matching the same scaling ratio as for the 1.3B model. We train models following the DE (S) + EN (L) setup. Our findings are in Figure~\ref{fig:1B_3B}. There is a limit to the size of models that can be trained with limited target data. Model performance for 250M tokens (more than available in mC4 for  $\sim$20 languages) shows limited improvement when scaling up from a 1.3B to a 2.7B parameter model. In contrast, increasing to 1B tokens doubles improvement when increasing model size to a 2.7B parameter model. Training a 2.7B parameter model necessitates training on at least 1B tokens of data to see larger gains in performance (more than available in mC4 for $\sim$40 languages).

\section{Conclusion}
This work studies how an auxiliary language for which an abundance of training data is available can boost pretraining for a target language for which only limited data is available. We find that adding auxiliary high quality data obtained by data filtering can improve performance in a target language. Moreover, our results indicate that most of these gains can be attributed to having relevant information in the auxiliary language, which may not be present in the target language. However, we find that results are inconsistent across target languages. We hypothesize that for languages further from English, better English datasets are not as helpful as information is not shared between them. Finally, we find limitations to scaling models for languages that are data constrained.  This work takes a step towards pretraining language models in languages with limited data, and can inspire more research into bilingual  learning under data constraints. %

\section{Limitations}
\label{sec:limit}

In this section we list some limitations of our work.

\paragraph{Evaluation data.}

When evaluating language models for languages other than English, one often needs to rely on translated evaluation sets. Not only may this introduce translation mistakes, the resulting evaluation set might also contain cultural biases. As a result, certain aspects of the evaluation may lead to improved performance when using English auxiliary or translated data. Additionally, translated data often exhibits a distribution different from that of real data in the target languages. Therefore, an important direction for future work is the development of evaluation datasets that are not based on translation, which is essential for more accurate evaluation of bi- and multilingual language models.

\paragraph{Languages included.}
The focus of this work is on training language models for data constrained langauges.  We note that there are many languages within mC4 (and more broadly) which can benefit from having auxiliary English data for pretraining.  Due to the aforementioned limited evaluation benchmarks and availability of target language data for comparison, we leave investigation for these languages to future work.

\section*{Acknowledgements}

We are grateful to Masha Fedzechkina Donaldson, Kunal Talwar, and Maureen de Seyssel for their helpful discussions, comments, and thoughtful feedback in reviewing this work.

\bibliography{main.bib}

\clearpage 
\appendix

\section{Hyperparameters and Training Details}
\label{sec:hyperparams}

The medium-scale (300M non-embedding parameter) model consists of 24 layers, 16 attention heads, and a hidden dimension size of 1024. The XL-scale (1.3B non-embedding parameter) model consists of 24 layers, 16 attention heads, and a hidden dimension size of 2048. Both models have a maximum sequence length of 1024. 

The baseline models are trained using NVIDIA’s Megatron-LM\footnote{\url{https://github.com/NVIDIA/Megatron-LM}} repository for pretraining language models. The medium size models are trained for a total of 30K steps, and 100K steps for the XL models at a batch size of 1024.  All models are trained using a maximum learning rate of $0.0003$ for the medium  model and $0.0002$ for the XL model, and a minimum learning rate of $0.00001$ with a cosine learning rate scheduler and warmup for $1\%$ of the total steps. For regularization, we use a weight decay of $0.01$, along with a gradient clipping norm of $1.0$. Models are trained with the Adam optimizer using $\beta_1=0.9$ and $\beta_2=0.999$.

The 2.7B parameter model consists of 32 layers with 2560 hidden dimension and 32 attention heads. 2.7B parameter models are trained for a total of 150K steps with a batch size of 1024 and context length of 2048.  All models are trained using a learning rate of $0.00016$ with the same optimization recipe as other models.  

The total training time for XL models on roughly 100B tokens is around 1000 GPUh on Nvidia H100 GPUs.  For medium size models, the total training time is around 200 hours for roughly 30B tokens.

\section{Dataset Details}
\label{sec:app_dataset}

\subsection{Train Sets}

\begin{itemize}[leftmargin=*]
    \item \textbf{mC4}: The primary pretraining corpus in our experiments is multilingual Colossal Clean Crawled Corpus (mC4), a curated text dataset comprising over 6.3T tokens. This corpus is derived from CommonCrawl and used for pretraining numerous language models~\cite{brown2020language,raffel2020exploring,touvron2023llama}. The dataset is chosen as all languages have similar data extraction pipelines including line length filter, {\tt cld3} language detection, and deduplication~\cite{xue2020mt5}.  The English portion contains 2.7T tokens, German contains 350B tokens, French contains 320B tokens, Spanish contains 430B tokens, Portuguese contains 146B tokens, Italian contains 160B tokens, Korean contains 26B tokens, Japanese contains 160B tokens, and Chinese contains 40B tokens. For both Chinese and Korean monolingual models: Target (L), we repeat data up to 100K steps of training (roughly 2-3 repetitions).
    \item \textbf{RedPajamav2}: A pretraining corpus with light filtering (primarily only deduplication) comprising 30T tokens and 20T tokens of English text.  We focus on the English portion of the dataset only and train using a random shuffled subset of both the head and middle portions~\cite{together2023redpajama}. 
    \item \textbf{RefinedWeb}: The dataset is also derived from the CommonCrawl, however has a more stringent filtering process including {\tt trafilatura} text extraction, document and line level rules, and fuzzy duplication over the original C4 processing~\cite{penedo2023refinedweb}.
    \item \textbf{FineWeb}: This dataset is derived from the CommonCrawl with the aim of replicating RefinedWeb at larger scales.  The dataset has some additional filtering including Gopher filtering \cite{rae2021scaling}, additional C4 filters, and custom filters for text quality ~\cite{penedo2024fineweb}.
    \item \textbf{FineWebEDU}: A subset of the FineWeb dataset which is filtered according to a classifier trained on annotations for educational quality from Llama-3 70B model~\cite{penedo2024fineweb}.
    \item \textbf{ChineseFineWeb-EDU}\footnote{\url{https://huggingface.co/datasets/opencsg/chinese-fineweb-edu-v2}}: An educational corpus in Chinese consisting of roughly 400B tokens of data.  Although it shares a similar name, the ChineseFineWeb-EDU does not share data from FineWebEDU and is collected from different sources.  Models trained on this data are trained with only 1 repetition.
\end{itemize}

\subsection{Zero Shot Evaluations}
\label{app:eval_desc}
\begin{itemize}[leftmargin=*]
    \item \textbf{SciQ}: A dataset of science exam questions, specifically designed to evaluate the ability of NLP models in understanding and reasoning within the scientific domain~\citep{welbl2017crowdsourcing}.
    \item \textbf{ARC Challenge (ARC-C)}: This dataset is part of the AI2 Reasoning Challenge (ARC)~\citep{clark2018think}, containing science exam questions from grades 3 to 9. The ARC Challenge set includes more difficult questions that necessitate higher-order reasoning.
    \item \textbf{ARC Easy (ARC-E)}: The Easy set of the AI2 Reasoning Challenge~\citep{clark2018think} features questions from the same source as ARC-C but are considered less challenging and do not require as advanced reasoning skills.
    \item \textbf{Winogrande (Wino.)}: This dataset challenges models on common sense reasoning in a language context, focusing on pronoun disambiguation tasks~\citep{sakaguchi2021winogrande}.
    \item \textbf{PIQA}: Physical Interaction Question Answering tests the understanding of everyday physical processes, an aspect of practical common sense~\citep{bisk2020piqa}.
\item \textbf{HellaSwag}: This dataset evaluates a model's ability to complete scenarios in a contextually and logically coherent manner, requiring both language understanding and common sense reasoning~\citep{zellers2019hellaswag}.
\end{itemize}

For each of the eval datasets, we include the number of samples for each translated evaluation in Table~\ref{tab:lm_eval_data_size}.  For our evaluations, we use the lm-eval-harness repository\footnote{\url{https://github.com/EleutherAI/lm-evaluation-harness}} for zero-shot accuracy on QA tasks.

We translate primarily using our own systems to 
\begin{enumerate*}[label=(\roman*)]
 \item ensure that translation artifacts will be consistent across evaluation tasks,
\item have translations for all languages we evaluate.
\end{enumerate*}
We note further that evaluation frameworks such as Okapi \cite{dac2023okapi} do not release evaluation for all languages we study, and only support HellaSwag as they do not translate the full ARC evaluation set. We did not evaluate on other multilingual evaluation tasks such as MT and MGSM as these are not traditionally knowledge tasks that require information sharing between the two languages.

\begin{table*}[h!]
    \centering
    \scalebox{0.8}{
        \begin{tabular}{l|rrrrrrrrr}
            \toprule
           \textbf{Dataset} & \textbf{EN} & \textbf{DE} & \textbf{FR} & \textbf{ZH} & \textbf{JA} & \textbf{PT} & \textbf{ES} & \textbf{IT} & \textbf{KO} \\
            \midrule
            ARC-C & 1,172 & 1,137 & 1,147 & 1,146 & 1,147 & 1,117 & 1,117 & 1,117 & 1,147 \\
            ARC-E & 2,376 & 2,260 & 2,271 & 2,271 & 2,271 & 2,271 & 2,271 & 2,271 & 2,271 \\
            HS & 10,042 & 9,368 & 9,338 & 9,266 & 10,033 & 9,229 & 9,374 & 9,193 & 10,025 \\
            PIQA & 1,838 & 1,838 & 1,838 & 1,838 & 1,838 & 1,838 & 1,838 & 1,838 & 1,838 \\
            SCIQ & 1,000 & 950 & 953 & 1,000 & 926 & 953 & 951 & 952 & 945\\
            WG & 1,267 & 1,184 & 1,215 & 1,059 & 1,096 & 1,232 & 1,239 & 1,233 & 1,177\\
            \bottomrule
        \end{tabular}
    }
    \caption{Evaluation set sizes for each language.}
    \label{tab:lm_eval_data_size}
\end{table*}

\subsection{Number of Data Files for Filtering Experiments}

The mC4 English portion of the dataset is split into roughly 11,264 files totaling 2.7T tokens of data \cite{xue2020mt5}.  For each of our experiments, data is filtered differently, and as such varying numbers of files are needed for training.  At a baseline, we consider the first 1500 files totaling roughly 350B tokens of data.  This number was selected to match the total amount of German data which is recorded as 347B tokens using the mT5 tokenizer \cite{xue2020mt5}. For the OH classifier, we use the first 10,000 files and filter down to 10\% of the dataset.  For German, Japanese, Spanish, Portuguese, Italian, and French models, we use the first two files of data totaling roughly 250-300M tokens of data.  For Chinese and Korean models, we use the first 7 files totaling roughly 250M tokens.

\subsection{License and Attribution}
All datasets used in this paper are supported by public licenses including ODC and Apache.  The pre-trained models including Mistral and OH FastText classifiers are also supported by public licenses including Apache and MIT licenses. We use the Megatron codebase under the Nvidia license for pre-training and the lm-eval-harness (MIT) for evaluations.  All models and datasets are collected from Huggingface via the datasets library where possible.
We use a proprietary translation system for fast translation at scale and are thus unable to provide details of the license at this time.

\section{Evaluation Metrics}
\label{sec:app_eval}

The metric utilized for evaluation is the \textit{macro token level perplexity}. Given a batch of encoded texts, the perplexity at the token level was computed as follows:

Given the accumulated loss over the entire dataset, denoted as \( L \), and the total number of tokens, represented by \( T \), the macro token-level perplexity, denoted as \( \mathcal{P} \), is calculated as:

\begin{equation}
\mathcal{P} = \exp\left(\min\left(20, \frac{L}{T}\right)\right)
\end{equation}

Where:
\begin{itemize}[nosep]
    \item \( \exp \) is the exponential function.
    \item \( L \) is the cumulative loss over all shifted logits and labels in the dataset.
    \item \( T \) is the total number of tokens in the dataset.
\end{itemize}

The value of 20 acts as an upper limit to stabilize the metric in cases of high loss values.

For zero-shot MCQ accuracy evaluations, we compute the perplexity of each sentence completion, and choose the lowest perplexity choice.  We use the lm-evaluation-harness and where possible evaluate with the length-normalized accuracy.  Unless otherwise stated, all evaluations are zero-shot.

\section{Synthetic Prompts and Examples}
\label{sec:app_syn}

For building the synthetic corpus used in our data selection experiments, we consider three prompts for generating science questions (similar to many downstream tasks), fact-based QA data, and instruction-based writing (such as emails, books, lists, etc.).  For generating science questions, we generate both the question and answer.  For the fact and instruction data, we first generate the questions using the prompt, and subsequently generate the answer without any additional prompting.

\noindent {\textbf{Science Question Prompt}}
\begin{tcolorbox}[colback=gray!20, colframe=gray!75, rounded corners, sharp corners=northeast, sharp corners=southwest]
\texttt{Give me a set of ten question and answer pairs on topics relating to Physics, Chemistry and Biology that a high school student would be able to answer.  The response should be in the form Question: <question> \textbackslash n Answer: <answer>  \textbackslash n  \textbackslash n with an answer that is less than ten words.  The response should not contain any other details or explanations about the question or answer.}
\end{tcolorbox}

\noindent {\textbf{Facts Question Prompt}}
\begin{tcolorbox}[colback=gray!20, colframe=gray!75, rounded corners, sharp corners=northeast, sharp corners=southwest]
\texttt{People from different social and educational backgrounds, beliefs, ethnicity and gender are asking an AI assistant for information. They are looking for detailed explanations about encyclopedic facts on Wikipedia and in textbooks, about philosophy, nature, science, entertainment, literature, geography, socialogy, law, history, etc. Write an interesting and difficult question that would be sent to the AI assistant:}
\end{tcolorbox}

\noindent {\textbf{Instruction Writing Prompt}}
\begin{tcolorbox}[colback=gray!20, colframe=gray!75, rounded corners, sharp corners=northeast, sharp corners=southwest]
\texttt{People from different social and educational backgrounds, beliefs, ethnicity and gender are asking an AI assistant to help them write a piece of text that they need for their work or their personal life. They can ask the AI Assistant to write a document (email, letter, official document...). Each request comes with a long, precise and detailed description of what needs to be in the text, and why they need this document. The request may also include information about the writing style, the tone, the target audience or the layout of the text. The description of the task is formal, detailed and clear. Each request is composed of a few paragraphs written in English, and starts with the tag <request>. Here is some of the most interesting and original requests sent to the AI assistant: }
\end{tcolorbox}

\section{Data Distributions}
\label{sec:data_dists}
\begin{figure*}[h]
    \centering
       \begin{subfigure}{0.44\textwidth}
        \centering
        \includegraphics[width=\textwidth]{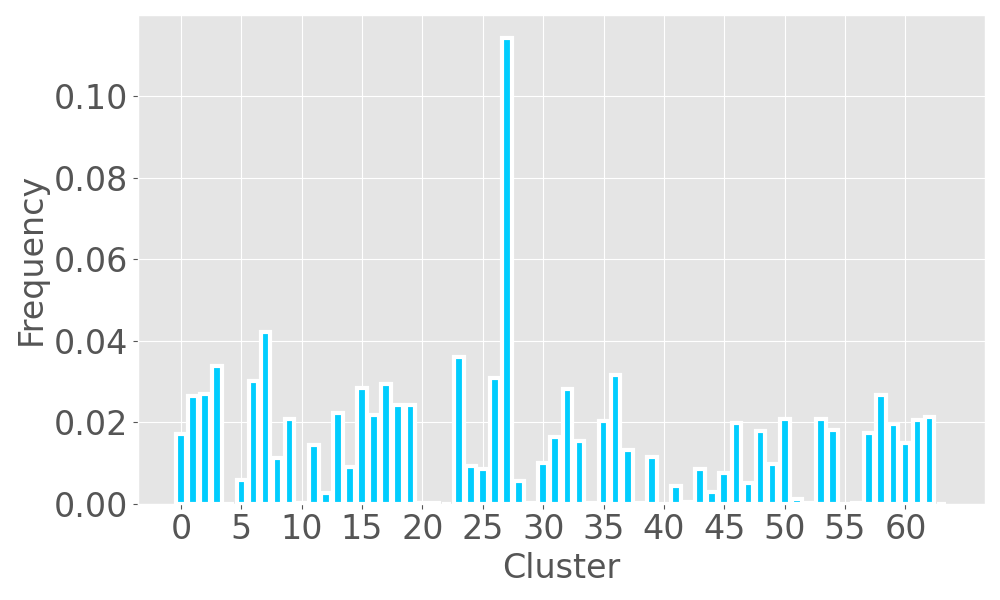}
        \vspace{-5mm}
        \caption{German}
        \label{fig:german_cluster_hist}
    \end{subfigure}
    \hfill
    \begin{subfigure}{0.44\textwidth}
        \centering
        \includegraphics[width=\textwidth]{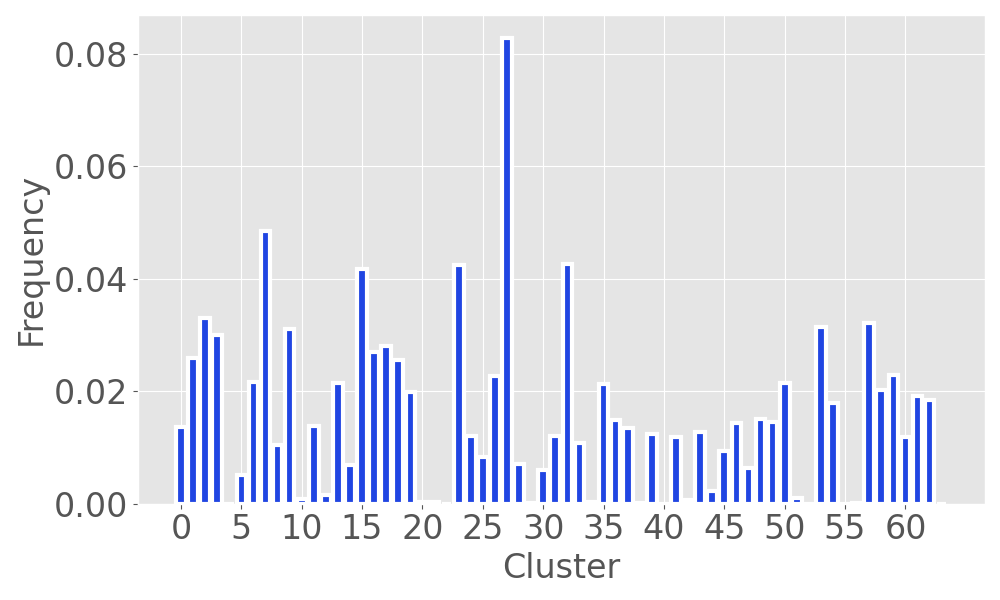}
        \vspace{-5mm}
        \caption{French}
        \label{fig:french_cluster_hist}
    \end{subfigure}
    \caption{Data distribution within mC4 English clusters for German and French mC4 data.}
    \label{fig:cluster_de_fr}
\end{figure*}

\begin{figure*}[h]
    \centering
       \begin{subfigure}{0.44\textwidth}
        \centering
        \includegraphics[width=\textwidth]{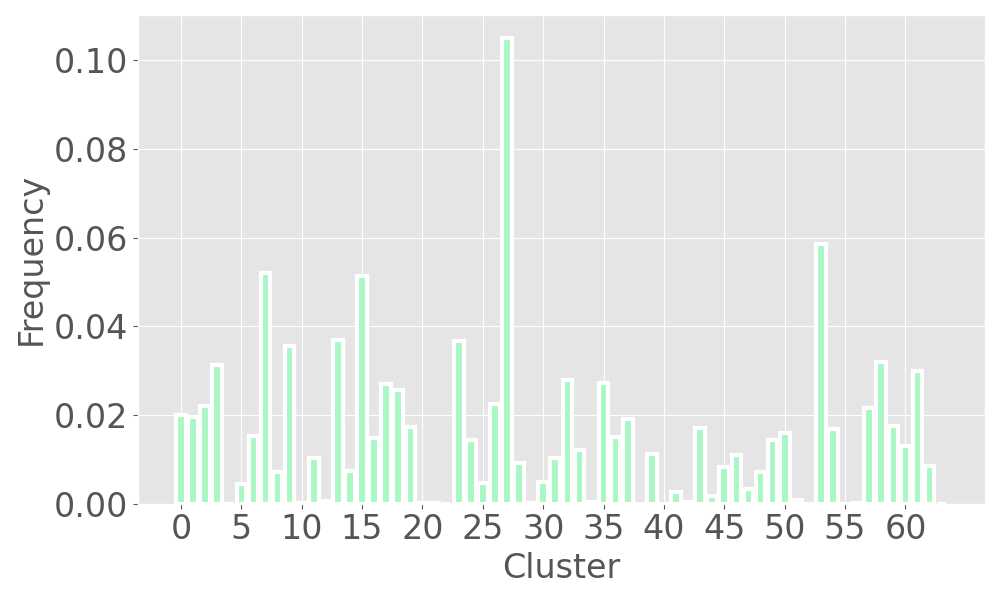}
        \vspace{-5mm}
        \caption{Spanish}
        \label{fig:spanish_cluster_hist}
    \end{subfigure}
    \hfill
    \begin{subfigure}{0.44\textwidth}
        \centering
        \includegraphics[width=\textwidth]{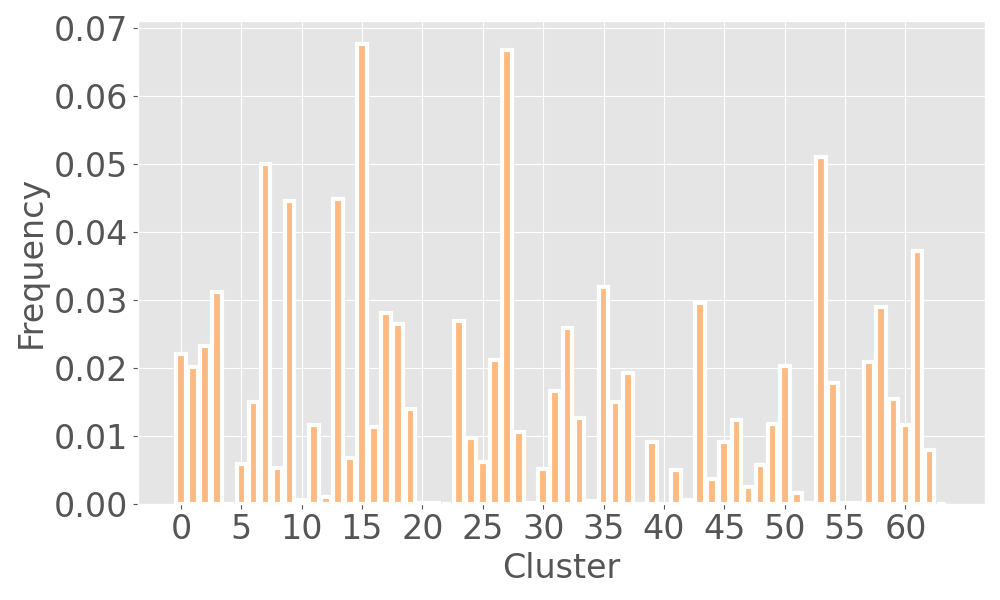}
        \vspace{-5mm}
        \caption{Portuguese}
        \label{fig:portuguese_cluster_hist}
    \end{subfigure}
    \caption{Data distribution within mC4 English clusters for Spanish and Portuguese mC4 data.}
    \label{fig:cluster_es_pt}
\end{figure*}

\begin{figure*}[h]
    \centering
       \begin{subfigure}{0.44\textwidth}
        \centering
        \includegraphics[width=\textwidth]{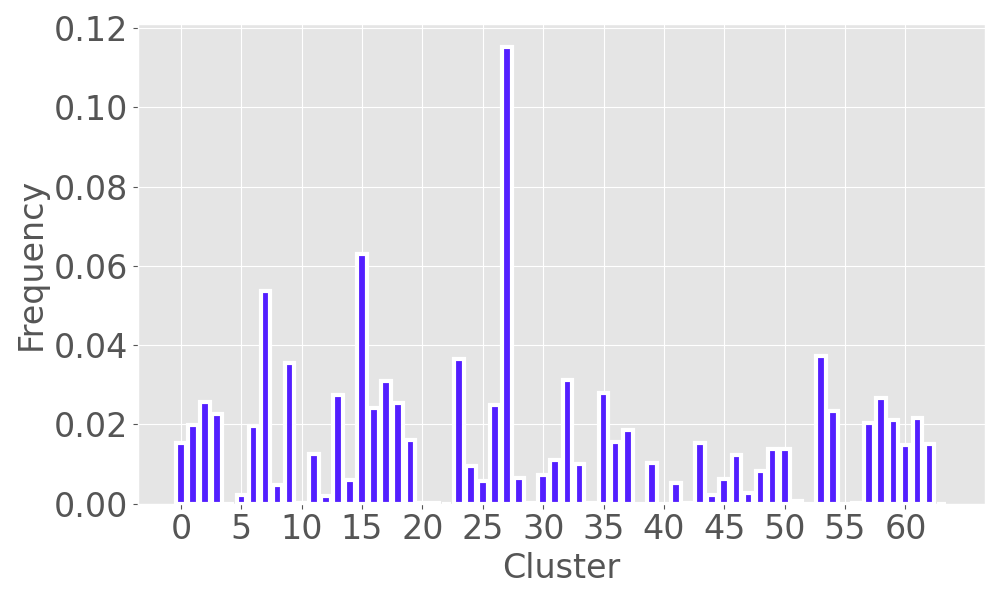}
        \vspace{-5mm}
        \caption{Italian}
        \label{fig:italian_cluster_hist}
    \end{subfigure}
    \hfill
    \begin{subfigure}{0.44\textwidth}
        \centering
        \includegraphics[width=\textwidth]{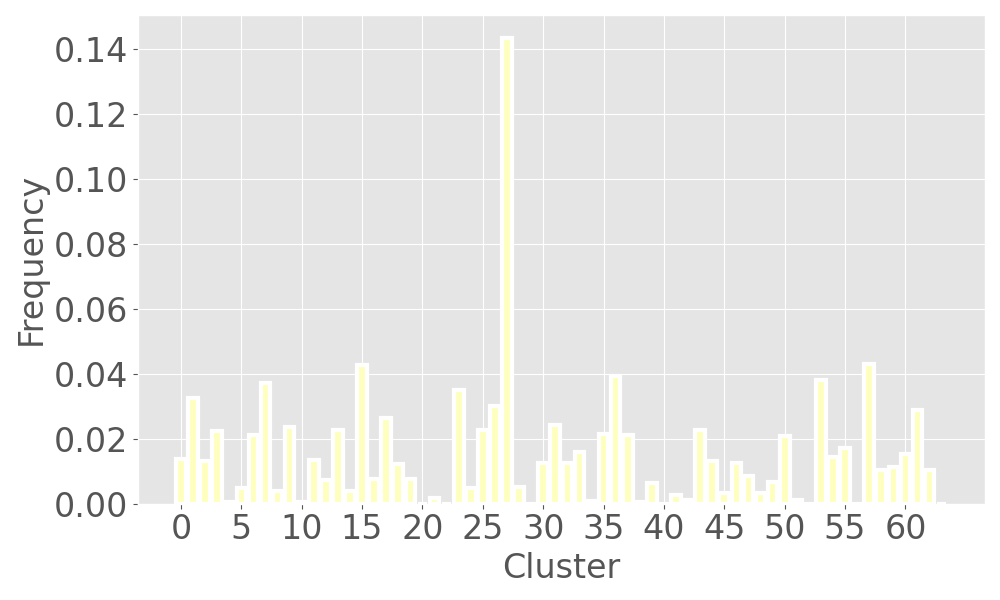}
        \vspace{-5mm}
        \caption{Korean}
        \label{fig:korean_cluster_hist}
    \end{subfigure}
    \caption{Data distribution within mC4 English clusters for Italian and Korean mC4 data.}
    \label{fig:cluster_it_ko}
\end{figure*}

\begin{figure*}[h]
    \centering
       \begin{subfigure}{0.44\textwidth}
        \centering
        \includegraphics[width=\textwidth]{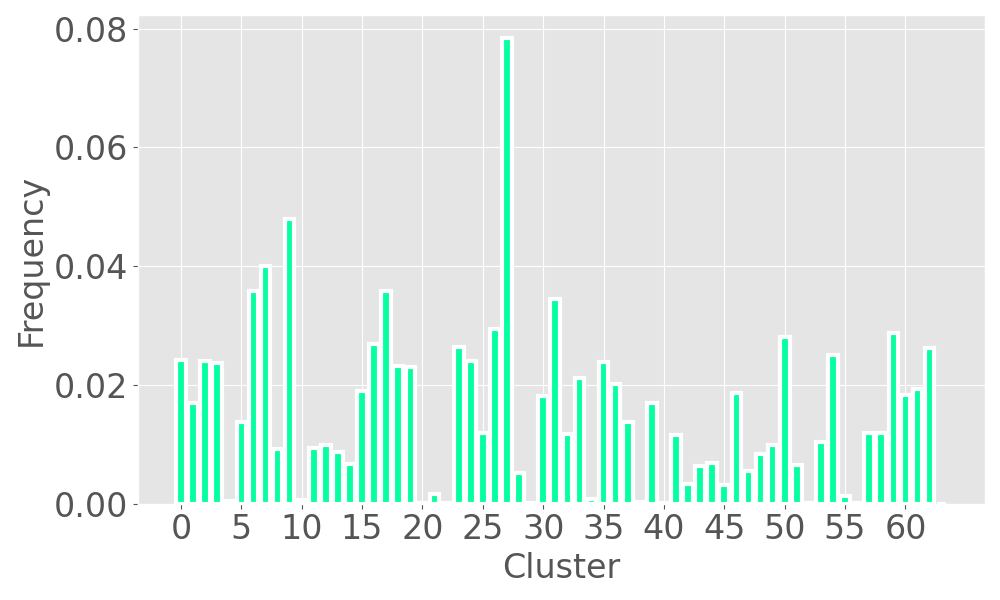}
        \vspace{-5mm}
        \caption{Japanese}
        \label{fig:japanese_cluster_hist}
    \end{subfigure}
    \hfill
    \begin{subfigure}{0.44\textwidth}
        \centering
        \includegraphics[width=\textwidth]{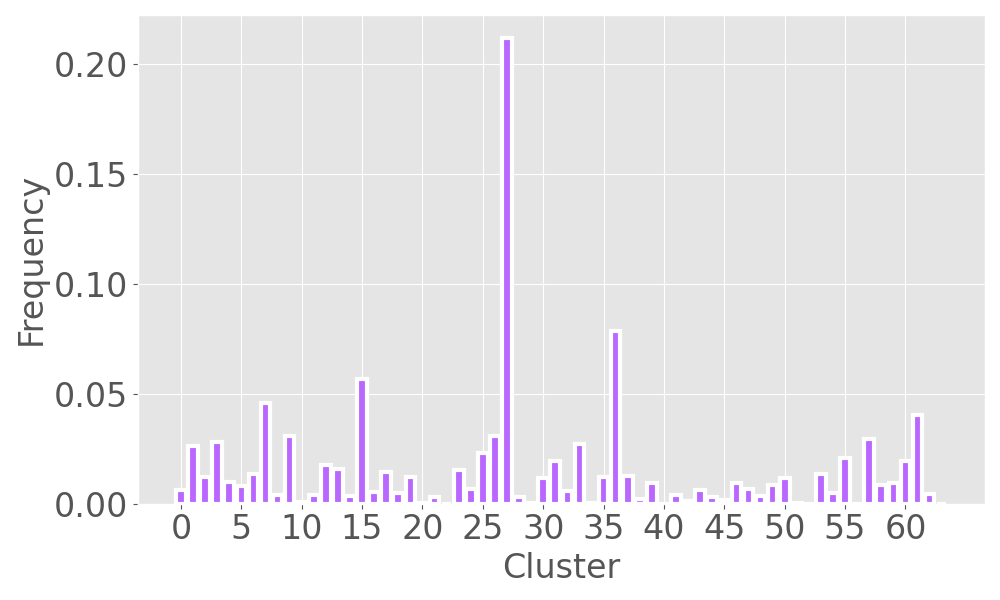}
        \vspace{-5mm}
        \caption{Chinese}
        \label{fig:chinese_cluster_hist}
    \end{subfigure}
    \caption{Data distribution within mC4 English clusters for Japanese and Chinese mC4 data.}
    \label{fig:cluster_ja_zh}
\end{figure*}

We present results comparing the data distributions for each dataset within the mC4 English clusters.  The purpose of these plots is to illustrate that the data from other languages follows a similar distribution to that of mC4 English. Figure~\ref{fig:cluster_de_fr} shows the distribution for French and German languages, Spanish and Portuguese in Figure~\ref{fig:cluster_es_pt}, Italian and Korean in Figure~\ref{fig:cluster_it_ko}, and for Japanese and Chinese languages in Figure~\ref{fig:cluster_ja_zh}. 

\section{Model Size Data Scaling}
\label{sec:model_size_scaling_app}

\begin{figure*}[htp!]
    \centering
       \begin{subfigure}{0.45\textwidth}
        \centering
        \includegraphics[width=\textwidth]{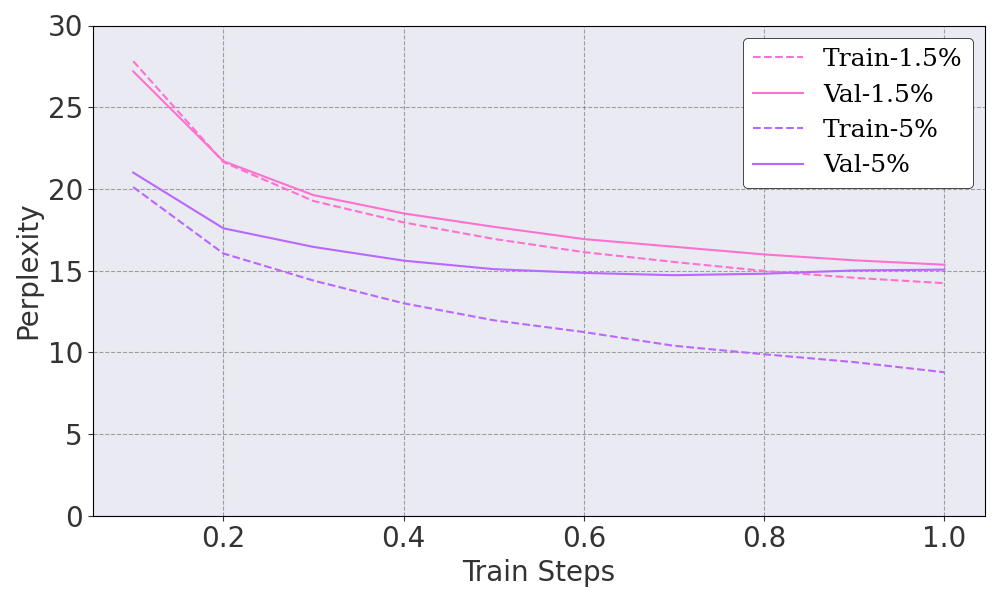}
        \caption{1B Model Perplexity}
        \label{fig:overfitting_1B}
    \end{subfigure}
    \hfill
    \begin{subfigure}{0.45\textwidth}
        \centering
        \includegraphics[width=\columnwidth]{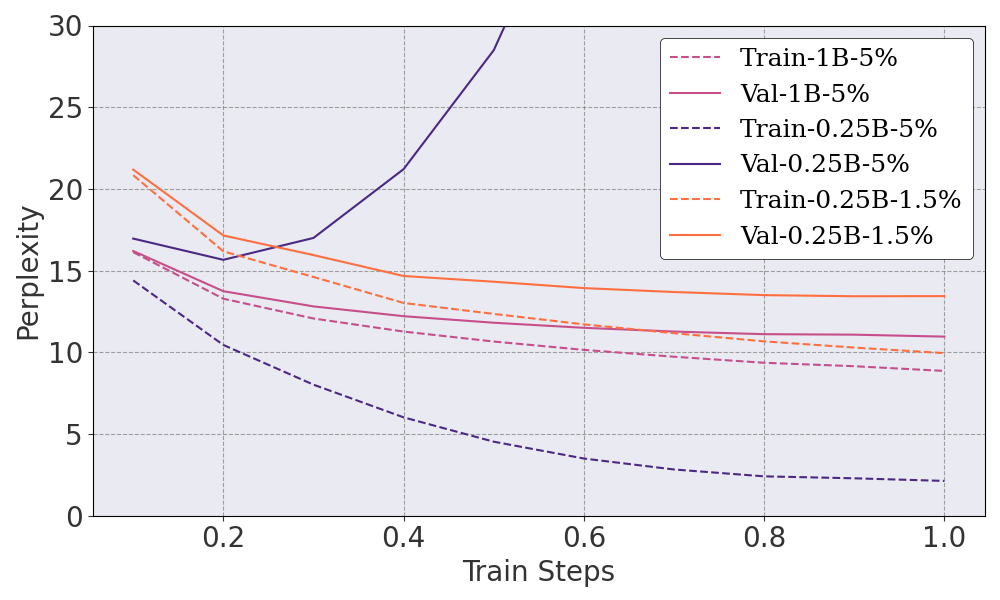}
        \caption{3B Model Perplexity}
        \label{fig:overfitting_3B}
    \end{subfigure}
    \caption{Train and validation perplexity for 1.3B and 2.7B parameters with varying amount of data.}
    \label{fig:perplexity_overfit}
\end{figure*}

In this section, we include additional findings accompanying the results presented in \S\ref{sec:model_size_scaling}.

 The perplexity of training and validation data is shown in Figure~\ref{fig:perplexity_overfit} for the 1.3B (left) and 2.7B models (right). For 1.3B models, we see little to no overfitting at 5\% ratio of training steps, and the model achieves slightly lower perplexity than a model trained with a lower ratio of 1.5\% target language data.  In contrast, for the 2.7B model, there is clear overfitting from as early as 25\% of the training with 5\% ratio of training steps.  This indicates that the number of repetitions is too high for the 2.7B model and performance may degrade. 

\begin{figure*}[htp!]
    \centering
       \begin{subfigure}{0.45\textwidth}
        \centering
         \includegraphics[width=\linewidth]{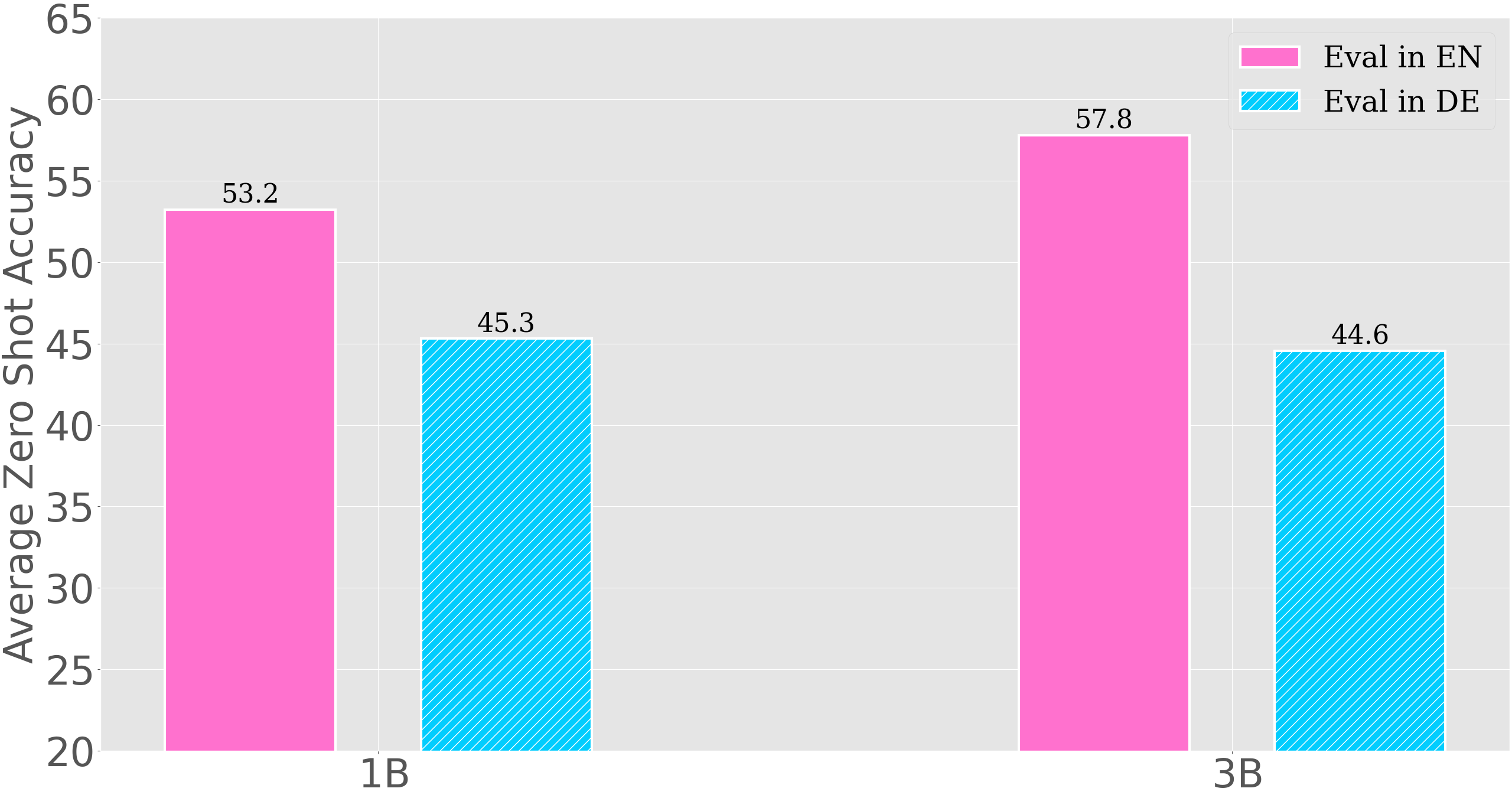}
        \caption{1B Model}
        \label{fig:barplot_1B_3B_250M_005}
    \end{subfigure}
    \hfill
    \begin{subfigure}{0.45\textwidth}
        \centering
         \includegraphics[width=\linewidth]{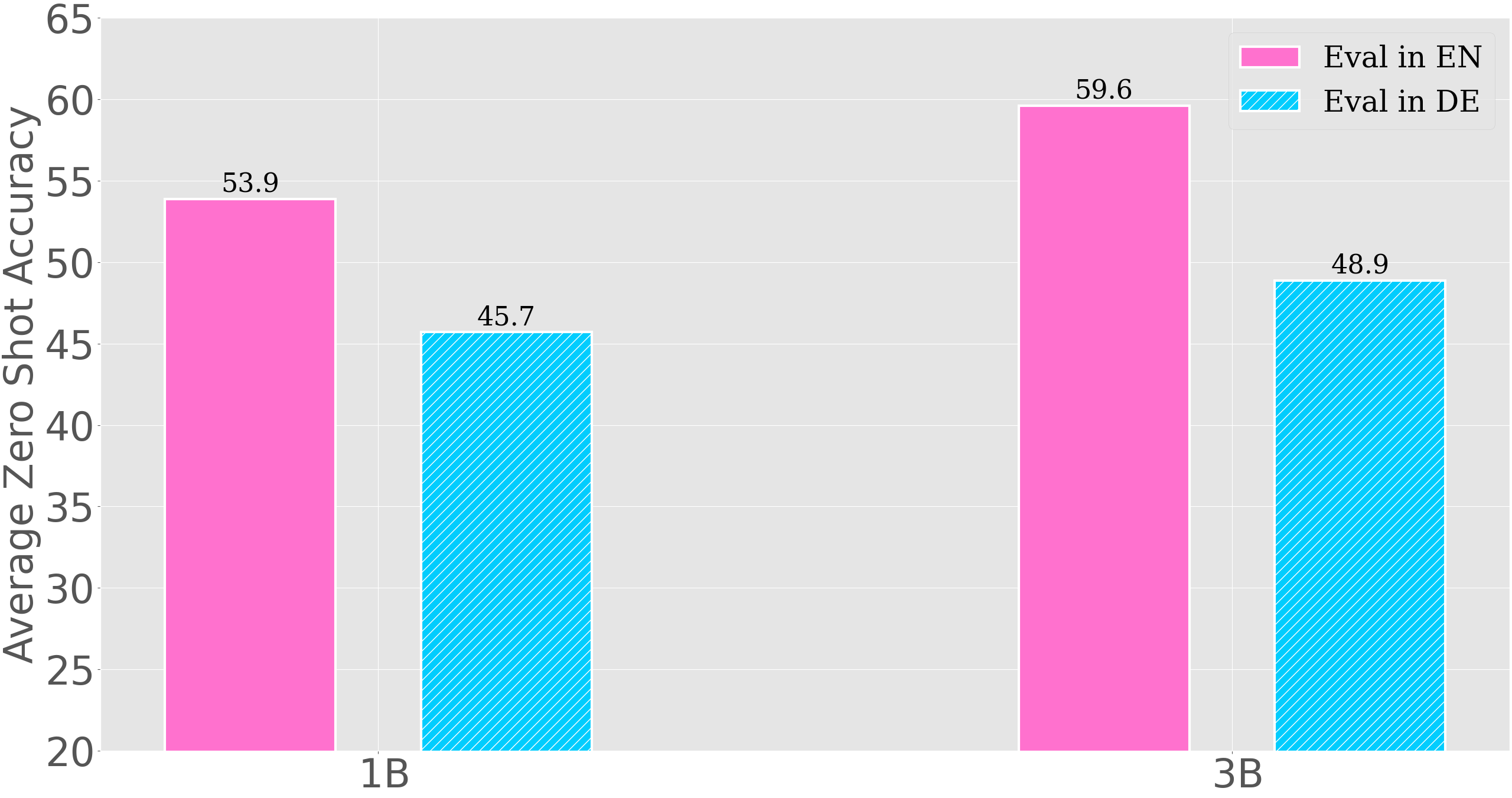}
        \caption{3B Model}
        \label{fig:barplot_1B_3B_1B_005}
    \end{subfigure}
    \caption{Average accuracy over zero-shot benchmark tasks in English and translated German comparing 1.3B and 2.7B models trained with 5\% of the training steps using (a) 250M tokens, and (b) 1B tokens in German. }
    \label{fig:barplot_1B_3B_005}
\end{figure*}

We evaluate both 1.3B and 2.7B models with 5\% data ratios with 250M tokens in the target language. Results are shown in Figure~\ref{fig:barplot_1B_3B_005}. 
Figure~\ref{fig:barplot_1B_3B_250M_005} shows that the 2.7B model performs slightly worse than the 1.3B model with the same 250M amount of target language data matching the similar perplexity values between the two models. However, increasing to 1B tokens results in larger improvements when increasing model size from 1.3B to 2.7B, as shown in Figure~\ref{fig:barplot_1B_3B_1B_005}.

We further conduct a comparison with different data ratios.  We experiment with different data ratios beyond 5\% used during training. For this experiment, we study the 1.3B parameter model and a 2.7B parameter model. Zero-shot accuracy is provided in Figure~\ref{fig:data_ratios}. We see that performance is around the same for 1.3B models, but 2.7B models perform worse with a 5\% data ratio.

\section{Results for 300M Models}
\label{sec:app_300M}

\begin{figure*}[htp!]
    \centering
       \begin{subfigure}{0.45\textwidth}
        \centering
        \includegraphics[width=\columnwidth]{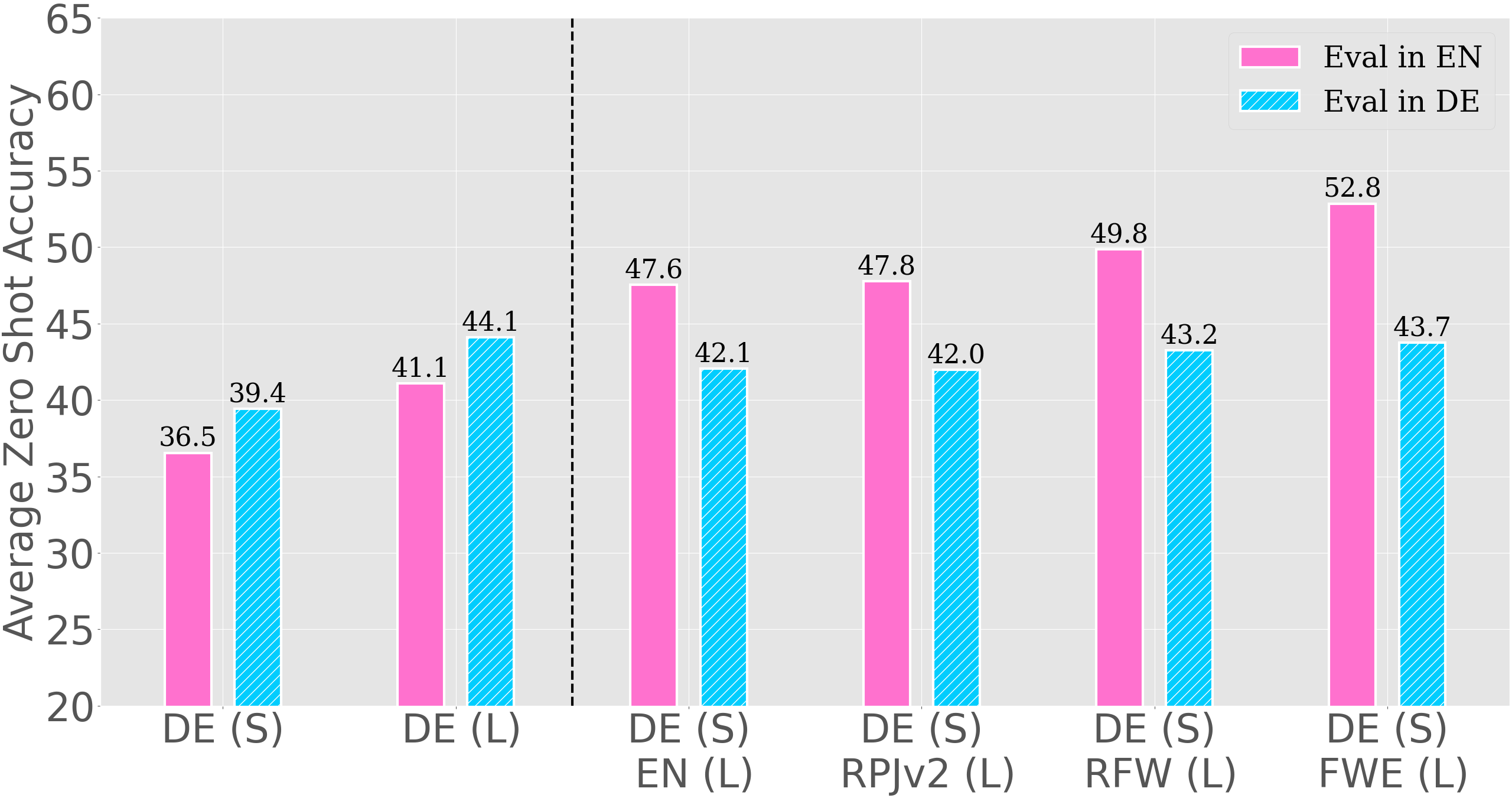}
    \caption{Higher quality English auxiliary data.}
    \label{fig:diff_datasets_300M}
    \end{subfigure}
    \vspace{5mm}
    \hfill
    \begin{subfigure}{0.45\textwidth}
        \centering
            \includegraphics[width=\columnwidth]{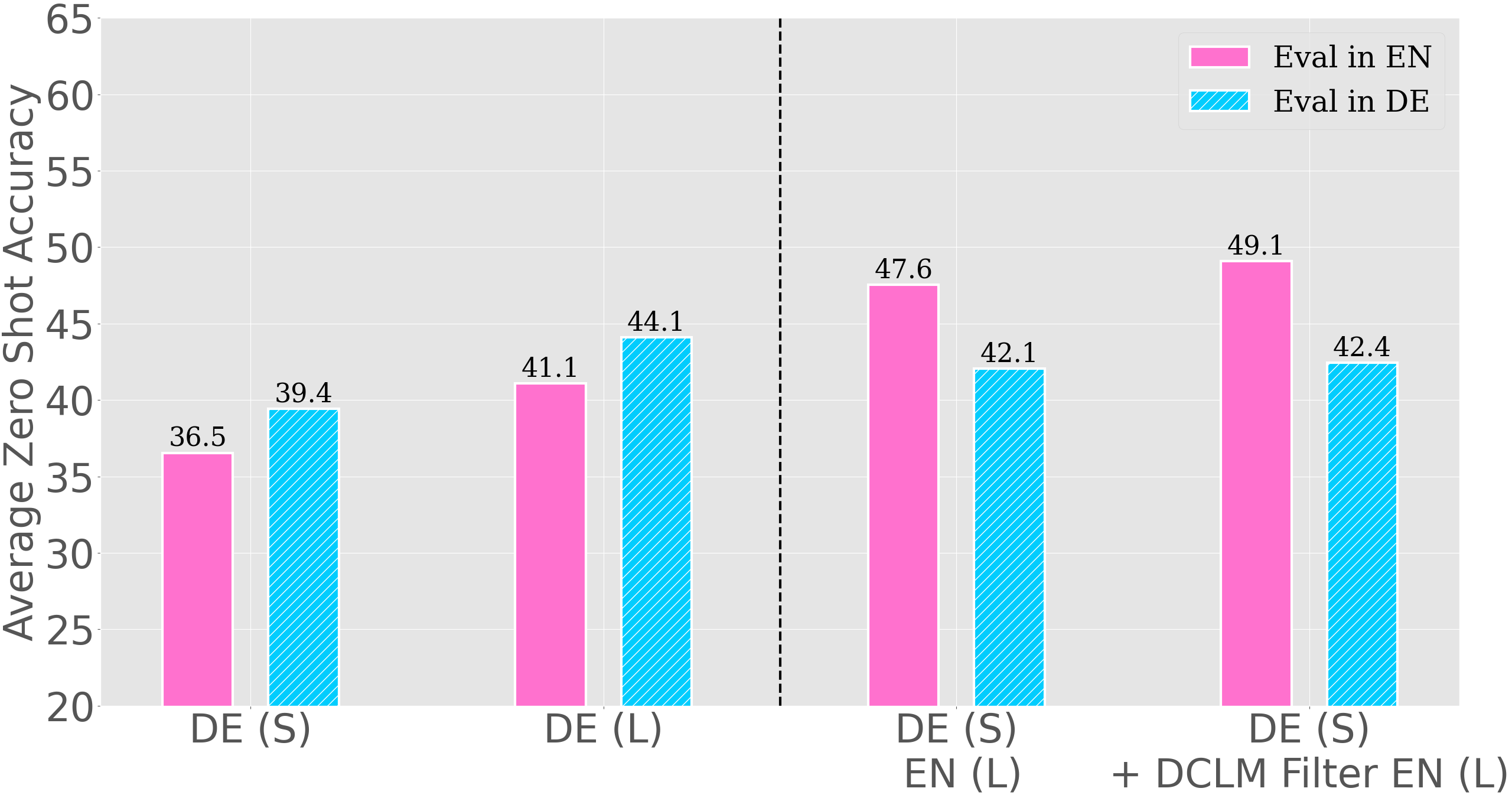}
    \caption{Model based filtering of English auxiliary data.}
    \label{fig:mb_filter_300M}
    \end{subfigure}
    \hfill
    \begin{subfigure}{0.45\textwidth}
        \centering
            \includegraphics[width=\columnwidth]{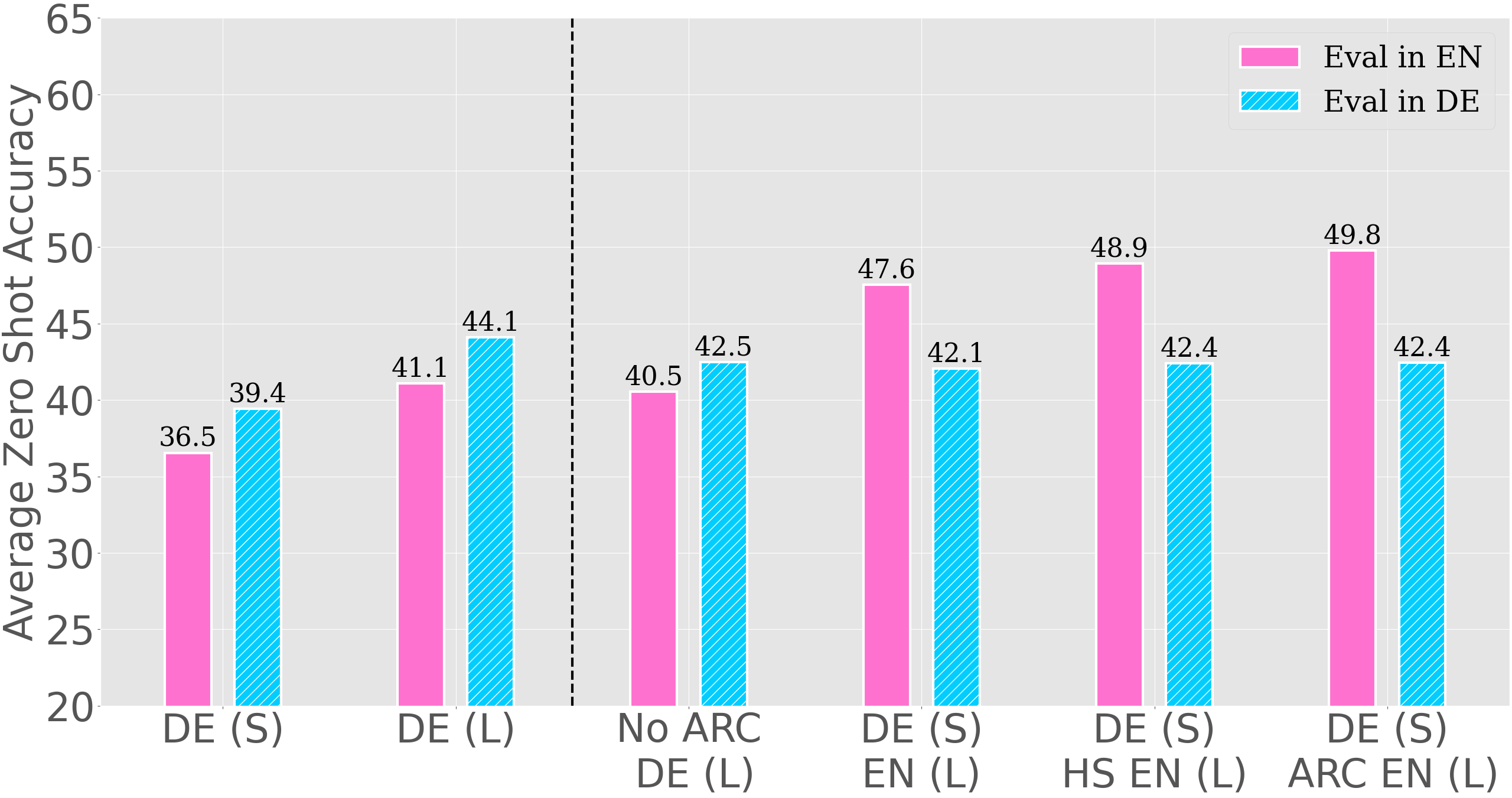}
    \caption{Upsampling of English auxiliary data.}
    \label{fig:data_selection_300M}
    \end{subfigure}
    \vspace{5mm}
    \hfill
    \begin{subfigure}{0.45\textwidth}
        \centering
    \includegraphics[width=\columnwidth]{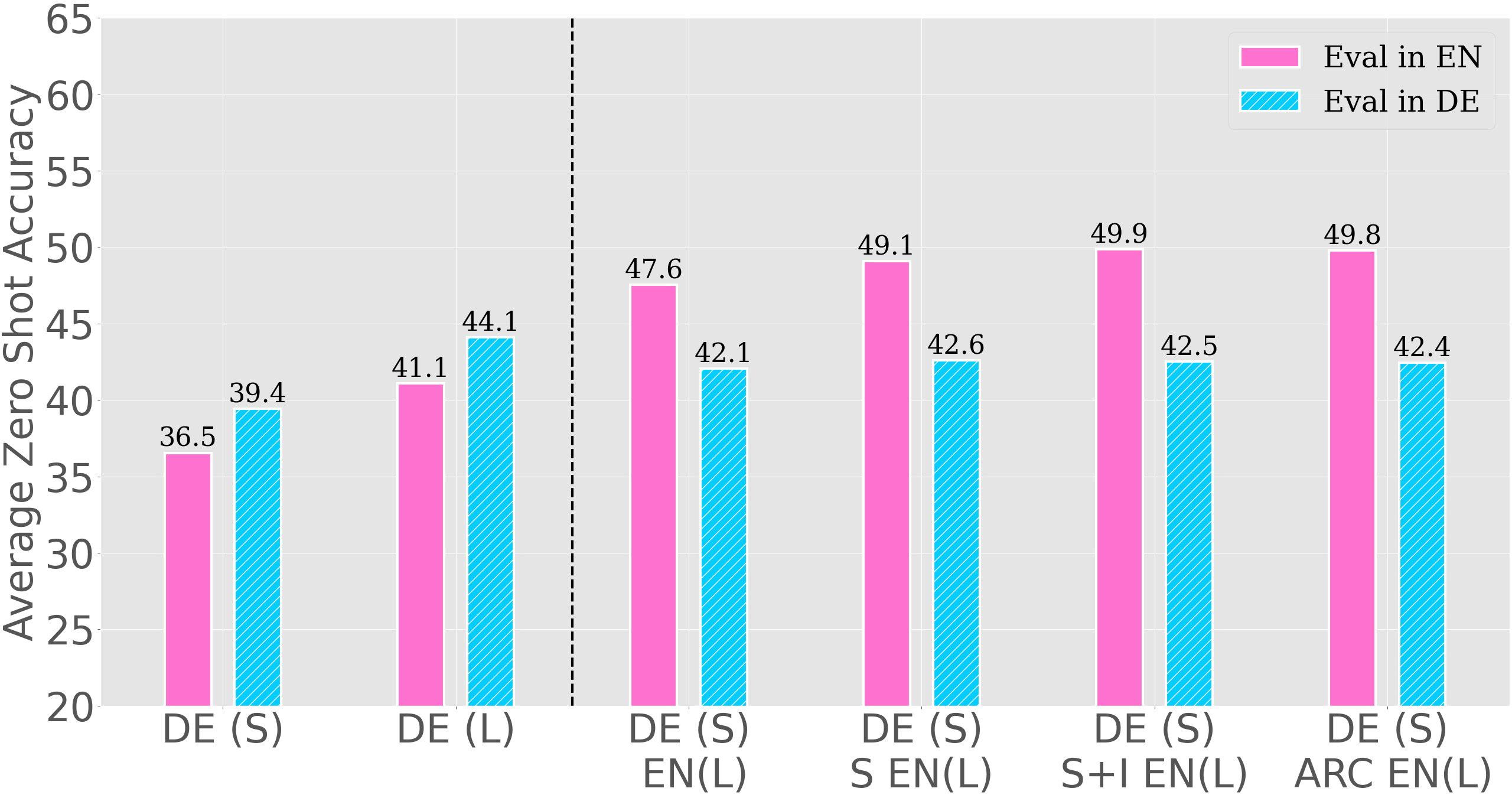}
    \caption{Synthetic data upsampling of English auxiliary data.}
    \label{fig:syn_selection_300M}
    \end{subfigure}
    \hfill
    \begin{subfigure}{0.45\textwidth}
        \centering
    \includegraphics[width=\columnwidth]{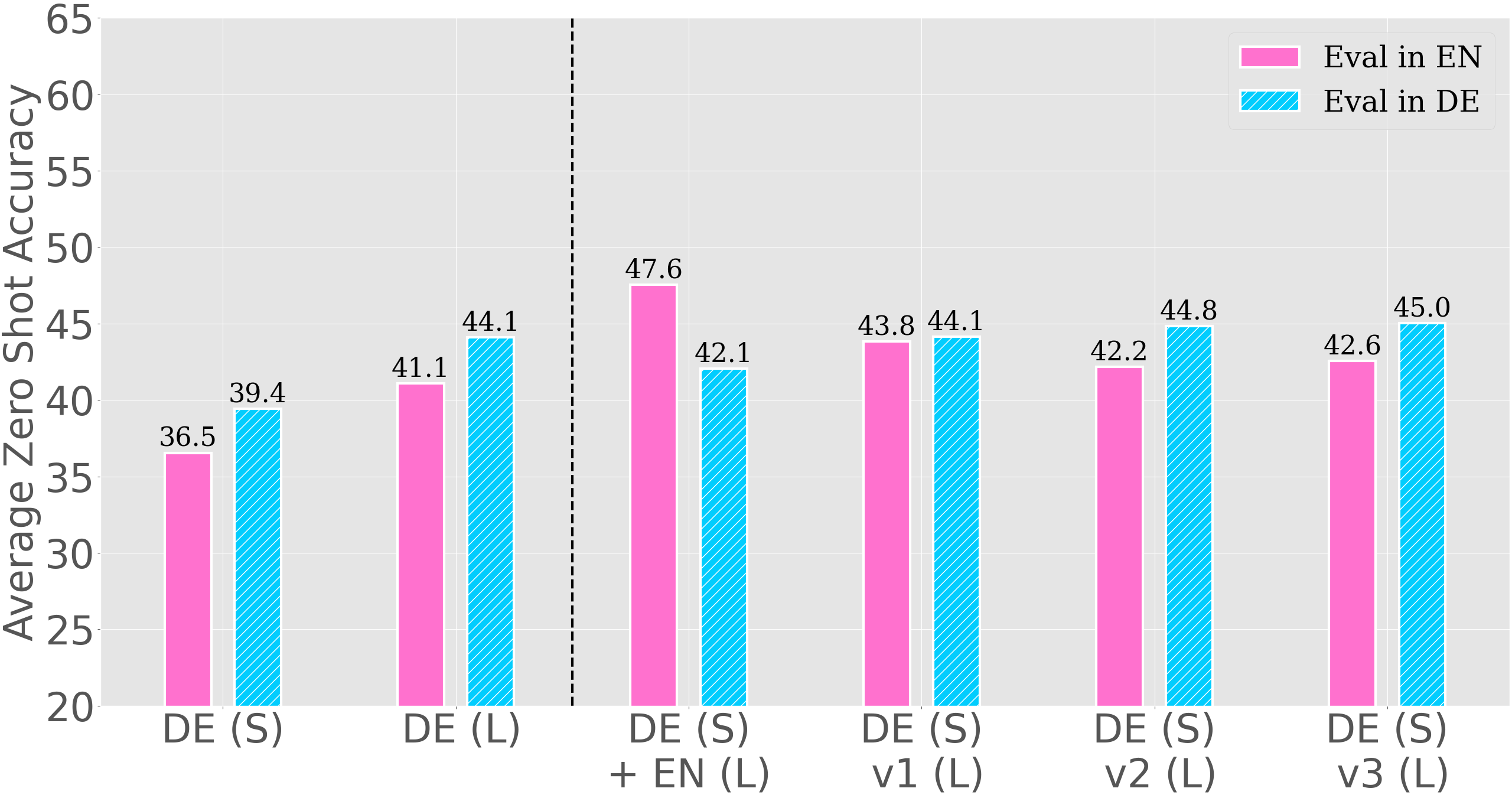}
    \caption{Translations of English auxiliary data.}
    \label{fig:translation_300M}
    \end{subfigure}
    \caption{Zero-shot accuracy of 300M models trained with different types of English auxiliary data.  Results are averaged over six evaluation datasets. For each setting evaluation is done in English (pink, left bars) and German (blue, right bars).}
    \label{fig:data_ratios}

\end{figure*}

We present results comparing approaches for 300M models with English as the auxiliary language and German as the target language. Figure~\ref{fig:diff_datasets_300M} shows performance with different datasets, Figure~\ref{fig:mb_filter_300M} shows performance with the DCLM filter, Figure~\ref{fig:data_selection_300M} shows performance with upsampling real data, Figure~\ref{fig:syn_selection_300M} shows performance upsampling synthetic data, and Figure~\ref{fig:translation_300M} shows performance with translation. All results show similar trends as in Section~\ref{sec:exp}-\ref{sec:methods_improve}, but the performance improvements are smaller.

\section{Results for Multiple Languages}
\label{sec:app_multilang_evals}
\subsection{Perplexity Evaluations for Translated Training Data}
\label{sec:appendix_ppl_evals}

In \S\ref{sec:multiple_languages} we found that performance trends were not the same across languages. In particular, French, German, Portuguese, and Spanish (belonging to the same language family) have similar patterns, but Chinese, Japanese, and Korean exhibit different patterns.  To further test whether the models retain knowledge from one language in another, we translate a small portion of the training set from FineWebEDU and mC4 English, totaling 10,000 documents.  We then translate the data using the v3 translation system from \S\ref{sec:translation}.  We measure both the macro perplexity of all documents as well as the fraction of times where the translated and original data from FineWebEDU (training set) have lower loss than the average loss of documents from mC4 English (not part of the training set but from a similar distribution). We refer to this quantity as translated and original exceedance. Having lower loss means the data is more familiar to the model, and having an equal exceedance across the original and translated data means the model can reason equally in either language. Our results are summarized in Table~\ref{tab:multilingual_perp}.  We find that perplexity is nearly identical for original data, but much higher for translated data in all languages.  For exceedance, in English, we see that the scores are all around 80\%.  However, we see that for Japanese and Chinese these values are much lower, indicating that seeing the data in English for these languages does not lower the perplexity in the target language and that the model is not making use of information in the other language. For Chinese evaluations, we note that the perplexity is much higher than for other languages, indicating that the translation system potentially causes higher perplexity and lower {exceedance.  However, we still note that for Japanese the exceedance is lower, and we expect that with better translation quality, the Chinese evaluations will be similar to Japanese. 

\begin{table*}
    \centering
    \scalebox{0.7}{
        \begin{tabular}{lcccccccc}
        \toprule
            Language & mC4-Train & mc4-Val & mC4-EN & mC4-EN Translated & FWE & FWE Translated & Original EX &  Translated EX\\
            
            German & 8.41 & 16.41 & 14.25 & 25.58 & 10.61 & 21.31 &  78.70 & 75.00\\
            French & 6.24 & 12.75 &  14.37 & 20.31 & 10.66 & 14.37 & 79.24 & 88.64\\
            \midrule 
            Japanese & 6.52 & 11.21 &  14.56 & 25.56 & 10.64 & 23.89 & 80.41 & 56.97\\
            Chinese & 4.35 & 21.38 & 15.37 & 165.90 & 10.69 & 210.43 & 84.07 & 27.71\\
        \bottomrule
        \end{tabular}
    }
        \caption{Perplexity evaluations for mC4 English and FineWeb-EDU comparing original data and translated versions for 1B models trained with 250M tokens from the target language and FineWeb-EDU as the auxiliary dataset.}
        \label{tab:multilingual_perp}
\end{table*}

\subsection{Other Auxiliary Languages}
\label{sec:auxiliary_zh}

In this section we give additional details for our experiments with other auxiliary languages than English, as presented in \S\ref{sec:multiple_languages}.
}

\begin{figure}[t]
    \centering
    \includegraphics[width=\linewidth]{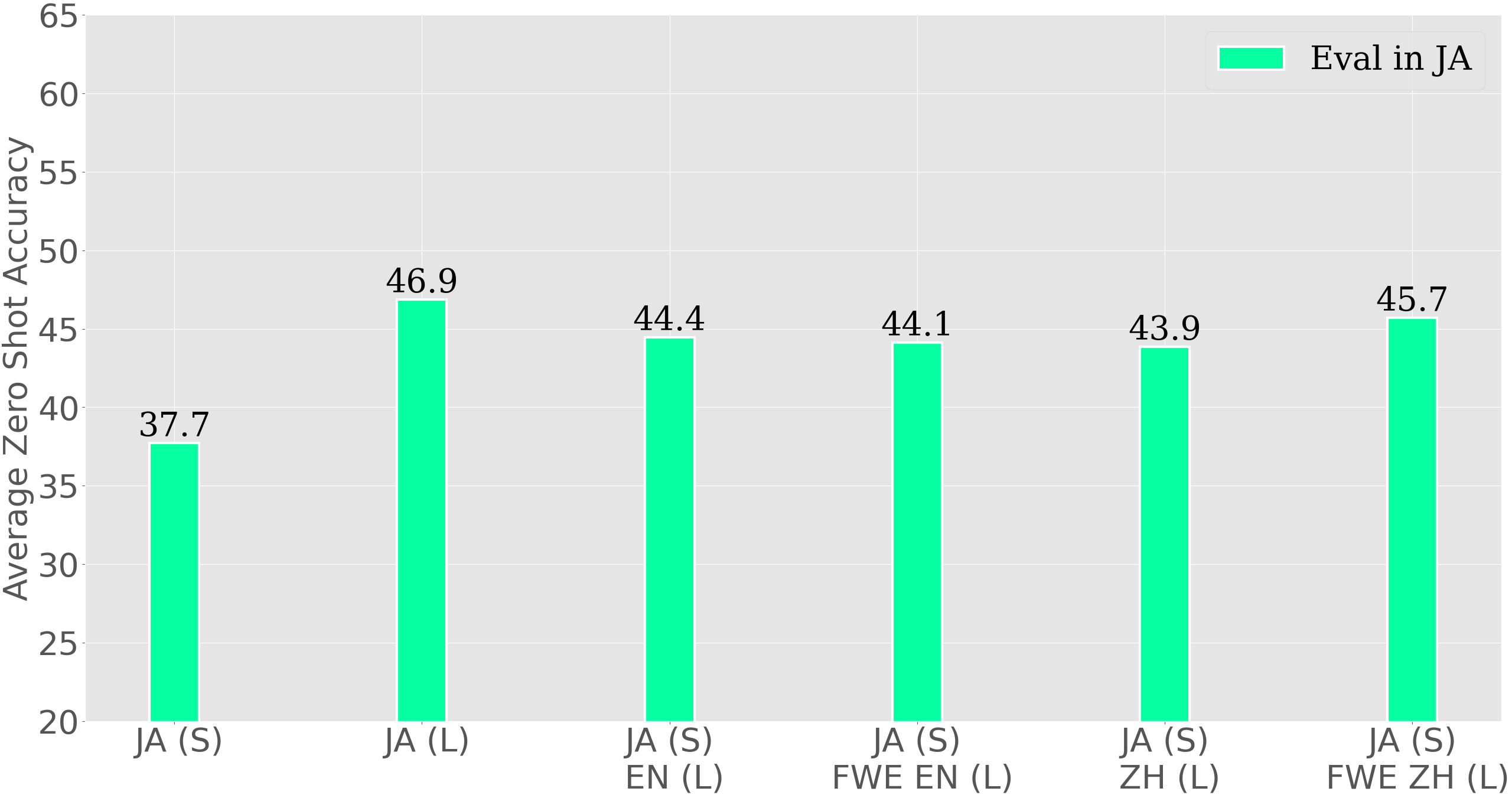}
    \caption{Average accuracy over zero-shot benchmark tasks in translated Japanese comparing Chinese and English auxiliary data. Models are 1.3B size and trained for 100B tokens.}
    \label{fig:barplot_chinese_aux_jp}
    \vspace{-2mm}
\end{figure}

\paragraph{Methodology.} We conduct experiments comparing using English and Chinese as an auxiliary language with Japanese as the target language.  For these experiments, we train a 1.3B model with the auxiliary dataset being either mC4 in English or Chinese, and English or Chinese versions of FineWebEDU\footnote{\url{https://huggingface.co/datasets/opencsg/chinese-fineweb-edu-v2}}. Our evaluations are on the Japanese translations.  

\paragraph{Findings.} We summarize the results in Figure~\ref{fig:barplot_chinese_aux_jp}. Our findings indicate that using FineWebEDU, a high quality Chinese dataset, as auxiliary data improves Japanese performance over mC4 in Chinese, and the performance increases over using English either from mC4 or FineWebEDU.  Note that the performance increases are consistent with improvements better English data for Indo-European langauges as seen in Figure~\ref{fig:monolingual_summary}.

\subsection{Multilingual Experiments}
\label{sec:multilingual_mix}

\paragraph{Motivation.} %
Our main findings investigate how auxiliary data benefits evaluations in a target language.  However, many models trained on other languages (beyond English) are trained with many languages simultaneously.  We investigate whether better auxiliary language datasets also improve multilingual model training.  

\paragraph{Methodology.}
We conduct experiments combining the German, French, Chinese, and Japanese language data towards multilingual training.  For these experiments, we train a 1.3B model with the auxiliary dataset being FineWebEDU and a mix of data from the four languages, totaling 5\% or 20\% of the training. We use this subset of four languages, as we keep the data ratios the same per language, and did not want to increase the amount of data in target languages beyond the typical multilingual ratios in large open-source models \cite{xu2024survey}.

\paragraph{Findings.}
We summarize the results in Table~\ref{tab:lm_eval_multilingual_1B}.
Our findings indicate that training with 20\% of the data being a combination of target languages yields similar performance to training with 5\%, resulting in a 1\% reduction in performance on average when training with 20\% multilingual data.  When compared with training a bilingual model, we observe performance decreases for German and French, and increases for Chinese and Japanese. %

\begin{table}[h!]
    \centering
        \begin{tabular}{lccccc}
           \toprule
            & \textbf{EN} & \textbf{DE} & \textbf{FR} & \textbf{JA} & \textbf{ZH} \\
            \midrule
            \textbf{5\%} & 61.07 & 46.02 & 46.64 & \textbf{44.00} & \textbf{45.92} \\
            \textbf{20\%} & 59.25 & 46.44 & 46.61 & 43.23 & 44.36 \\
            Bi & & \textbf{47.16} & \textbf{47.52} & 42.73 & 44.38\\
            \bottomrule
        \end{tabular}
    \caption{Evaluation of XL models in multilingual setting on ``General Understanding Tasks''  focusing on general reasoning, language understanding, and science knowledge in translated languages. Rows are the average accuracy for the respective language, with 5\% or 20\% of the training coming from a mix of the four languages. `Bi' refers to the bilingual models trained with target (S) + FWE EN (L).}
    \label{tab:lm_eval_multilingual_1B}
\end{table}

\subsection{Average Zero Shot Accuracy Plots}
\label{sec:app_multilang_avg}

\begin{figure*}[h]
    \centering
       \begin{subfigure}{0.44\textwidth}
        \centering
        \includegraphics[width=\textwidth]{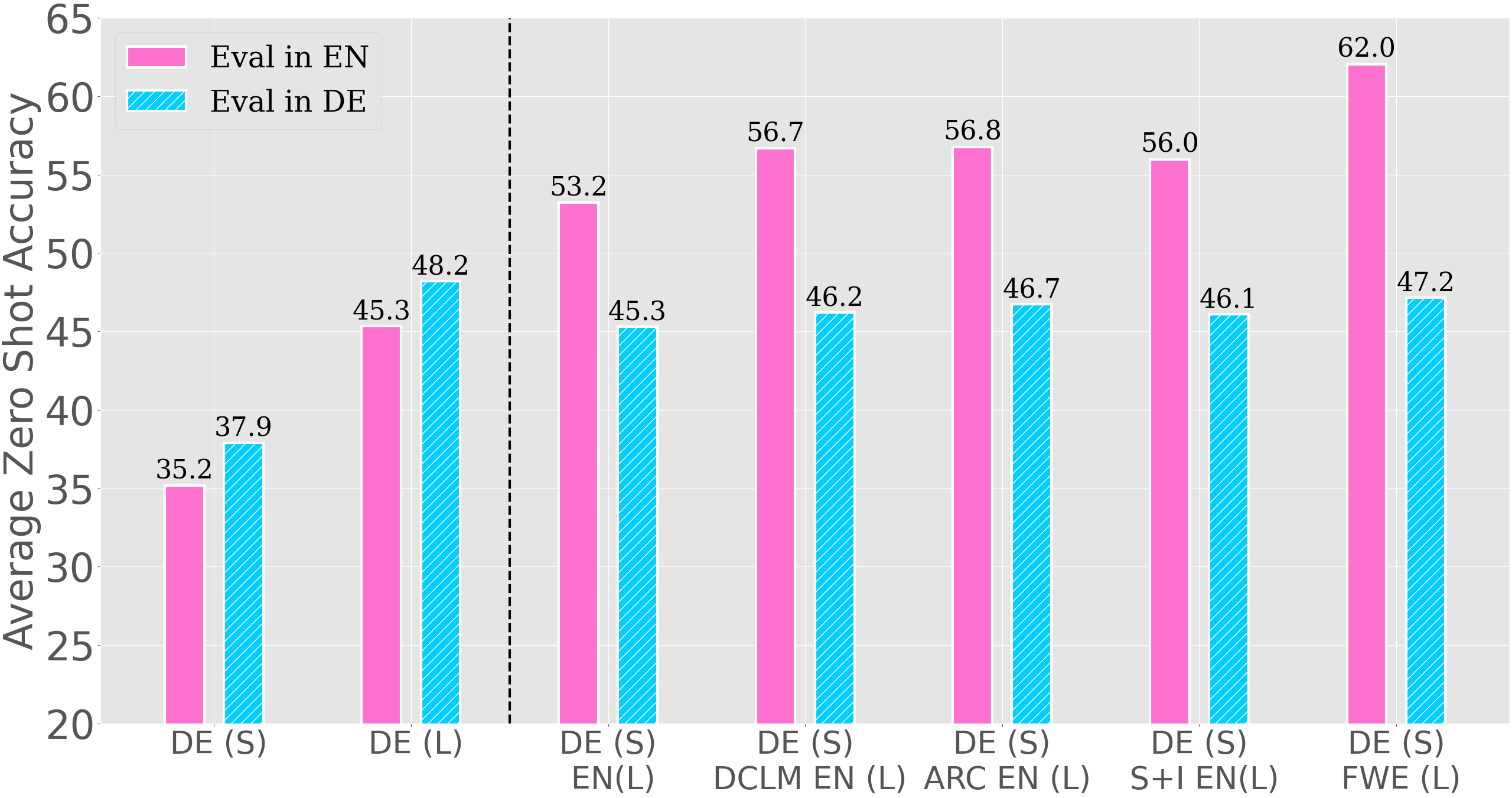}
        \caption{German}
        \label{fig:multilang_german_1B}
    \end{subfigure}
    \hfill
    \begin{subfigure}{0.44\textwidth}
        \centering
        \includegraphics[width=\textwidth]{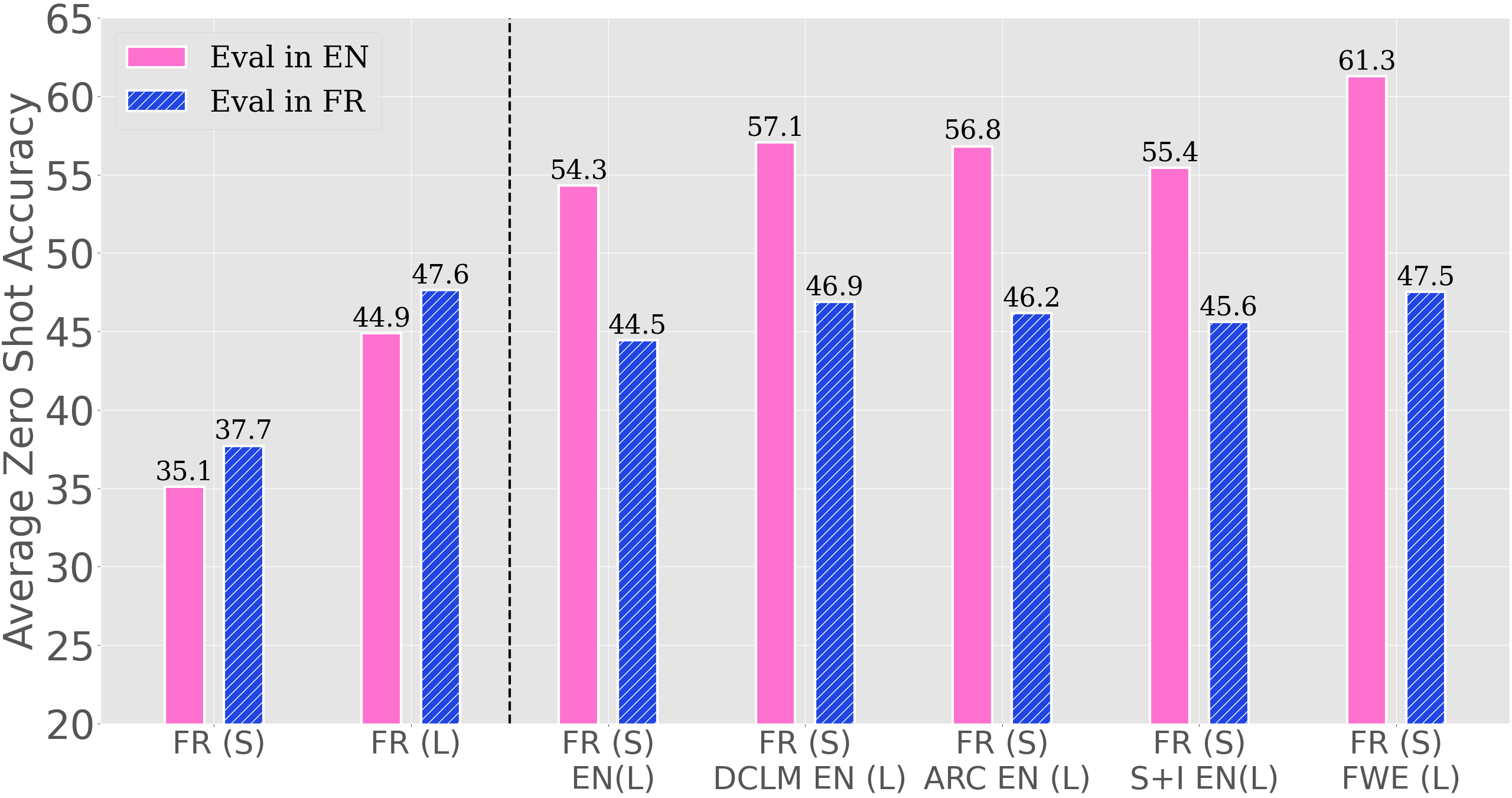}
        \caption{French}
        \label{fig:multilang_french_1B}
    \end{subfigure}
    \caption{Zero-shot accuracy of XL models trained with  various English auxiliary data for German and French.  Results are averaged over six eval datasets.}
    \label{fig:multilang_de_fr}
\end{figure*}

\begin{figure*}[h]
    \centering
    \begin{subfigure}{0.44\textwidth}
        \centering
        \includegraphics[width=\textwidth]{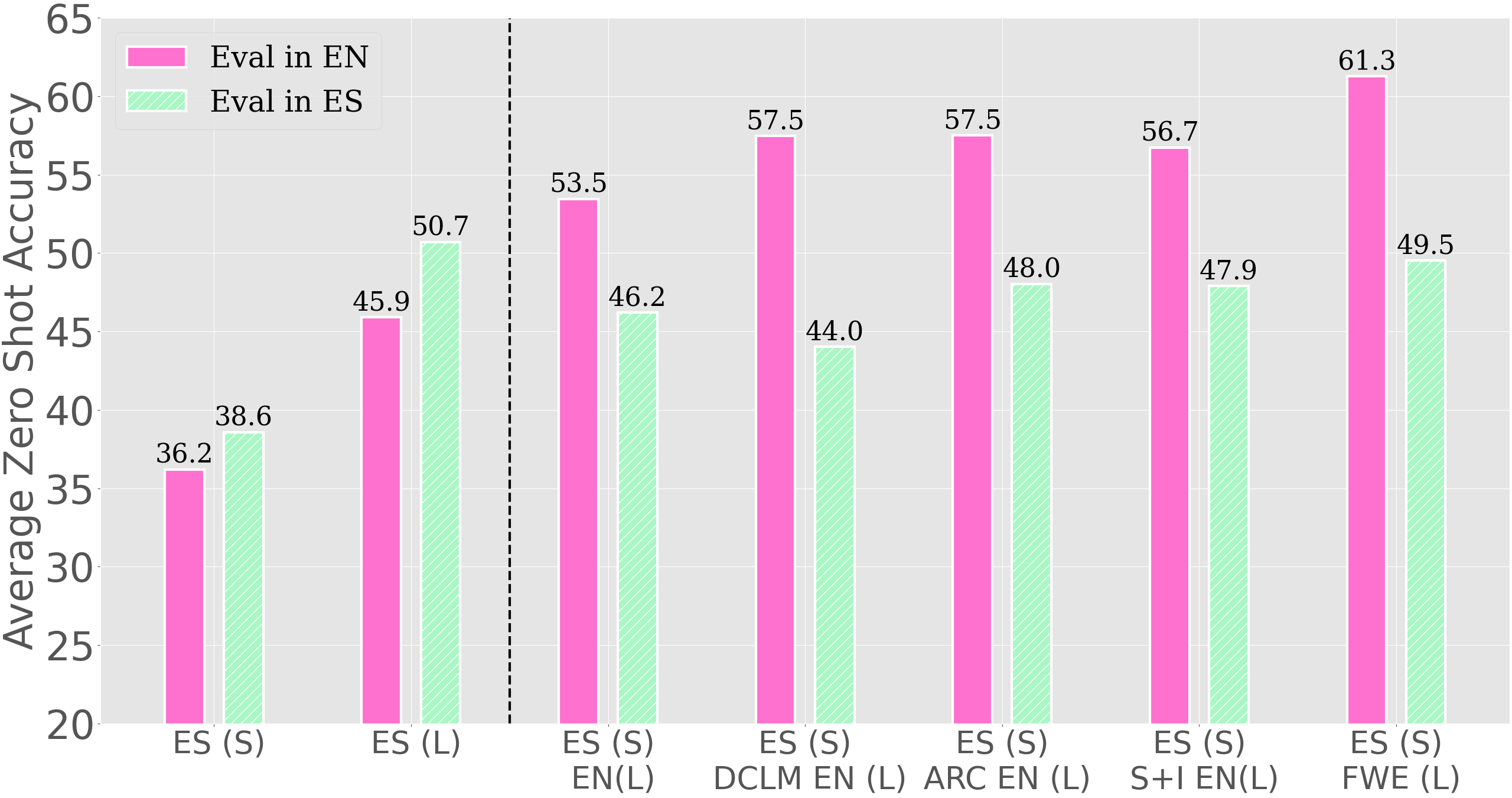}
        \caption{Spanish}
        \label{fig:multilang_spanish_1B}
    \end{subfigure}
    \hfill
    \begin{subfigure}{0.44\textwidth}
        \centering
        \includegraphics[width=\textwidth]{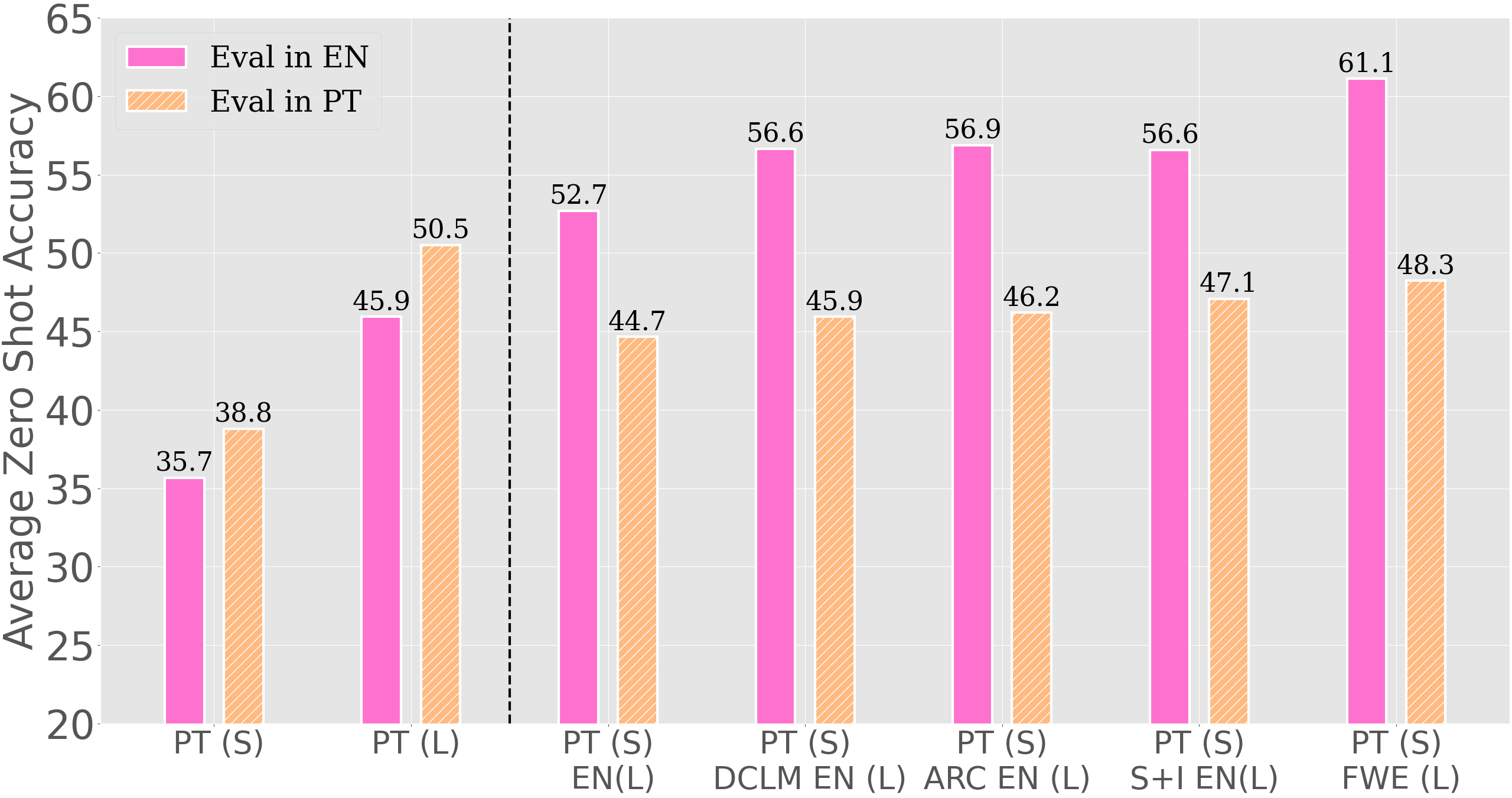}
        \caption{Portuguese}
        \label{fig:multilang_portuguese_1B}
    \end{subfigure}
    
    \caption{Zero-shot accuracy of XL models trained with  various English auxiliary data for Spanish and Portuguese.  Results are averaged over six eval datasets.}
    \label{fig:multilang_es_pt}
\end{figure*}

\begin{figure*}[h]
    \centering
    \begin{subfigure}{0.44\textwidth}
        \centering
        \includegraphics[width=\textwidth]{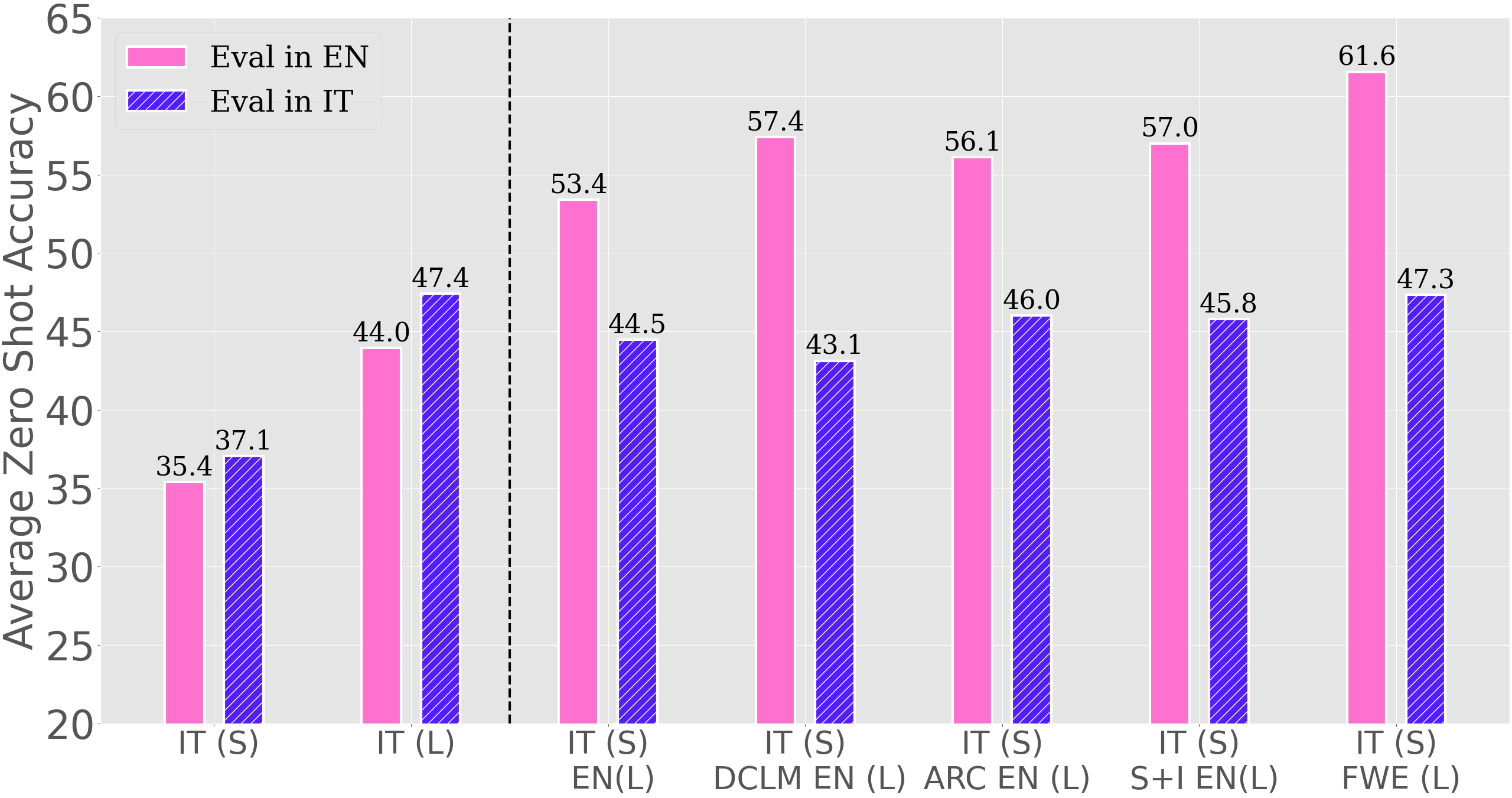}
        \caption{Italian}
        \label{fig:multilang_italian_1B}
    \end{subfigure}
    \hfill
    \begin{subfigure}{0.44\textwidth}
        \centering
        \includegraphics[width=\textwidth]{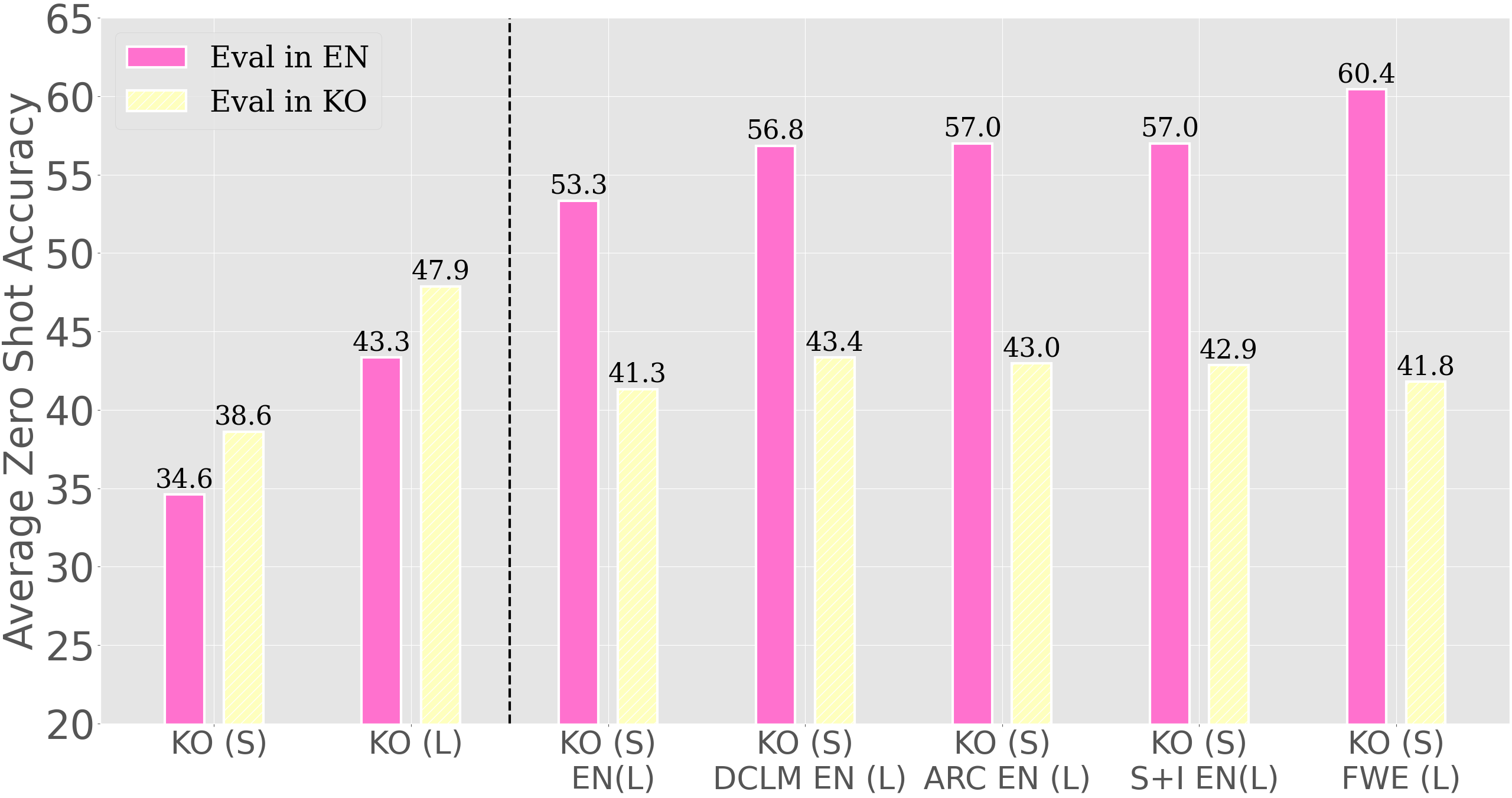}
        \caption{Korean}
        \label{fig:multilang_korean_1B}
    \end{subfigure}
    
    \caption{Zero-shot accuracy of XL models trained with  various English auxiliary data for Italian and Korean.  Results are averaged over six eval datasets.}
    \label{fig:multilang_it_ko}
\end{figure*}

\begin{figure*}[h]
    \centering
    \begin{subfigure}{0.44\textwidth}
        \centering
        \includegraphics[width=\textwidth]{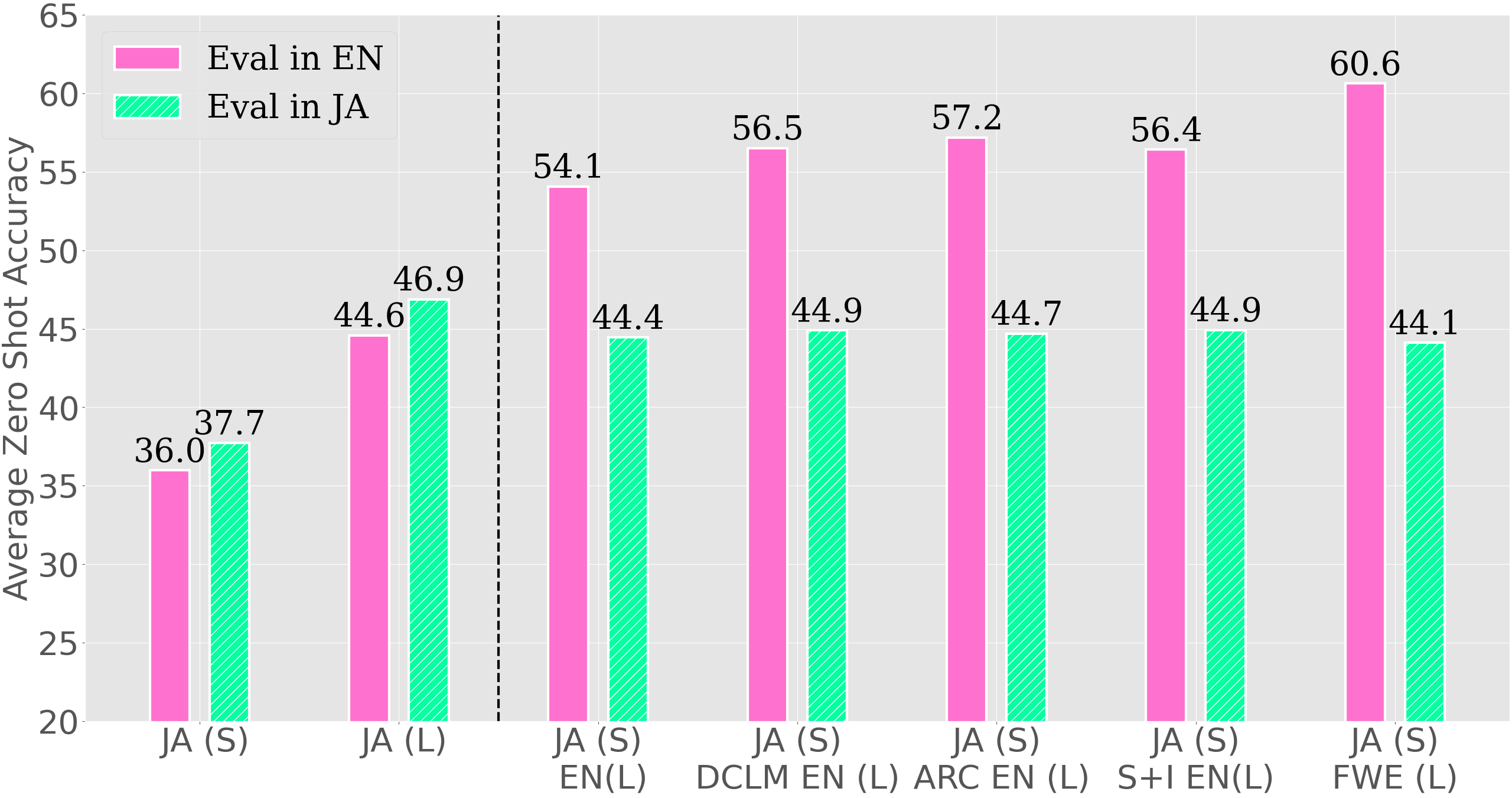}
        \caption{Japanese}
        \label{fig:multilang_japanese_1B}
    \end{subfigure}
    \hfill
    \begin{subfigure}{0.44\textwidth}
        \centering
        \includegraphics[width=\textwidth]{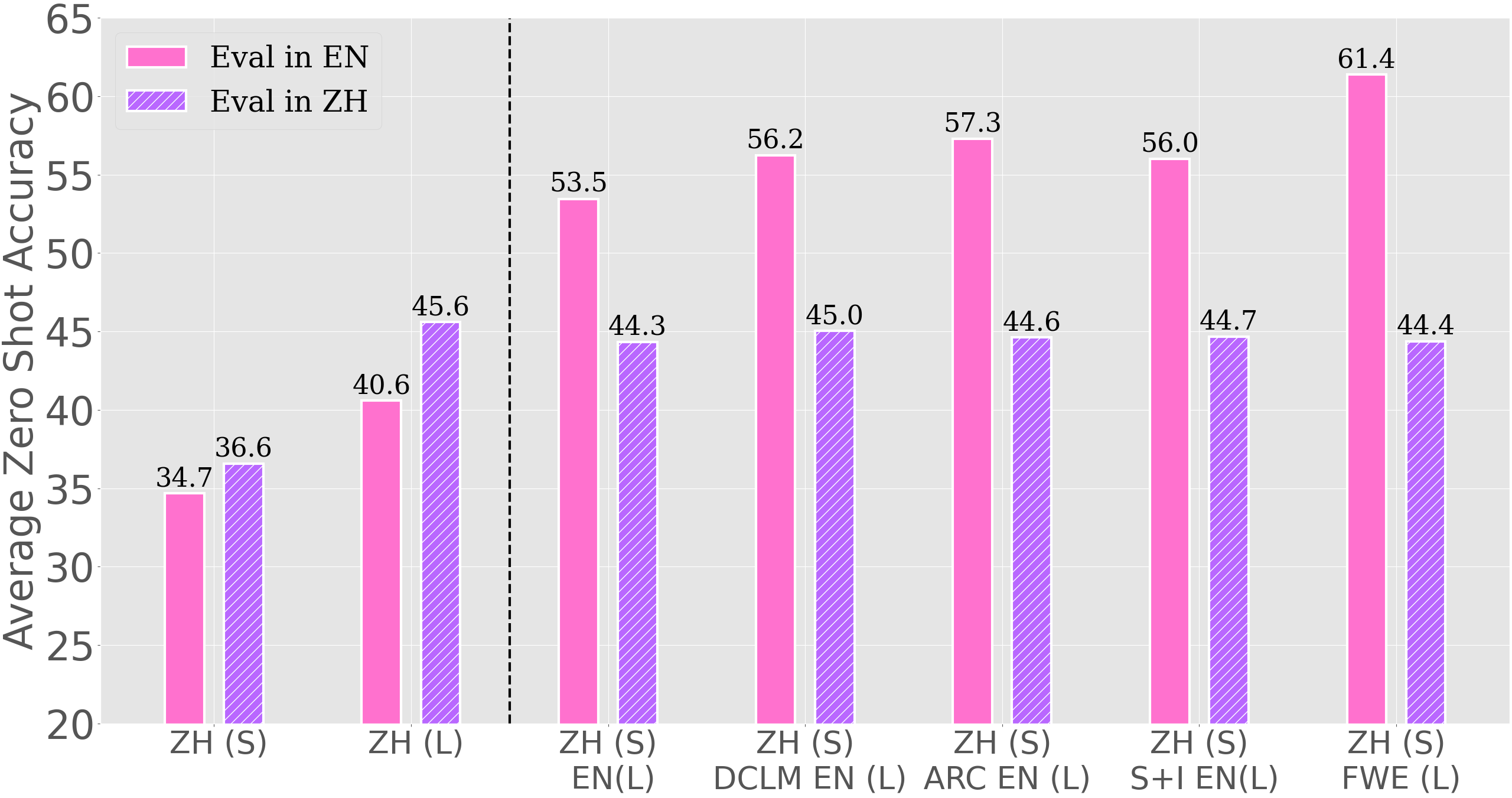}
        \caption{Chinese}
        \label{fig:multilang_chinese_1B}
    \end{subfigure}
    
    \caption{Zero-shot accuracy of XL models trained with  various English auxiliary data for Japanese and Chinese.  Results are averaged over six eval datasets.}
    \label{fig:multilang_ja_zh}
\end{figure*}

We present experimental results comparing the best performing approaches for French and German languages in Figure~\ref{fig:multilang_de_fr}, Spanish and Portuguese in Figure~\ref{fig:multilang_es_pt}, Italian and Korean in Figure~\ref{fig:multilang_it_ko}, and for Japanese and Chinese languages in Figure~\ref{fig:multilang_ja_zh}. For each language, we take the best performing dataset and model found in German from \S\ref{sec:dataset_comparisons}-\ref{sec:multiple_languages}.

\subsection{Individual Eval Dataset Results}

Results for individual evaluation datasets are shown for all languages in Tables~\ref{tab:lm_eval_1B_de_en_indiv}-\ref{tab:lm_eval_1B_zh_en_indiv}.
\clearpage

\begin{table*}[h!]
    \centering
    \scalebox{0.7}{
        \begin{tabular}{lcccccc>{\columncolor{LightCyan}}c}
            \toprule
            Model Name & ARC-C & ARC-E & HS & PIQA & SCIQ & WGrande & AVG \\
            \midrule
            DE (S) & $20.73 \pm 1.18$ & $27.53 \pm 0.92$ & $25.94 \pm 0.44$ & $52.88 \pm 1.16$ & $36.80 \pm 1.53$ & $47.12 \pm 1.40$ & $35.17$ \\
            DE (L) & $22.78 \pm 1.23$ & $37.92 \pm 1.00$ & $33.06 \pm 0.47$ & $62.35 \pm 1.13$ & $64.70 \pm 1.51$ & $51.14 \pm 1.40$ & $45.33$ \\
            DE (S)
             EN(L) & $25.94 \pm 1.28$ & $48.15 \pm 1.03$ & $48.50 \pm 0.50$ & $71.55 \pm 1.05$ & $73.10 \pm 1.40$ & $52.01 \pm 1.40$ & $53.21$ \\
            DE (S)
             DCLM EN (L) & $29.52 \pm 1.33$ & $57.03 \pm 1.02$ & $49.96 \pm 0.50$ & $71.87 \pm 1.05$ & $77.90 \pm 1.31$ & $53.83 \pm 1.40$ & $56.69$ \\
            DE (S)
             ARC EN (L) & $31.83 \pm 1.36$ & $57.24 \pm 1.02$ & $48.80 \pm 0.50$ & $73.07 \pm 1.04$ & $77.00 \pm 1.33$ & $52.72 \pm 1.40$ & $56.78$ \\
            DE (S)
             S+I EN(L) & $29.10 \pm 1.33$ & $55.22 \pm 1.02$ & $48.34 \pm 0.50$ & $71.71 \pm 1.05$ & $78.50 \pm 1.30$ & $53.04 \pm 1.40$ & $55.98$ \\
            DE (S)
             FWE (L) & $38.14 \pm 1.42$ & $66.37 \pm 0.97$ & $54.88 \pm 0.50$ & $72.25 \pm 1.04$ & $84.60 \pm 1.14$ & $56.04 \pm 1.39$ & $62.05$ \\
            \midrule 
            DE (S) & $22.69 \pm 1.24$ & $30.31 \pm 0.97$ & $28.99 \pm 0.47$ & $54.84 \pm 1.16$ & $40.00 \pm 1.59$ & $50.42 \pm 1.45$ & $37.88$ \\
            DE (L) & $27.44 \pm 1.32$ & $38.81 \pm 1.03$ & $39.53 \pm 0.51$ & $63.17 \pm 1.13$ & $67.68 \pm 1.52$ & $52.62 \pm 1.45$ & $48.21$ \\
            DE (S)
             EN(L) & $25.07 \pm 1.29$ & $37.70 \pm 1.02$ & $36.18 \pm 0.50$ & $59.36 \pm 1.15$ & $63.05 \pm 1.57$ & $50.51 \pm 1.45$ & $45.31$ \\
            DE (S)
             DCLM EN (L) & $25.77 \pm 1.30$ & $40.00 \pm 1.03$ & $35.94 \pm 0.50$ & $59.25 \pm 1.15$ & $65.05 \pm 1.55$ & $51.18 \pm 1.45$ & $46.20$ \\
            DE (S)
             ARC EN (L) & $26.91 \pm 1.32$ & $39.34 \pm 1.03$ & $35.77 \pm 0.50$ & $58.65 \pm 1.15$ & $67.58 \pm 1.52$ & $52.11 \pm 1.45$ & $46.73$ \\
            DE (S)
             S+I EN(L) & $25.24 \pm 1.29$ & $39.38 \pm 1.03$ & $35.34 \pm 0.49$ & $59.74 \pm 1.14$ & $64.21 \pm 1.56$ & $52.70 \pm 1.45$ & $46.10$ \\
            DE (S)
             FWE (L) & $26.91 \pm 1.32$ & $42.39 \pm 1.04$ & $37.40 \pm 0.50$ & $60.45 \pm 1.14$ & $65.37 \pm 1.54$ & $50.42 \pm 1.45$ & $47.16$ \\
            \bottomrule
        \end{tabular}
    }
    \caption{Evaluation of $1$B parameter XL model on ``General Understanding Tasks''  focusing on general reasoning, language understanding, and science knowledge in English followed by translated German. Results show the length normalized accuracy for individual datasets and the average over all datasets for all datasets.}
    \label{tab:lm_eval_1B_de_en_indiv}
\end{table*}

\begin{table*}[h!]
    \centering
    \scalebox{0.7}{
        \begin{tabular}{lcccccc>{\columncolor{LightCyan}}c}
            \toprule
            Model Name & ARC-C & ARC-E & HS & PIQA & SCIQ & WGrande & AVG \\
            \midrule
            FR (S) & $19.88 \pm 1.17$ & $26.60 \pm 0.91$ & $26.61 \pm 0.44$ & $52.56 \pm 1.17$ & $33.70 \pm 1.50$ & $51.14 \pm 1.40$ & $35.08$ \\
            FR (L) & $24.66 \pm 1.26$ & $36.78 \pm 0.99$ & $32.89 \pm 0.47$ & $60.66 \pm 1.14$ & $63.90 \pm 1.52$ & $50.51 \pm 1.41$ & $44.90$ \\
            FR (S)
             EN(L) & $26.28 \pm 1.29$ & $49.92 \pm 1.03$ & $48.62 \pm 0.50$ & $71.98 \pm 1.05$ & $75.20 \pm 1.37$ & $53.91 \pm 1.40$ & $54.32$ \\
            FR (S)
             DCLM EN (L) & $29.52 \pm 1.33$ & $56.44 \pm 1.02$ & $49.92 \pm 0.50$ & $72.74 \pm 1.04$ & $79.70 \pm 1.27$ & $54.06 \pm 1.40$ & $57.06$ \\
            FR (S)
             ARC EN (L) & $29.69 \pm 1.34$ & $57.28 \pm 1.02$ & $48.96 \pm 0.50$ & $72.91 \pm 1.04$ & $78.90 \pm 1.29$ & $53.12 \pm 1.40$ & $56.81$ \\
            FR (S)
             S+I EN(L) & $28.41 \pm 1.32$ & $53.16 \pm 1.02$ & $47.44 \pm 0.50$ & $70.51 \pm 1.06$ & $78.90 \pm 1.29$ & $54.14 \pm 1.40$ & $55.43$ \\
            FR (S)
             FWE (L) & $36.69 \pm 1.41$ & $65.40 \pm 0.98$ & $54.81 \pm 0.50$ & $72.74 \pm 1.04$ & $82.50 \pm 1.20$ & $55.41 \pm 1.40$ & $61.26$ \\
            \midrule 
            FR (S) & $22.06 \pm 1.22$ & $28.53 \pm 0.95$ & $29.25 \pm 0.47$ & $55.93 \pm 1.16$ & $40.29 \pm 1.59$ & $50.04 \pm 1.44$ & $37.68$ \\
            FR (L) & $25.98 \pm 1.30$ & $38.53 \pm 1.02$ & $41.71 \pm 0.51$ & $64.25 \pm 1.12$ & $62.12 \pm 1.57$ & $53.25 \pm 1.43$ & $47.64$ \\
            FR (S)
             EN(L) & $23.54 \pm 1.25$ & $37.03 \pm 1.01$ & $37.66 \pm 0.50$ & $58.98 \pm 1.15$ & $60.44 \pm 1.58$ & $49.05 \pm 1.43$ & $44.45$ \\
            FR (S)
             DCLM EN (L) & $26.50 \pm 1.30$ & $38.93 \pm 1.02$ & $38.85 \pm 0.50$ & $60.50 \pm 1.14$ & $66.11 \pm 1.53$ & $50.37 \pm 1.43$ & $46.88$ \\
            FR (S)
             ARC EN (L) & $25.54 \pm 1.29$ & $39.32 \pm 1.03$ & $38.17 \pm 0.50$ & $59.41 \pm 1.15$ & $64.22 \pm 1.55$ & $50.29 \pm 1.44$ & $46.16$ \\
            FR (S)
             S+I EN(L) & $25.72 \pm 1.29$ & $38.71 \pm 1.02$ & $37.61 \pm 0.50$ & $59.03 \pm 1.15$ & $63.06 \pm 1.56$ & $49.55 \pm 1.43$ & $45.61$ \\
            FR (S)
             FWE (L) & $27.38 \pm 1.32$ & $40.51 \pm 1.03$ & $40.67 \pm 0.51$ & $60.94 \pm 1.14$ & $64.85 \pm 1.55$ & $50.78 \pm 1.43$ & $47.52$ \\
            \bottomrule
        \end{tabular}
    }
    \caption{Evaluation of $1$B parameter XL model on ``General Understanding Tasks''  focusing on general reasoning, language understanding, and science knowledge in English followed by French. Results show the length normalized accuracy for individual datasets and the average over all datasets for all datasets.}
    \label{tab:lm_eval_1B_fr_en_indiv}
\end{table*}

\begin{table*}[h!]
    \centering
    \scalebox{0.7}{
        \begin{tabular}{lcccccc>{\columncolor{LightCyan}}c}
            \toprule
            Model Name & ARC-C & ARC-E & HS & PIQA & SCIQ & WGrande & AVG \\
            \midrule
            ES (S) & $21.16 \pm 1.19$ & $28.45 \pm 0.93$ & $26.33 \pm 0.44$ & $52.61 \pm 1.16$ & $38.50 \pm 1.54$ & $50.04 \pm 1.41$ & $36.18$ \\
            ES (L) & $23.29 \pm 1.24$ & $38.55 \pm 1.00$ & $33.86 \pm 0.47$ & $60.83 \pm 1.14$ & $67.90 \pm 1.48$ & $51.07 \pm 1.40$ & $45.92$ \\
            ES (S)
             EN(L) & $26.19 \pm 1.28$ & $48.95 \pm 1.03$ & $48.16 \pm 0.50$ & $71.11 \pm 1.06$ & $74.30 \pm 1.38$ & $52.01 \pm 1.40$ & $53.45$ \\
            ES (S)
             DCLM EN (L) & $29.86 \pm 1.34$ & $58.08 \pm 1.01$ & $50.55 \pm 0.50$ & $72.36 \pm 1.04$ & $79.40 \pm 1.28$ & $54.62 \pm 1.40$ & $57.48$ \\
            ES (S)
             ARC EN (L) & $30.89 \pm 1.35$ & $58.71 \pm 1.01$ & $49.41 \pm 0.50$ & $73.45 \pm 1.03$ & $78.50 \pm 1.30$ & $54.06 \pm 1.40$ & $57.50$ \\
            ES (S)
             S+I EN(L) & $29.44 \pm 1.33$ & $57.49 \pm 1.01$ & $48.89 \pm 0.50$ & $70.57 \pm 1.06$ & $79.70 \pm 1.27$ & $54.22 \pm 1.40$ & $56.72$ \\
            ES (S)
             FWE (L) & $36.60 \pm 1.41$ & $64.86 \pm 0.98$ & $54.95 \pm 0.50$ & $71.87 \pm 1.05$ & $81.50 \pm 1.23$ & $57.85 \pm 1.39$ & $61.27$ \\
            \midrule 
            ES (S) & $21.80 \pm 1.22$ & $29.90 \pm 0.96$ & $29.84 \pm 0.47$ & $57.45 \pm 1.15$ & $41.85 \pm 1.60$ & $50.61 \pm 1.42$ & $38.57$ \\
            ES (L) & $27.99 \pm 1.33$ & $44.12 \pm 1.04$ & $43.64 \pm 0.51$ & $65.94 \pm 1.11$ & $69.72 \pm 1.49$ & $52.70 \pm 1.42$ & $50.69$ \\
            ES (S)
             EN(L) & $25.28 \pm 1.28$ & $38.71 \pm 1.02$ & $39.32 \pm 0.50$ & $60.94 \pm 1.14$ & $65.83 \pm 1.54$ & $47.13 \pm 1.42$ & $46.20$ \\
            ES (S)
             DCLM EN (L) & $24.41 \pm 1.27$ & $35.31 \pm 1.00$ & $33.01 \pm 0.49$ & $56.64 \pm 1.16$ & $64.35 \pm 1.55$ & $50.36 \pm 1.42$ & $44.01$ \\
            ES (S)
             ARC EN (L) & $25.89 \pm 1.29$ & $42.01 \pm 1.04$ & $39.49 \pm 0.50$ & $62.02 \pm 1.13$ & $66.25 \pm 1.53$ & $52.54 \pm 1.42$ & $48.03$ \\
            ES (S)
             S+I EN(L) & $25.89 \pm 1.29$ & $41.08 \pm 1.03$ & $39.35 \pm 0.50$ & $61.37 \pm 1.14$ & $67.93 \pm 1.51$ & $51.82 \pm 1.42$ & $47.91$ \\
            ES (S)
             FWE (L) & $27.99 \pm 1.33$ & $43.24 \pm 1.04$ & $42.05 \pm 0.51$ & $62.19 \pm 1.13$ & $70.45 \pm 1.48$ & $51.17 \pm 1.42$ & $49.51$ \\
            \bottomrule
        \end{tabular}
    }
    \caption{Evaluation of $1$B parameter XL model on ``General Understanding Tasks''  focusing on general reasoning, language understanding, and science knowledge in English followed by Spanish. Results show the length normalized accuracy for individual datasets and the average over all datasets for all datasets.}
    \label{tab:lm_eval_1B_es_en_indiv}
\end{table*}

\begin{table*}[h!]
    \centering
    \scalebox{0.7}{
        \begin{tabular}{lcccccc>{\columncolor{LightCyan}}c}
            \toprule
            Model Name & ARC-C & ARC-E & HS & PIQA & SCIQ & WGrande & AVG \\
            \midrule
            PT (S) & $23.38 \pm 1.24$ & $26.77 \pm 0.91$ & $26.01 \pm 0.44$ & $53.48 \pm 1.16$ & $33.50 \pm 1.49$ & $50.91 \pm 1.41$ & $35.67$ \\
            PT (L) & $23.98 \pm 1.25$ & $39.14 \pm 1.00$ & $33.26 \pm 0.47$ & $60.99 \pm 1.14$ & $66.30 \pm 1.50$ & $52.01 \pm 1.40$ & $45.95$ \\
            PT (S)
             EN(L) & $25.34 \pm 1.27$ & $48.02 \pm 1.03$ & $47.19 \pm 0.50$ & $71.60 \pm 1.05$ & $74.00 \pm 1.39$ & $50.04 \pm 1.41$ & $52.70$ \\
            PT (S)
             DCLM EN (L) & $29.44 \pm 1.33$ & $53.79 \pm 1.02$ & $49.53 \pm 0.50$ & $71.49 \pm 1.05$ & $80.30 \pm 1.26$ & $55.33 \pm 1.40$ & $56.65$ \\
            PT (S)
             ARC EN (L) & $30.72 \pm 1.35$ & $57.83 \pm 1.01$ & $48.79 \pm 0.50$ & $73.01 \pm 1.04$ & $78.30 \pm 1.30$ & $52.64 \pm 1.40$ & $56.88$ \\
            PT (S)
             S+I EN(L) & $29.35 \pm 1.33$ & $57.45 \pm 1.01$ & $48.26 \pm 0.50$ & $71.55 \pm 1.05$ & $78.80 \pm 1.29$ & $53.99 \pm 1.40$ & $56.56$ \\
            PT (S)
             FWE (L) & $35.07 \pm 1.39$ & $64.39 \pm 0.98$ & $54.81 \pm 0.50$ & $72.69 \pm 1.04$ & $82.10 \pm 1.21$ & $57.62 \pm 1.39$ & $61.11$ \\
             \midrule 
             PT (S) & $25.37 \pm 1.29$ & $30.74 \pm 0.97$ & $29.18 \pm 0.47$ & $55.39 \pm 1.16$ & $41.97 \pm 1.60$ & $50.00 \pm 1.43$ & $38.77$ \\
            PT (L) & $30.25 \pm 1.36$ & $43.55 \pm 1.04$ & $42.36 \pm 0.51$ & $64.80 \pm 1.11$ & $69.88 \pm 1.49$ & $52.11 \pm 1.42$ & $50.49$ \\
            PT (S)
             EN(L) & $24.06 \pm 1.26$ & $38.04 \pm 1.02$ & $36.04 \pm 0.50$ & $59.68 \pm 1.14$ & $59.18 \pm 1.59$ & $51.06 \pm 1.42$ & $44.68$ \\
            PT (S)
             DCLM EN (L) & $25.37 \pm 1.29$ & $39.59 \pm 1.03$ & $37.15 \pm 0.50$ & $58.71 \pm 1.15$ & $66.11 \pm 1.53$ & $48.70 \pm 1.42$ & $45.94$ \\
            PT (S)
             ARC EN (L) & $27.90 \pm 1.32$ & $38.79 \pm 1.02$ & $36.87 \pm 0.50$ & $60.07 \pm 1.14$ & $64.32 \pm 1.55$ & $49.35 \pm 1.42$ & $46.22$ \\
            PT (S)
             S+I EN(L) & $27.03 \pm 1.31$ & $40.38 \pm 1.03$ & $37.72 \pm 0.50$ & $60.72 \pm 1.14$ & $66.00 \pm 1.54$ & $50.73 \pm 1.42$ & $47.10$ \\
            PT (S)
             FWE (L) & $28.86 \pm 1.34$ & $43.11 \pm 1.04$ & $39.81 \pm 0.51$ & $60.34 \pm 1.14$ & $66.84 \pm 1.53$ & $50.57 \pm 1.42$ & $48.25$ \\
            \bottomrule
        \end{tabular}
    }
    \caption{Evaluation of $1$B parameter XL model on ``General Understanding Tasks''  focusing on general reasoning, language understanding, and science knowledge in English followed by Portuguese. Results show the length normalized accuracy for individual datasets and the average over all datasets for all datasets.}
    \label{tab:lm_eval_1B_pt_en_indiv}
\end{table*}

\begin{table*}[h!]
    \centering
    \scalebox{0.7}{
        \begin{tabular}{lcccccc>{\columncolor{LightCyan}}c}
            \toprule
            Model Name & ARC-C & ARC-E & HS & PIQA & SCIQ & WGrande & AVG \\
            \midrule
            IT (S) & $23.46 \pm 1.24$ & $27.23 \pm 0.91$ & $26.30 \pm 0.44$ & $53.21 \pm 1.16$ & $32.40 \pm 1.48$ & $49.88 \pm 1.41$ & $35.41$ \\
            IT (L) & $21.59 \pm 1.20$ & $35.77 \pm 0.98$ & $32.54 \pm 0.47$ & $59.09 \pm 1.15$ & $63.70 \pm 1.52$ & $51.07 \pm 1.40$ & $43.96$ \\
            IT (S)
             EN(L) & $26.11 \pm 1.28$ & $48.53 \pm 1.03$ & $47.16 \pm 0.50$ & $71.71 \pm 1.05$ & $73.50 \pm 1.40$ & $53.35 \pm 1.40$ & $53.39$ \\
            IT (S)
             DCLM EN (L) & $31.14 \pm 1.35$ & $57.37 \pm 1.01$ & $50.89 \pm 0.50$ & $72.42 \pm 1.04$ & $79.50 \pm 1.28$ & $53.12 \pm 1.40$ & $57.40$ \\
            IT (S)
             ARC EN (L) & $30.46 \pm 1.34$ & $56.06 \pm 1.02$ & $48.78 \pm 0.50$ & $73.01 \pm 1.04$ & $75.90 \pm 1.35$ & $52.57 \pm 1.40$ & $56.13$ \\
            IT (S)
             S+I EN(L) & $30.38 \pm 1.34$ & $56.78 \pm 1.02$ & $48.49 \pm 0.50$ & $71.27 \pm 1.06$ & $80.20 \pm 1.26$ & $54.85 \pm 1.40$ & $56.99$ \\
            IT (S)
             FWE (L) & $37.03 \pm 1.41$ & $65.61 \pm 0.97$ & $54.91 \pm 0.50$ & $72.74 \pm 1.04$ & $84.20 \pm 1.15$ & $54.85 \pm 1.40$ & $61.56$ \\
            \midrule 
            IT (S) & $23.63 \pm 1.25$ & $29.19 \pm 0.95$ & $28.49 \pm 0.47$ & $56.31 \pm 1.16$ & $36.76 \pm 1.56$ & $48.01 \pm 1.42$ & $37.07$ \\
            IT (L) & $26.33 \pm 1.30$ & $40.38 \pm 1.03$ & $39.89 \pm 0.51$ & $64.74 \pm 1.11$ & $61.76 \pm 1.58$ & $51.42 \pm 1.42$ & $47.42$ \\
            IT (S)
             EN(L) & $25.81 \pm 1.29$ & $36.02 \pm 1.01$ & $35.29 \pm 0.50$ & $58.65 \pm 1.15$ & $58.82 \pm 1.60$ & $52.23 \pm 1.42$ & $44.47$ \\
            IT (S)
             DCLM EN (L) & $24.93 \pm 1.28$ & $32.01 \pm 0.98$ & $31.76 \pm 0.49$ & $54.95 \pm 1.16$ & $61.97 \pm 1.57$ & $53.20 \pm 1.42$ & $43.14$ \\
            IT (S)
             ARC EN (L) & $25.72 \pm 1.29$ & $39.15 \pm 1.02$ & $36.18 \pm 0.50$ & $60.55 \pm 1.14$ & $63.24 \pm 1.56$ & $51.34 \pm 1.42$ & $46.03$ \\
            IT (S)
             S+I EN(L) & $26.07 \pm 1.30$ & $40.03 \pm 1.03$ & $36.67 \pm 0.50$ & $58.76 \pm 1.15$ & $62.29 \pm 1.57$ & $50.93 \pm 1.42$ & $45.79$ \\
            IT (S)
             FWE (L) & $29.90 \pm 1.35$ & $41.04 \pm 1.03$ & $38.24 \pm 0.51$ & $62.68 \pm 1.13$ & $61.03 \pm 1.58$ & $51.18 \pm 1.42$ & $47.34$ \\
            \bottomrule
        \end{tabular}
    }
    \caption{Evaluation of $1$B parameter XL model on ``General Understanding Tasks''  focusing on general reasoning, language understanding, and science knowledge in English followed by Italian. Results show the length normalized accuracy for individual datasets and the average over all datasets for all datasets.}
    \label{tab:lm_eval_1B_it_en_indiv}
\end{table*}

\begin{table*}[h!]
    \centering
    \scalebox{0.7}{
        \begin{tabular}{lcccccc>{\columncolor{LightCyan}}c}
            \toprule
            Model Name & ARC-C & ARC-E & HS & PIQA & SCIQ & WGrande & AVG \\
            \midrule
            KO (S) & $23.04 \pm 1.23$ & $26.14 \pm 0.90$ & $25.41 \pm 0.43$ & $50.98 \pm 1.17$ & $29.50 \pm 1.44$ & $52.57 \pm 1.40$ & $34.61$ \\
            KO (L) & $22.10 \pm 1.21$ & $37.46 \pm 0.99$ & $28.99 \pm 0.45$ & $59.19 \pm 1.15$ & $60.70 \pm 1.55$ & $51.62 \pm 1.40$ & $43.34$ \\
            KO (S)
             EN(L) & $26.02 \pm 1.28$ & $47.47 \pm 1.02$ & $47.71 \pm 0.50$ & $71.11 \pm 1.06$ & $75.00 \pm 1.37$ & $52.72 \pm 1.40$ & $53.34$ \\
            KO (S)
             DCLM EN (L) & $30.03 \pm 1.34$ & $56.69 \pm 1.02$ & $50.38 \pm 0.50$ & $71.76 \pm 1.05$ & $78.90 \pm 1.29$ & $53.20 \pm 1.40$ & $56.83$ \\
            KO (S)
             ARC EN (L) & $30.03 \pm 1.34$ & $57.37 \pm 1.01$ & $49.41 \pm 0.50$ & $73.78 \pm 1.03$ & $78.20 \pm 1.31$ & $53.12 \pm 1.40$ & $56.98$ \\
            KO (S)
             S+I EN(L) & $30.97 \pm 1.35$ & $58.54 \pm 1.01$ & $48.16 \pm 0.50$ & $71.49 \pm 1.05$ & $80.30 \pm 1.26$ & $52.33 \pm 1.40$ & $56.97$ \\
            KO (S)
             FWE (L) & $36.01 \pm 1.40$ & $63.59 \pm 0.99$ & $54.41 \pm 0.50$ & $72.96 \pm 1.04$ & $80.10 \pm 1.26$ & $55.56 \pm 1.40$ & $60.44$ \\
            \midrule 
            KO (S) & $24.41 \pm 1.27$ & $28.84 \pm 0.95$ & $27.63 \pm 0.45$ & $52.39 \pm 1.17$ & $47.41 \pm 1.63$ & $50.89 \pm 1.46$ & $38.60$ \\
            KO (L) & $28.07 \pm 1.33$ & $42.18 \pm 1.04$ & $35.48 \pm 0.48$ & $60.66 \pm 1.14$ & $71.64 \pm 1.47$ & $49.28 \pm 1.46$ & $47.89$ \\
            KO (S)
             EN(L) & $22.76 \pm 1.24$ & $33.82 \pm 0.99$ & $28.88 \pm 0.45$ & $55.22 \pm 1.16$ & $56.08 \pm 1.62$ & $51.23 \pm 1.46$ & $41.33$ \\
            KO (S)
             DCLM EN (L) & $24.85 \pm 1.28$ & $34.13 \pm 1.00$ & $29.35 \pm 0.45$ & $55.55 \pm 1.16$ & $63.92 \pm 1.56$ & $52.34 \pm 1.46$ & $43.35$ \\
            KO (S)
             ARC EN (L) & $25.28 \pm 1.28$ & $33.73 \pm 0.99$ & $29.33 \pm 0.45$ & $53.86 \pm 1.16$ & $64.44 \pm 1.56$ & $51.15 \pm 1.46$ & $42.97$ \\
            KO (S)
             S+I EN(L) & $23.19 \pm 1.25$ & $34.26 \pm 1.00$ & $30.01 \pm 0.46$ & $55.60 \pm 1.16$ & $64.02 \pm 1.56$ & $50.04 \pm 1.46$ & $42.86$ \\
            KO (S)
             FWE (L) & $22.93 \pm 1.24$ & $32.32 \pm 0.98$ & $29.09 \pm 0.45$ & $53.26 \pm 1.16$ & $64.76 \pm 1.55$ & $48.43 \pm 1.46$ & $41.80$ \\
            \bottomrule
        \end{tabular}
    }
    \caption{Evaluation of $1$B parameter XL model on ``General Understanding Tasks''  focusing on general reasoning, language understanding, and science knowledge in English followed by Korean. Results show the length normalized accuracy for individual datasets and the average over all datasets for all datasets.}
    \label{tab:lm_eval_1B_ko_en_indiv}
\end{table*}

\begin{table*}[h!]
    \centering
    \scalebox{0.7}{
        \begin{tabular}{lcccccc>{\columncolor{LightCyan}}c}
                \toprule
                Model Name & ARC-C & ARC-E & HS & PIQA & SCIQ & WGrande & AVG \\
                \midrule
                JA (S) & $22.61 \pm 1.22$ & $28.49 \pm 0.93$ & $26.29 \pm 0.44$ & $53.21 \pm 1.16$ & $33.80 \pm 1.50$ & $51.46 \pm 1.40$ & $35.98$ \\
                JA (L) & $23.46 \pm 1.24$ & $37.42 \pm 0.99$ & $29.07 \pm 0.45$ & $59.14 \pm 1.15$ & $68.10 \pm 1.47$ & $50.28 \pm 1.41$ & $44.58$ \\
                JA (S)
                 EN(L) & $26.19 \pm 1.28$ & $49.24 \pm 1.03$ & $48.69 \pm 0.50$ & $72.63 \pm 1.04$ & $74.40 \pm 1.38$ & $53.28 \pm 1.40$ & $54.07$ \\
                JA (S)
                DCLM EN (L) & $29.78 \pm 1.34$ & $55.30 \pm 1.02$ & $50.26 \pm 0.50$ & $72.14 \pm 1.05$ & $78.90 \pm 1.29$ & $52.72 \pm 1.40$ & $56.52$ \\
                JA (S)
                 ARC EN (L) & $29.35 \pm 1.33$ & $56.94 \pm 1.02$ & $49.05 \pm 0.50$ & $73.78 \pm 1.03$ & $79.80 \pm 1.27$ & $54.22 \pm 1.40$ & $57.19$ \\
                JA (S)
                S+I EN(L) & $30.38 \pm 1.34$ & $56.14 \pm 1.02$ & $48.32 \pm 0.50$ & $72.42 \pm 1.04$ & $78.60 \pm 1.30$ & $52.64 \pm 1.40$ & $56.42$ \\
                JA (S)
                 FWE EN (L) & $36.09 \pm 1.40$ & $63.93 \pm 0.99$ & $54.74 \pm 0.50$ & $73.18 \pm 1.03$ & $80.00 \pm 1.27$ & $55.80 \pm 1.40$ & $60.62$ \\
                \midrule
                JA (S) & $22.58 \pm 1.24$ & $30.07 \pm 0.96$ & $29.32 \pm 0.45$ & $53.81 \pm 1.16$ & $41.79 \pm 1.62$ & $48.81 \pm 1.51$ & $37.73$ \\
                JA (L) & $25.28 \pm 1.28$ & $40.69 \pm 1.03$ & $35.11 \pm 0.48$ & $58.98 \pm 1.15$ & $69.98 \pm 1.51$ & $51.19 \pm 1.51$ & $46.87$ \\
                JA (S)
                 EN(L) & $25.28 \pm 1.28$ & $36.15 \pm 1.01$ & $31.71 \pm 0.46$ & $56.26 \pm 1.16$ & $66.41 \pm 1.55$ & $50.82 \pm 1.51$ & $44.44$ \\
                JA (S)
                DCLM EN (L) & $26.50 \pm 1.30$ & $36.99 \pm 1.01$ & $31.11 \pm 0.46$ & $56.96 \pm 1.16$ & $67.60 \pm 1.54$ & $50.18 \pm 1.51$ & $44.89$ \\
                JA (S)
                 ARC EN (L) & $27.55 \pm 1.32$ & $37.43 \pm 1.02$ & $31.56 \pm 0.46$ & $57.29 \pm 1.15$ & $65.55 \pm 1.56$ & $48.72 \pm 1.51$ & $44.68$ \\
                JA (S)
                S+I EN(L) & $27.38 \pm 1.32$ & $36.42 \pm 1.01$ & $31.44 \pm 0.46$ & $56.58 \pm 1.16$ & $67.28 \pm 1.54$ & $50.36 \pm 1.51$ & $44.91$ \\
                JA (S)
                 FWE EN (L) & $25.89 \pm 1.29$ & $35.71 \pm 1.01$ & $31.06 \pm 0.46$ & $56.20 \pm 1.16$ & $66.09 \pm 1.56$ & $49.64 \pm 1.51$ & $44.10$ \\
                \bottomrule
        \end{tabular}
    }
    \caption{Evaluation of $1$B parameter XL model on ``General Understanding Tasks''  focusing on general reasoning, language understanding, and science knowledge in English followed by translated Japanese. Results show the length normalized accuracy for individual datasets and the average over all datasets for all datasets.}
    \label{tab:lm_eval_1B_ja_en_indiv}
\end{table*}

\begin{table*}[h!]
    \centering
    \scalebox{0.7}{
        \begin{tabular}{lcccccc>{\columncolor{LightCyan}}c}
            \toprule
            Model Name & ARC-C & ARC-E & HS & PIQA & SCIQ & WGrande & AVG \\
            \midrule
           ZH (S) & $22.10 \pm 1.21$ & $26.14 \pm 0.90$ & $25.74 \pm 0.44$ & $52.50 \pm 1.17$ & $31.10 \pm 1.46$ & $50.43 \pm 1.41$ & $34.67$ \\
            ZH (L) & $21.16 \pm 1.19$ & $33.16 \pm 0.97$ & $27.63 \pm 0.45$ & $55.98 \pm 1.16$ & $56.60 \pm 1.57$ & $49.09 \pm 1.41$ & $40.61$ \\
            ZH (S)
             EN(L) & $25.85 \pm 1.28$ & $47.90 \pm 1.03$ & $48.54 \pm 0.50$ & $71.93 \pm 1.05$ & $73.90 \pm 1.39$ & $52.64 \pm 1.40$ & $53.46$ \\
            ZH (S)
             DCLM EN (L) & $28.84 \pm 1.32$ & $55.85 \pm 1.02$ & $49.63 \pm 0.50$ & $71.22 \pm 1.06$ & $78.70 \pm 1.30$ & $53.12 \pm 1.40$ & $56.23$ \\
            ZH (S)
             ARC EN (L) & $30.72 \pm 1.35$ & $57.49 \pm 1.01$ & $48.38 \pm 0.50$ & $73.29 \pm 1.03$ & $80.10 \pm 1.26$ & $53.75 \pm 1.40$ & $57.29$ \\
            ZH (S)
             S+I EN(L) & $30.38 \pm 1.34$ & $56.31 \pm 1.02$ & $47.66 \pm 0.50$ & $71.16 \pm 1.06$ & $78.10 \pm 1.31$ & $52.49 \pm 1.40$ & $56.02$ \\
            ZH (S)
             FWE (L) & $36.18 \pm 1.40$ & $67.13 \pm 0.96$ & $54.07 \pm 0.50$ & $73.99 \pm 1.02$ & $80.90 \pm 1.24$ & $55.96 \pm 1.40$ & $61.37$ \\
            \midrule
            ZH (S) & $22.77 \pm 1.24$ & $28.58 \pm 0.95$ & $28.46 \pm 0.47$ & $50.71 \pm 1.17$ & $40.80 \pm 1.55$ & $48.06 \pm 1.54$ & $36.56$ \\
            ZH (L) & $25.65 \pm 1.29$ & $38.88 \pm 1.02$ & $33.07 \pm 0.49$ & $56.37 \pm 1.16$ & $69.90 \pm 1.45$ & $49.86 \pm 1.54$ & $45.62$ \\
            ZH (S)
             EN(L) & $25.22 \pm 1.28$ & $36.77 \pm 1.01$ & $31.62 \pm 0.48$ & $56.09 \pm 1.16$ & $68.00 \pm 1.48$ & $48.35 \pm 1.54$ & $44.34$ \\
            ZH (S)
             DCLM EN (L) & $23.56 \pm 1.25$ & $38.22 \pm 1.02$ & $32.45 \pm 0.49$ & $54.68 \pm 1.16$ & $68.50 \pm 1.47$ & $52.79 \pm 1.53$ & $45.03$ \\
            ZH (S)
             ARC EN (L) & $23.21 \pm 1.25$ & $37.52 \pm 1.02$ & $32.06 \pm 0.48$ & $55.44 \pm 1.16$ & $70.90 \pm 1.44$ & $48.54 \pm 1.54$ & $44.61$ \\
            ZH (S)
             S+I EN(L) & $22.16 \pm 1.23$ & $38.00 \pm 1.02$ & $32.24 \pm 0.49$ & $54.03 \pm 1.16$ & $68.60 \pm 1.47$ & $53.07 \pm 1.53$ & $44.68$ \\
            ZH (S)
             FWE (L) & $25.04 \pm 1.28$ & $36.50 \pm 1.01$ & $31.89 \pm 0.48$ & $54.62 \pm 1.16$ & $66.30 \pm 1.50$ & $51.94 \pm 1.54$ & $44.38$ \\
            \bottomrule
        \end{tabular}
    }
    \caption{Evaluation of $1$B parameter XL model on ``General Understanding Tasks''  focusing on general reasoning, language understanding, and science knowledge in English followed by translated Chinese. Results show the length normalized accuracy for individual datasets and the average over all datasets for all datasets.}
    \label{tab:lm_eval_1B_zh_en_indiv}
\end{table*}

\section{Results for Individual datasets for German Target Language Models}
\label{sec:app_indiv}

Results for the 300M models on individual eval datasets are also provides in Tables~\ref{tab:lm_eval_300M_de_en}-\ref{tab:lm_eval_300M_de_de}. Results for 1B models on English evaluation tasks are shown in Table~\ref{tab:lm_eval_1B_de_en} and for translated German benchmarks in Table~\ref{tab:lm_eval_1B_de_de}.

\begin{table*}[h!]
    \centering
    \scalebox{0.6}{
        \begin{tabular}{lcccccc>{\columncolor{LightCyan}}c}
        \midrule
Small DE & $21.33 \pm 1.20$ & $27.90 \pm 0.92$ & $26.20 \pm 0.44$ & $51.20 \pm 1.17$ & $42.70 \pm 1.56$ & $49.72 \pm 1.41$ & $36.51$ \\
No ARC Large DE & $21.08 \pm 1.19$ & $31.52 \pm 0.95$ & $28.09 \pm 0.45$ & $56.47 \pm 1.16$ & $54.80 \pm 1.57$ & $51.30 \pm 1.40$ & $40.54$ \\
Large DE & $20.73 \pm 1.18$ & $33.33 \pm 0.97$ & $28.15 \pm 0.45$ & $57.40 \pm 1.15$ & $56.30 \pm 1.57$ & $50.59 \pm 1.41$ & $41.09$ \\
Small DE
+ Large EN & $23.81 \pm 1.24$ & $41.92 \pm 1.01$ & $35.73 \pm 0.48$ & $67.03 \pm 1.10$ & $66.10 \pm 1.50$ & $50.75 \pm 1.41$ & $47.56$ \\
\midrule 
Small DE
+ Large EN DCLM Filter & $25.43 \pm 1.27$ & $46.17 \pm 1.02$ & $35.82 \pm 0.48$ & $67.19 \pm 1.10$ & $67.50 \pm 1.48$ & $52.49 \pm 1.40$ & $49.10$ \\
\midrule 
Small DE
+ Large HS EN & $23.72 \pm 1.24$ & $42.42 \pm 1.01$ & $39.64 \pm 0.49$ & $70.13 \pm 1.07$ & $67.20 \pm 1.49$ & $50.43 \pm 1.41$ & $48.93$ \\
Small DE
 +Large ARC EN & $26.28 \pm 1.29$ & $46.80 \pm 1.02$ & $36.44 \pm 0.48$ & $67.46 \pm 1.09$ & $70.30 \pm 1.45$ & $51.38 \pm 1.40$ & $49.78$ \\
\midrule 
+ Large S EN & $26.62 \pm 1.29$ & $49.33 \pm 1.03$ & $31.94 \pm 0.47$ & $63.44 \pm 1.12$ & $72.40 \pm 1.41$ & $50.83 \pm 1.41$ & $49.09$ \\
Small DE
+ Large S+I EN & $26.19 \pm 1.28$ & $46.93 \pm 1.02$ & $36.02 \pm 0.48$ & $66.21 \pm 1.10$ & $73.10 \pm 1.40$ & $50.67 \pm 1.41$ & $49.85$ \\
\midrule 
Small DE
+ Large v1 & $21.16 \pm 1.19$ & $36.83 \pm 0.99$ & $29.19 \pm 0.45$ & $60.07 \pm 1.14$ & $63.50 \pm 1.52$ & $52.33 \pm 1.40$ & $43.84$ \\
Small DE
+ Large v2 & $21.16 \pm 1.19$ & $34.89 \pm 0.98$ & $29.60 \pm 0.46$ & $57.45 \pm 1.15$ & $59.50 \pm 1.55$ & $50.36 \pm 1.41$ & $42.16$ \\
Small DE
+ Large v3 & $19.54 \pm 1.16$ & $35.02 \pm 0.98$ & $29.46 \pm 0.45$ & $59.47 \pm 1.15$ & $61.40 \pm 1.54$ & $50.51 \pm 1.41$ & $42.57$ \\
\midrule 
RPJv2 & $25.09 \pm 1.27$ & $43.27 \pm 1.02$ & $37.23 \pm 0.48$ & $65.02 \pm 1.11$ & $66.30 \pm 1.50$ & $49.80 \pm 1.41$ & $47.78$ \\
RefinedWeb & $24.40 \pm 1.26$ & $43.98 \pm 1.02$ & $39.75 \pm 0.49$ & $68.66 \pm 1.08$ & $69.80 \pm 1.45$ & $52.49 \pm 1.40$ & $49.85$ \\
FineWeb & $25.00 \pm 1.27$ & $44.23 \pm 1.02$ & $40.89 \pm 0.49$ & $69.53 \pm 1.07$ & $68.40 \pm 1.47$ & $51.78 \pm 1.40$ & $49.97$ \\
FineWebEDU & $28.67 \pm 1.32$ & $56.06 \pm 1.02$ & $40.85 \pm 0.49$ & $66.65 \pm 1.10$ & $72.60 \pm 1.41$ & $52.09 \pm 1.40$ & $52.82$ \\
\bottomrule
        \end{tabular}
        }
    \caption{Evaluation of $300$M parameter medium model on ``General Understanding Tasks''  focusing on general reasoning, language understanding, and science knowledge in English. Results show the length normalized accuracy for individual datasets and the average over all datasets for all datasets.}
    \label{tab:lm_eval_300M_de_en}
\end{table*}

\begin{table*}[h!]
    \centering
    \scalebox{0.6}{
        \begin{tabular}{lcccccc>{\columncolor{LightCyan}}c}
        \toprule
Model Name & ARC-C-DE & ARC-E-DE & HS-DE & PIQA-DE & SCIQ-DE & WGrande-DE & AVG-DE \\
\midrule
Small DE & $23.83 \pm 1.26$ & $30.71 \pm 0.97$ & $28.33 \pm 0.47$ & $53.92 \pm 1.16$ & $47.58 \pm 1.62$ & $52.20 \pm 1.45$ & $39.43$ \\
No ARC Large DE & $23.48 \pm 1.26$ & $32.30 \pm 0.98$ & $30.45 \pm 0.48$ & $56.09 \pm 1.16$ & $60.32 \pm 1.59$ & $52.11 \pm 1.45$ & $42.46$ \\
Large DE & $24.98 \pm 1.28$ & $34.87 \pm 1.00$ & $32.28 \pm 0.48$ & $59.79 \pm 1.14$ & $61.26 \pm 1.58$ & $51.52 \pm 1.45$ & $44.12$ \\
Small DE
+ Large EN & $23.83 \pm 1.26$ & $30.66 \pm 0.97$ & $29.69 \pm 0.47$ & $56.37 \pm 1.16$ & $60.84 \pm 1.58$ & $51.01 \pm 1.45$ & $42.07$ \\
\midrule 
Small DE
+ Large EN DCLM Filter & $24.27 \pm 1.27$ & $32.74 \pm 0.99$ & $29.68 \pm 0.47$ & $55.88 \pm 1.16$ & $60.42 \pm 1.59$ & $51.69 \pm 1.45$ & $42.45$ \\
\midrule 
Small DE
+ Large HS EN & $23.75 \pm 1.26$ & $31.81 \pm 0.98$ & $30.32 \pm 0.47$ & $57.40 \pm 1.15$ & $59.68 \pm 1.59$ & $51.52 \pm 1.45$ & $42.41$ \\
Small DE
 +Large ARC EN & $23.83 \pm 1.26$ & $33.05 \pm 0.99$ & $29.59 \pm 0.47$ & $56.37 \pm 1.16$ & $60.11 \pm 1.59$ & $51.60 \pm 1.45$ & $42.43$ \\
\midrule 
Small DE
+ Large S EN & $23.66 \pm 1.26$ & $32.21 \pm 0.98$ & $28.84 \pm 0.47$ & $56.26 \pm 1.16$ & $61.58 \pm 1.58$ & $52.96 \pm 1.45$ & $42.58$ \\
Small DE
+ Large S+I EN & $23.22 \pm 1.25$ & $33.67 \pm 0.99$ & $29.42 \pm 0.47$ & $55.93 \pm 1.16$ & $61.68 \pm 1.58$ & $51.27 \pm 1.45$ & $42.53$ \\
\midrule 
Small DE
+ Large v1 & $23.92 \pm 1.27$ & $35.27 \pm 1.01$ & $32.79 \pm 0.49$ & $59.74 \pm 1.14$ & $62.42 \pm 1.57$ & $50.68 \pm 1.45$ & $44.14$ \\
Small DE
+ Large v2 & $24.27 \pm 1.27$ & $36.19 \pm 1.01$ & $32.72 \pm 0.48$ & $59.41 \pm 1.15$ & $63.89 \pm 1.56$ & $52.53 \pm 1.45$ & $44.84$ \\
Small DE
+ Large v3 & $23.83 \pm 1.26$ & $36.19 \pm 1.01$ & $34.06 \pm 0.49$ & $61.04 \pm 1.14$ & $63.26 \pm 1.56$ & $51.77 \pm 1.45$ & $45.03$ \\
\midrule 
Small DE
+ Large RPJv2 & $24.10 \pm 1.27$ & $33.10 \pm 0.99$ & $30.06 \pm 0.47$ & $55.44 \pm 1.16$ & $59.05 \pm 1.60$ & $50.08 \pm 1.45$ & $41.97$ \\
Small DE
+ Large RFW & $25.33 \pm 1.29$ & $32.12 \pm 0.98$ & $30.56 \pm 0.48$ & $57.34 \pm 1.15$ & $62.32 \pm 1.57$ & $51.77 \pm 1.45$ & $43.24$ \\
Small DE
+ Large FWE & $25.59 \pm 1.29$ & $35.00 \pm 1.00$ & $31.19 \pm 0.48$ & $56.86 \pm 1.16$ & $62.32 \pm 1.57$ & $51.52 \pm 1.45$ & $43.75$ \\
\bottomrule
        \end{tabular}
    }
    \caption{Evaluation of $300$M parameter model on ``General Understanding Tasks''  focusing on general reasoning, language understanding, and science knowledge in translated German. Results show the length normalized accuracy for individual datasets and the average over all datasets for all datasets.}
    \label{tab:lm_eval_300M_de_de}
\end{table*}

\begin{table*}[h!]
    \centering
    \scalebox{0.6}{
        \begin{tabular}{lcccccc>{\columncolor{LightCyan}}c}
        \toprule
Model Name & ARC-C & ARC-E & HS & PIQA & SCIQ & WGrande & AVG \\
\midrule
Small DE & $20.73 \pm 1.18$ & $27.53 \pm 0.92$ & $25.94 \pm 0.44$ & $52.88 \pm 1.16$ & $36.80 \pm 1.53$ & $47.12 \pm 1.40$ & $35.17$ \\
No ARC Large DE & $21.33 \pm 1.20$ & $35.40 \pm 0.98$ & $30.88 \pm 0.46$ & $59.14 \pm 1.15$ & $60.50 \pm 1.55$ & $50.67 \pm 1.41$ & $42.99$ \\
Large DE & $22.78 \pm 1.23$ & $37.92 \pm 1.00$ & $33.06 \pm 0.47$ & $62.35 \pm 1.13$ & $64.70 \pm 1.51$ & $51.14 \pm 1.40$ & $45.33$ \\
Small DE
+ Large EN & $25.94 \pm 1.28$ & $48.15 \pm 1.03$ & $48.50 \pm 0.50$ & $71.55 \pm 1.05$ & $73.10 \pm 1.40$ & $52.01 \pm 1.40$ & $53.21$ \\
\midrule
Small DE
+ Large EN DCLM Filter & $29.52 \pm 1.33$ & $57.03 \pm 1.02$ & $49.96 \pm 0.50$ & $71.87 \pm 1.05$ & $77.90 \pm 1.31$ & $53.83 \pm 1.40$ & $56.69$ \\
\midrule
Small DE
+ Large HS EN & $28.84 \pm 1.32$ & $49.83 \pm 1.03$ & $54.83 \pm 0.50$ & $75.14 \pm 1.01$ & $75.10 \pm 1.37$ & $54.85 \pm 1.40$ & $56.43$ \\
Small DE
 +Large ARC EN & $31.83 \pm 1.36$ & $57.24 \pm 1.02$ & $48.80 \pm 0.50$ & $73.07 \pm 1.04$ & $77.00 \pm 1.33$ & $52.72 \pm 1.40$ & $56.78$ \\
\midrule
Small DE
+ Large S EN & $30.63 \pm 1.35$ & $57.28 \pm 1.02$ & $39.79 \pm 0.49$ & $68.28 \pm 1.09$ & $80.40 \pm 1.26$ & $53.91 \pm 1.40$ & $55.05$ \\
Small DE
+ Large S+I EN & $29.10 \pm 1.33$ & $55.22 \pm 1.02$ & $48.34 \pm 0.50$ & $71.71 \pm 1.05$ & $78.50 \pm 1.30$ & $53.04 \pm 1.40$ & $55.98$ \\
\midrule
Small DE + Large v1 & $20.73 \pm 1.18$ & $38.97 \pm 1.00$ & $36.09 \pm 0.48$ & $63.33 \pm 1.12$ & $65.40 \pm 1.51$ & $52.01 \pm 1.40$ & $46.09$ \\
Small DE + Large v2 & $22.70 \pm 1.22$ & $38.97 \pm 1.00$ & $36.55 \pm 0.48$ & $64.80 \pm 1.11$ & $69.30 \pm 1.46$ & $51.22 \pm 1.40$ & $47.26$ \\
Small DE + Large v3 & $20.65 \pm 1.18$ & $40.36 \pm 1.01$ & $36.57 \pm 0.48$ & $63.66 \pm 1.12$ & $70.50 \pm 1.44$ & $51.54 \pm 1.40$ & $47.21$ \\
\midrule
Small DE
+ Large RPJv2 & $26.96 \pm 1.30$ & $50.42 \pm 1.03$ & $51.10 \pm 0.50$ & $70.57 \pm 1.06$ & $77.60 \pm 1.32$ & $55.64 \pm 1.40$ & $55.38$ \\
Small DE
+ Large RFW & $27.90 \pm 1.31$ & $54.59 \pm 1.02$ & $54.91 \pm 0.50$ & $73.23 \pm 1.03$ & $77.60 \pm 1.32$ & $56.35 \pm 1.39$ & $57.43$ \\
Small DE
+ Large FWE & $38.14 \pm 1.42$ & $66.37 \pm 0.97$ & $54.88 \pm 0.50$ & $72.25 \pm 1.04$ & $84.60 \pm 1.14$ & $56.04 \pm 1.39$ & $62.05$ \\
\bottomrule
        \end{tabular}
    }
    \caption{Evaluation of $1.3$B parameter model on ``General Understanding Tasks''  focusing on general reasoning, language understanding, and science knowledge in English. Results show the length normalized accuracy for individual datasets and the average over all datasets for all datasets.}
    \label{tab:lm_eval_1B_de_en}
\end{table*}

\begin{table*}[h!]
    \centering
    \scalebox{0.6}{
        \begin{tabular}{lcccccc>{\columncolor{LightCyan}}c}
\toprule
Model Name & ARC-C-DE & ARC-E-DE & HS-DE & PIQA-DE & SCIQ-DE & WGrande-DE & AVG-DE \\
\midrule
Small DE & $22.69 \pm 1.24$ & $30.31 \pm 0.97$ & $28.99 \pm 0.47$ & $54.84 \pm 1.16$ & $40.00 \pm 1.59$ & $50.42 \pm 1.45$ & $37.88$ \\
No ARC Large DE & $25.07 \pm 1.29$ & $36.95 \pm 1.02$ & $35.64 \pm 0.49$ & $59.74 \pm 1.14$ & $64.53 \pm 1.55$ & $52.62 \pm 1.45$ & $45.76$ \\
Large DE & $27.44 \pm 1.32$ & $38.81 \pm 1.03$ & $39.53 \pm 0.51$ & $63.17 \pm 1.13$ & $67.68 \pm 1.52$ & $52.62 \pm 1.45$ & $48.21$ \\
Small DE
+ Large EN & $25.07 \pm 1.29$ & $37.70 \pm 1.02$ & $36.18 \pm 0.50$ & $59.36 \pm 1.15$ & $63.05 \pm 1.57$ & $50.51 \pm 1.45$ & $45.31$ \\
\midrule 
Small DE
+ Large EN DCLM Filter & $25.77 \pm 1.30$ & $40.00 \pm 1.03$ & $35.94 \pm 0.50$ & $59.25 \pm 1.15$ & $65.05 \pm 1.55$ & $51.18 \pm 1.45$ & $46.20$ \\
\midrule
Small DE
+ Large HS EN & $24.89 \pm 1.28$ & $36.90 \pm 1.02$ & $37.11 \pm 0.50$ & $60.77 \pm 1.14$ & $63.47 \pm 1.56$ & $52.87 \pm 1.45$ & $46.00$ \\
Small DE
 +Large ARC EN & $26.91 \pm 1.32$ & $39.34 \pm 1.03$ & $35.77 \pm 0.50$ & $58.65 \pm 1.15$ & $67.58 \pm 1.52$ & $52.11 \pm 1.45$ & $46.73$ \\
\midrule
+ Large S EN & $27.35 \pm 1.32$ & $39.38 \pm 1.03$ & $33.23 \pm 0.49$ & $58.76 \pm 1.15$ & $64.21 \pm 1.56$ & $51.18 \pm 1.45$ & $45.69$ \\
Small DE
+ Large S+I EN & $25.24 \pm 1.29$ & $39.38 \pm 1.03$ & $35.34 \pm 0.49$ & $59.74 \pm 1.14$ & $64.21 \pm 1.56$ & $52.70 \pm 1.45$ & $46.10$ \\
\midrule 
Small DE + Large v1 & $26.12 \pm 1.30$ & $40.93 \pm 1.03$ & $40.39 \pm 0.51$ & $61.48 \pm 1.14$ & $67.58 \pm 1.52$ & $51.94 \pm 1.45$ & $48.07$ \\
Small DE + Large v2 & $25.51 \pm 1.29$ & $42.30 \pm 1.04$ & $40.75 \pm 0.51$ & $62.19 \pm 1.13$ & $71.37 \pm 1.47$ & $52.45 \pm 1.45$ & $49.09$ \\
Small DE + Large v3 & $26.21 \pm 1.30$ & $40.84 \pm 1.03$ & $43.08 \pm 0.51$ & $64.09 \pm 1.12$ & $66.74 \pm 1.53$ & $51.52 \pm 1.45$ & $48.75$ \\
\midrule
Small DE
+ Large RPJv2 & $24.71 \pm 1.28$ & $37.17 \pm 1.02$ & $36.53 \pm 0.50$ & $58.65 \pm 1.15$ & $65.26 \pm 1.55$ & $52.70 \pm 1.45$ & $45.84$ \\
Small DE
+ Large RFW & $25.42 \pm 1.29$ & $38.81 \pm 1.03$ & $38.25 \pm 0.50$ & $58.38 \pm 1.15$ & $64.21 \pm 1.56$ & $51.01 \pm 1.45$ & $46.01$ \\
Small DE
+ Large FWE & $26.91 \pm 1.32$ & $42.39 \pm 1.04$ & $37.40 \pm 0.50$ & $60.45 \pm 1.14$ & $65.37 \pm 1.54$ & $50.42 \pm 1.45$ & $47.16$ \\
\bottomrule
        \end{tabular}
    }
    \caption{Evaluation of $1.3$B parameter model on ``General Understanding Tasks''  focusing on general reasoning, language understanding, and science knowledge in translated German. Results show the length normalized accuracy for individual datasets and the average over all datasets for all datasets.}
    \label{tab:lm_eval_1B_de_de}
\end{table*}

\end{document}